\documentclass{article}

\PassOptionsToPackage{numbers,sort&compress}{natbib}
\PassOptionsToPackage{table}{xcolor}
\usepackage[eandd,final]{neurips_2026}

\usepackage[utf8]{inputenc}
\usepackage[T1]{fontenc}
\usepackage{hyperref}
\usepackage{url}
\usepackage{booktabs}
\usepackage{amsfonts}
\usepackage{nicefrac}
\usepackage{microtype}
\usepackage{xcolor}

\usepackage{dashrule}
\usepackage{enumitem}
\usepackage{graphicx}
\usepackage{multirow}
\usepackage{algorithm}
\usepackage{algpseudocode}
\usepackage{amsmath}
\usepackage{tabularx}
\usepackage{pifont}
\usepackage{makecell}

\usepackage{wrapfig}
\usepackage{tikz}
\usepackage{tcolorbox}
\tcbuselibrary{listings, breakable}
\usepackage{subcaption}
\newcommand{\pos}[1]{$+#1$}
\newcommand{\nneg}[1]{$-#1$}

\newcommand{\spark}[3]{
  \begin{tikzpicture}[baseline=-0.5ex,x=0.9cm,y=0.30cm]
    \pgfmathsetmacro{\vmin}{min(#1,#2,#3)}
    \pgfmathsetmacro{\vmax}{max(#1,#2,#3)}
    \pgfmathsetmacro{\vrng}{\vmax-\vmin+0.0001}
    \pgfmathsetmacro{\ya}{(#1-\vmin)/\vrng}
    \pgfmathsetmacro{\yb}{(#2-\vmin)/\vrng}
    \pgfmathsetmacro{\yc}{(#3-\vmin)/\vrng}
    \draw[gray!30,line width=0.25pt] (0,0) -- (1,0);
    \draw[gray!30,line width=0.25pt] (0,1) -- (1,1);
    \draw[black,line width=0.7pt] (0,\ya) -- (0.5,\yb) -- (1,\yc);
    \fill[black] (0,\ya) circle (1pt);
    \fill[black] (0.5,\yb) circle (1pt);
    \fill[black] (1,\yc) circle (1pt);
  \end{tikzpicture}
}
\newcommand{\dpos}[1]{\scriptsize$+#1$}
\newcommand{\dneg}[1]{\scriptsize$-#1$}

\usepackage{array}
\hypersetup{
    hidelinks}
\usepackage{rotating}

\title{CardioLens: Revealing the Clinical Reality Gap of MLLMs via Multi-Sequence Cardiac MRI Evaluations}

\author{
Zixian Su$^{1}$\thanks{Equal Contributions.} \quad
Hongkai Zhang$^{2}$\footnotemark[1] \quad
Fan Gao$^{1,3}$\footnotemark[1]\quad
Encheng Su$^{1}$ \quad
Taiping Qu$^{1}$ \\
\textbf{Jingwei Guo}$^{4}$ \quad
\textbf{Nan Zhang}$^{2}$ \quad
\textbf{Hui Wang}$^{2}$ \quad
\textbf{Zhen Zhou}$^{2}$ \quad
\textbf{Kairui Bo}$^{2}$ \quad
\textbf{Yan Chen}$^{2}$ \\
\textbf{Yue Ren}$^{2}$ \quad
\textbf{Shuai Li}$^{3}$ \quad
\textbf{Lei Xu}$^{2}$\thanks{Corresponding author.} \quad
\textbf{Henggui Zhang}$^{1}$\footnotemark[2] \\
\\
$^{1}$Beijing Academy of Artificial Intelligence\\
$^{2}$Beijing Anzhen Hospital, Capital Medical University\\
$^{3}$Beihang University\\
$^{4}$King Abdullah University of Science and Technology\\
\texttt{zixian12138@163.com, leixu2001@mail.ccmu.edu.cn, hgzhang@baai.ac.cn}
}

\providecommand{\CRrecent}[1]{#1}

\begin{document}

\maketitle

\begin{abstract}
Multimodal Large Language Models (MLLMs) have shown strong performance on medical public benchmarks, but existing evaluations remain limited as clinically grounded assessments because they often rely on isolated inputs lacking study-level structure and simplified, recognition-style tasks that are far from real diagnostic workflows. We introduce \textbf{CardioLens}, a leakage-resistant evaluation testbed using multi-sequence Cardiovascular Magnetic Resonance (CMR), constructed from non-public hospital archives via a rigorous report-to-QA construction and verification pipeline. CardioLens is built from 473,896 slices and 13,494 verified QA pairs across 4D Cine, LGE, perfusion, and T2-weighted imaging, and evaluates three stages of CMR interpretation: image understanding, report generation, and disease diagnosis. Across 24 state-of-the-art MLLMs, CardioLens reveals a substantial clinical reality gap. Models perform poorly overall, with performance degrading along the CMR workflow. Confusion analysis further reveals a category-collapse failure mode: models often default to frequent abnormal categories rather than reliably distinguishing clinical categories.
To assess the influence of MLLM-compatible input construction, we compare random, clinically-motivated, and data-driven selection protocols under different slice budgets; performance changes only marginally, typically by about 1\%.
Explicit reasoning prompts also fail to rescue performance, with reduced recall suggesting a shift toward more conservative predictions.
These results show that current MLLMs remain far from reliable CMR interpretation, where clinical decisions require integrating distributed evidence across sequences, views, and temporal phases. By exposing where and how current MLLMs fall short, CardioLens provides a clinically grounded testbed for developing the next generation of MLLMs toward real-world clinical deployment.
\end{abstract}

\section{Introduction}
\label{sec:intro}

\begin{figure*}[t]
    \centering
    \includegraphics[width=\textwidth]{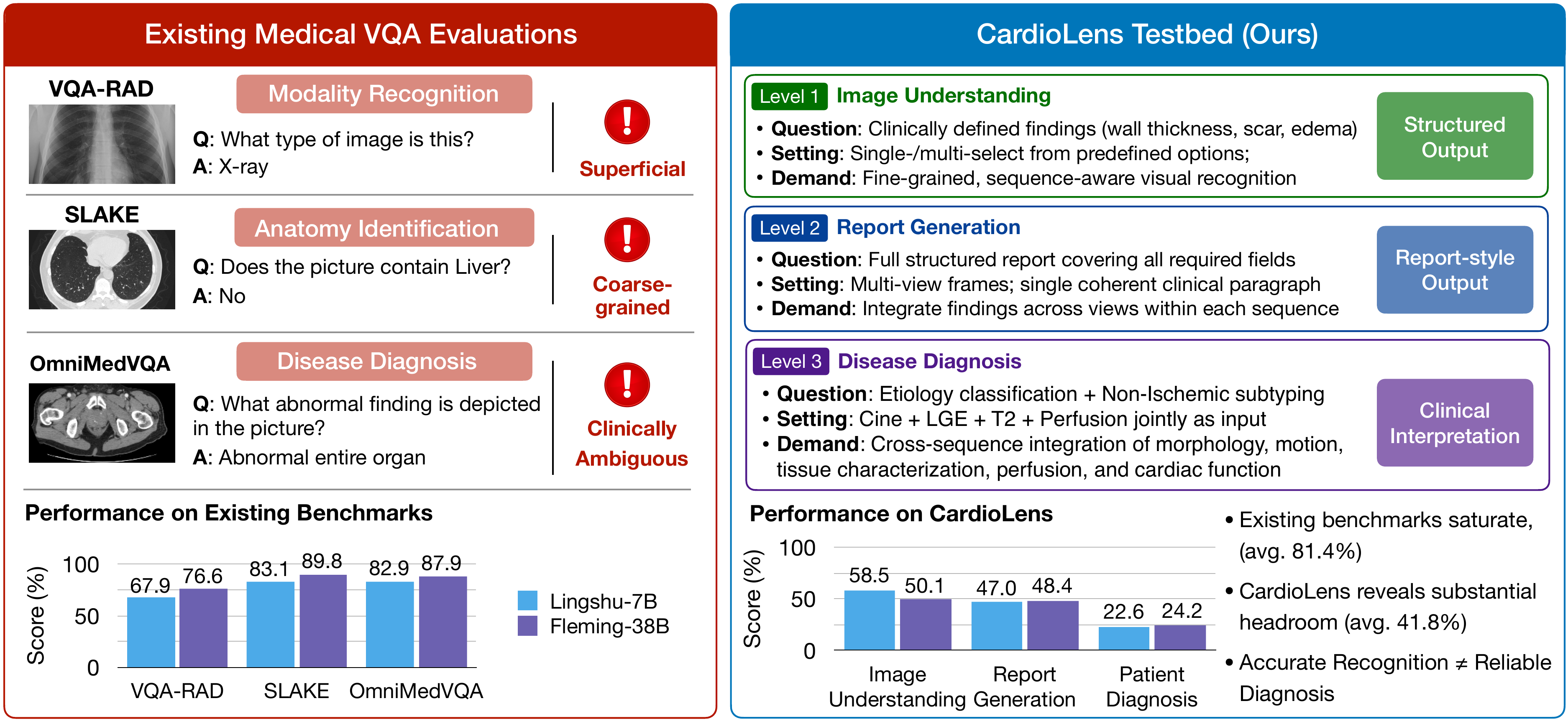}
    \caption{\textbf{Existing medical VQA benchmarks vs. our CardioLens testbed.} Results on VQA-RAD, SLAKE, and OmniMedVQA are adopted from prior reports~\cite{xu2025lingshu,liu2025fleming}.
    }
    \label{fig:data_quality_pipeline}
\end{figure*}

The rapid progress of Multimodal Large Language Models (MLLMs) has been demonstrated by strong results on public medical benchmarks~\cite{achiam2023gpt,wang2024qwen2,guo2025deepseek,wu2025qwen,sellergren2025medgemma}. However, the evidence provided by existing evaluations remains incomplete for clinical assessments. Most current benchmarks suffer from three limitations. First, many are assembled from publicly available datasets, which may overlap with web-scale pretraining corpora and cause data leakage~\cite{jin2020disease,jin2019pubmedqa,hu2024omnimedvqa,chen2024gmai,xia2024cares,bai2024m3d,wu2025towards,gu2025illusion,asadi2026mirage}. Second, their visual inputs are often simplified into isolated images or single-volume cases, whereas real clinical studies are typically high-dimensional, containing multiple anatomical views, temporal phases, and complementary imaging sequences~\cite{lau2018dataset,liu2021slake,hu2024omnimedvqa,chen2024gmai,xia2024cares}. Third, their tasks are frequently framed as image-level recognition or short-form VQA, which can test visual-textual association but does not sufficiently probe the staged evidence synthesis required for clinical decision-making. As a result, strong benchmark performance does not necessarily establish that an MLLM can reason over the complex evidence structure of real medical studies (see Figure~\ref{fig:data_quality_pipeline}).

Cardiovascular Magnetic Resonance (CMR) provides a clinically important setting in which these limitations become especially salient. A routine CMR examination combines multiple complementary sequences
-- Cine for cardiac motion, Late Gadolinium Enhancement (LGE) for myocardial fibrosis and scar, perfusion for microvascular blood flow, and T2-weighted imaging for edema and inflammation -- across multiple anatomical views~\cite{ye2025continuous,koehler2025deep,chiribiri2025society,tavakoli2025scarnet,lehtonen2023cardiac}.
Complete interpretation therefore requires dynamic motion assessment, three-dimensional localization, cross-sequence evidence integration, and patient-level diagnostic synthesis, rather than recognizing isolated findings in single frames. CMR therefore offers a natural test case for evaluating whether medical MLLMs can move beyond image-level perception toward actionable medical intelligence.

In light of this, we introduce \textbf{CardioLens}, a leakage-resistant evaluation testbed for multi-sequence CMR. Developed from non-public hospital archives through a rigorous data construction pipeline, CardioLens comprises 473,896 slices and 13,494 verified report-derived QA pairs from real patient studies, covering 4D Cine, LGE, perfusion, and T2-weighted imaging.
It contains three progressively more demanding tiers along the CMR interpretation workflow: fine-grained image understanding for clinical findings, structured report generation for standardized {clinical synthesis}, and disease diagnosis for patient-level decision-making.
This design enables systematic evaluation of MLLMs on data that are private, high-dimensional, multi-sequence, and aligned with real diagnostic workflows.

Our evaluation reveals a substantial clinical reality gap across 24 state-of-the-art MLLMs.
Models perform poorly overall on CardioLens, with performance degrading as tasks move along the CMR workflow.
This decline indicates that current MLLMs do not reliably translate isolated CMR recognition into clinically meaningful study-level interpretation. Beyond overall performance degradation, our analysis further identifies category collapse as an important failure mode. Confusion matrices show that models often default to frequent abnormal categories rather than reliably distinguishing distinct clinical findings. Such behavior suggests that current MLLMs place excessive weight on common diagnostic categories relative to the available multi-sequence CMR evidence, undermining the reliability of their clinical predictions.

We further examine whether the observed clinical performance gap is influenced by the way complete CMR studies are adapted to MLLM input constraints. Because most current models cannot directly incorporate complete 3D/4D examinations, CMR data must be represented by selected subsets during evaluation. CardioLens therefore compares multiple input-selection strategies -- random sampling, clinically-motivated heuristic sampling, and data-driven sampling -- under different slice budgets. We observe that
varying input-construction choices
only marginally alters performance, typically by about 1\%. {This pattern further confirms that current MLLMs struggle to organize distributed CMR cues into clinically grounded predictions.}

We also use explicit clinical reasoning prompts as a probe to better understand model behavior.
Keeping the visual inputs fixed, we compare standard visual prompts against variants that request structured clinical justifications.
This intervention does not produce consistent gains; instead, performance often decreases as recall falls, a pattern consistent with more conservative positive predictions.
These results reveal a deeper instability in current MLLMs: their mapping of CMR visual evidence to clinical findings remains unreliable.
Under prompts requiring explicit clinical justification, models may still miss genuine abnormalities rather than recover overlooked visual evidence.

Our findings reveal a clear mismatch between current benchmark performance and clinically grounded CMR interpretation. Strong results on conventional public benchmarks do not guarantee reliable performance in multi-sequence, dynamic CMR setting.
Through the lens of multi-sequence Cardiac MRI, we reveal these limitations and provide a controlled testbed for guiding future medical MLLM evaluation from isolated, simplified recognition toward real-world clinical utility.

\section{Related Work}\label{sec:related_work}

\textbf{Advances and Open Challenges in Medical MLLMs.} Recent multimodal large language models (MLLMs) have achieved strong visual-language understanding and instruction following, with general-domain systems such as Qwen-VL, LLaVA, and InternVL~\cite{bai2023qwen,li2024llava,chen2024internvl} forming the basis for many medical variants. Medical MLLMs, including LLaVA-Med, Lingshu, and other radiology-oriented models~\cite{li2023llava,he2024meddr,xu2025lingshu,wu2025towards}, further adapt these capabilities to medical VQA, report generation, and diagnostic assistance. However, most progress is still evaluated under a static image-text alignment paradigm, where clinical competence is approximated by recognizing findings from isolated images~\cite{radford2021learning,alayrac2022flamingo,li2023llava}. This leaves open whether medical MLLMs can handle clinical studies in which evidence is distributed across views, temporal phases, and complementary sequences.
CardioLens probes this gap through image understanding, sequence-level report generation, and patient-level diagnosis from multi-sequence CMR studies.

\textbf{Progress and Evaluation Limitations in Medical Benchmarks.}
Medical MLLM benchmarks have expanded from single-image VQA datasets like VQA-RAD and SLAKE to more diverse evaluations such as OmniMedVQA, GMAI-MMBench, CARES, M3D-Bench~\cite{lau2018dataset,liu2021slake,hu2024omnimedvqa,chen2024gmai,xia2024cares,bai2024m3d} (see more details in Appendix~\ref{app:Datasets_of_Medical_MLLM}). Despite this progress, existing benchmarks remain insufficient for supporting claims of clinical competence, as they often rely on \textit{leakage-prone public data}, \textit{isolated inputs}, and \textit{clinically over-simplified tasks}. As a result, strong benchmark performance may not translate into clinical utility for physicians. CardioLens is designed against all three pitfalls—drawn from a private hospital archive to limit leakage, preserving study-level organization of multi-sequence data, and framing tasks along the actual diagnostic workflow rather than recognition shortcuts.

\textbf{Cardiac MRI for High-Dimensional Clinical Evaluation.}
Cardiac MRI provides a natural clinical context for evaluating MLLMs beyond isolated image recognition, as CMR interpretation requires integrating dynamic function, tissue characterization, perfusion, and edema information across anatomical views and temporal phases~\cite{chiribiri2025society,ye2025continuous,koehler2025deep,tavakoli2025scarnet}. Existing datasets have advanced cardiac image analysis, but they mainly target segmentation or task-specific quantification, including Cine-based ventricular segmentation, LGE-based scar analysis, atrial analysis, and multi-sequence myocardial segmentation~\cite{bernard2018deep,martin2023deep,lalande2020emidec,li2022review,li2023myops}. These resources are valuable for specialized CMR algorithms, but they target component-level tasks rather than the full clinical interpretation pipeline. CardioLens repurposes multi-sequence CMR as a clinically grounded testbed for MLLMs, with tasks spanning image understanding, report generation, and disease diagnosis.

\begin{table}[t]
\centering
\caption{CardioLens statistics. Image understanding subtasks are single-label (S) or multi-label (M) with different classes(\#C). {Full per-sequence clinical roles are described in Appendix~\ref{app:sequence-details}}}
\label{tab:dataset_statistics}
\setlength{\tabcolsep}{4pt}
\renewcommand{\arraystretch}{1.0}
\resizebox{1.0\textwidth}{!}{
\footnotesize
\begin{tabular}{@{}lcrr@{\hskip 10pt}lcrr@{\hskip 10pt}lcrr@{\hskip 10pt}lcrr@{}}
\toprule
\multicolumn{4}{c}{\textbf{Cine (5,223)}} & \multicolumn{4}{c}{\textbf{LGE (3,264)}} & \multicolumn{4}{c}{\textbf{Perfusion (985)}} & \multicolumn{4}{c}{\textbf{T2 (928)}} \\
\cmidrule(r){1-4}\cmidrule(lr){5-8}\cmidrule(lr){9-12}\cmidrule(l){13-16}
Subtask & T & \#C & \#QA & Subtask & T & \#C & \#QA & Subtask & T & \#C & \#QA & Subtask & T & \#C & \#QA \\
\midrule
Wall Thickness      & M & 5 & 580 & Enhance. Status  & S & 2 & 581 & Perf.\ Status       & S & 2 & 565 & T2 Signal        & S & 4 & 507 \\
Special Signs       & M & 6 & 580 & Special Findings    & M & 4 & 581 & Signal Char.        & S & 3 & 129 & Abn.\ Regions    & M & 6 & 144 \\
Pericardial Eff.    & S & 2 & 580 & Abn.\ Signal        & S & 2 & 468 & Abn.\ Regions       & M & 6 & 116 & Abn.\ Segments   & M & 4 & 127 \\
Pleural Eff.        & S & 2 & 580 & High-Sig Pattern      & M & 5 & 412 & Abn.\ Segments      & M & 4 &  91 & Sig.\ Distrib.   & M & 5 & 109 \\
Mitral Valve        & S & 2 & 579 & High-Sig Layer        & M & 5 & 384 & Myo.\ Layer     & M & 5 &  84 & Myo.\ Layer  & M & 5 &  41 \\
Tricuspid Valve     & S & 2 & 579 & High-Sig Region       & M & 6 & 382 &                     &   &   &     &                  &   &   &     \\
Aortic Valve        & S & 2 & 570 & High-Sig Segment      & M & 5 & 378 &                     &   &   &     &                  &   &   &     \\
Motion Ampl.        & M & 5 & 435 & Low-Sig Pattern      & M & 5 &  40 &                     &   &   &     &                  &   &   &     \\
Systolic Func.      & S & 3 & 398 & Low-Sig Region       & M & 6 &  18 &                     &   &   &     &                  &   &   &     \\
Motion Coord.       & S & 2 & 195 & Low-Sig Segment      & M & 4 &  14 &                     &   &   &     &                  &   &   &     \\
Diastolic Func.     & S & 2 & 147 & Low-Sig Layer        & M & 5 &   6 &                     &   &   &     &                  &   &   &     \\
\midrule
\multicolumn{16}{@{}l}{\textbf{Report Generation (2,246):}~~Cine 584~~$\mid$~~LGE 581~~$\mid$~~Perfusion 570~~$\mid$~~T2 511} \\
\midrule
\multicolumn{16}{@{}l}{\textbf{Diagnosis (848):}~~\textit{Etiology (553):} NH 102 / IHD 156 / NICM 295~~$\mid$~~\textit{NICM Subtype (295):} HCM 107 / Myocarditis 101 / DCM 72 / RCM 13 / ACM 2} \\
\bottomrule
\end{tabular}
}
\end{table}

\section{CardioLens}
\label{sec:CardioLens}

\begin{figure*}[t]
    \centering
    \includegraphics[width=0.98\textwidth]{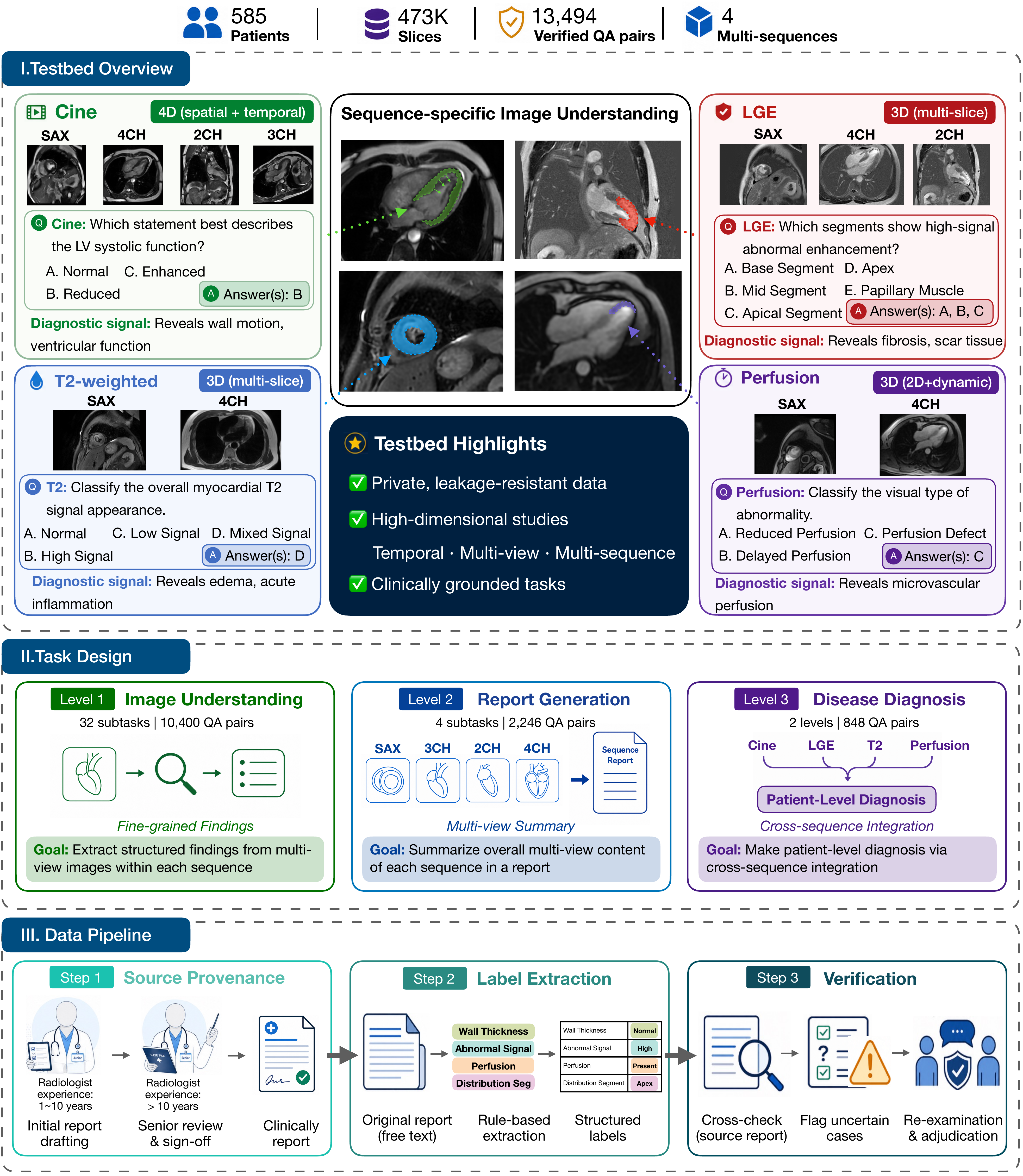}

    \caption{
\textbf{Overview of CardioLens.}
CardioLens is a private multi-sequence CMR evaluation testbed constructed via a radiologist-guided report-to-QA pipeline and organized into three clinical tasks.
}
    \label{fig:CardioLens}
\end{figure*}

CardioLens is constructed from private hospital archives comprising multi-sequence cardiovascular MRI studies from 585 patients. It covers four clinically essential CMR sequences -- Cine, Late Gadolinium Enhancement (LGE), Perfusion, and T2-weighted imaging -- and {is built from} 5,969 imaging files, 473,896 slices, and 13,494 expert-verified question-answer pairs derived from authentic clinical reports. Detailed statistics and an overview are provided in Table~\ref{tab:dataset_statistics} and Figure~\ref{fig:CardioLens}, respectively.

\subsection{Task Definitions}
\label{sec:task_definition}

We design a three-tier evaluation following the CMR interpretation workflow: \textbf{image understanding} identifies sequence-specific findings, \textbf{report generation} synthesizes visual evidence into structured sequence-level reports, and \textbf{disease diagnosis} predicts patient-level disease categories directly from multi-sequence images.

\textbf{Image Understanding (32 subtasks, 10{,}400 QA pairs).}
Image understanding tasks extract structured clinical findings from single-sequence inputs (11 Cine, 11 LGE, 5 T2, 5 Perfusion subtasks), each formulated as single- or multi-label classification. The taxonomy mirrors how radiologists structure their reading: Cine covers motion, wall thickness, ventricular function, valvular regurgitation, and effusions; LGE follows an explicit decision tree from enhancement presence to signal type, then segmental, regional, distributional, and laminar characteristics; T2 and Perfusion progress analogously from signal status to spatial localization and distribution. This evaluates per-sequence perceptual grounding before any cross-sequence or patient-level interpretation.

\textbf{Report Generation (4 subtasks, 2{,}246 QA pairs).}
For each patient, one sequence-level report is generated for Cine, LGE, Perfusion, and T2. Reference reports are constructed by rule-based concatenation of structured image-understanding labels using predefined templates, ensuring logical consistency between report text and structured annotations. This task evaluates whether models can convert distributed visual findings into clinically structured narrative reports.

\textbf{Disease Diagnosis (2 subtasks, 848 QA pairs).}
Diagnosis requires mapping multi-sequence images directly to patient-level disease categories and is organized as two tasks of increasing specificity. The \emph{etiology classification} task distinguishes Normal Heart (NH), Ischemic Heart Disease (IHD), and Non-Ischemic Cardiomyopathy (NICM) -- the most fundamental differential framework in CMR practice; the \emph{NICM subtyping} task further resolves NICM into HCM, DCM, RCM, ACM, and Myocarditis (Appendix~\ref{app:diagnosis-subtasks}). Each subtype shows a characteristic multi-sequence phenotype, demanding full-sequence pattern matching.

\subsection{Data Construction Pipeline}
\label{sec:data_construction}

\begin{itemize}[leftmargin=1.4em, itemsep=0.0em, topsep=0.0em]
    \item \textbf{Source provenance.}
    Each report is first drafted by a radiologist with 1--10 years of cardiac imaging experience and then reviewed and signed off by a senior radiologist with more than 10 years of experience. The finalized clinical report serves as the authoritative source for image-understanding and report-generation targets.

    \item \textbf{Label extraction.}
    Free-text reports are converted into structured subtask labels using rule-based extraction with regular expressions and subtask-specific clinical vocabularies.
    These labels define the answers for image-understanding tasks and provide the structured basis for report-generation references (diagnosis labels are taken from discharge records to reflect clinical conclusions).

    \item \textbf{Verification.}
    Extracted labels are manually checked against the source reports. Uncertain or ambiguous cases are flagged, re-examined, and adjudicated to improve consistency between the structured annotations and the original clinical reports.
\end{itemize}

\subsection{Slice Selection Protocols}\label{sec:slice_selection}

CMR studies are retained in NIfTI format with their spatial and temporal indexing (Appendix~\ref{app:heuristic:framework}).

Because current MLLMs cannot process the full 3D/4D CMR examination directly, each sequence is sampled along its native spatial or temporal axes to select original acquisition slices or frames before inference. CardioLens therefore varies two input-construction factors: the slice-selection protocol and the slice budget $K$. By comparing random, clinically-motivated, and data-driven selection under different budgets, we examine whether the performance gap can be reduced by providing either better-selected slices or more visual evidence.

\begin{itemize}[leftmargin=1.4em, itemsep=0.1em, topsep=0.1em]

\item \textbf{Random Sampling.} Random sampling uniformly selects $K$ slices from the available slices of a given sequence without using clinical prior knowledge. This provides an uninformed baseline for estimating what a model can extract from an arbitrary subset of the study.

\item \textbf{Clinically-motivated Heuristic Sampling.} Heuristic sampling uses radiologist-defined task-to-view priors to select slices that are clinically most relevant to each subtask. For example, Cine short-axis views are prioritized for wall thickness, long-axis views for motion, valvular and functional assessment, and LGE short-axis views for regional enhancement localization. Full details are provided in Appendix~\ref{app:heuristic}.

\item \textbf{Data-driven MIL Sampling.} As a data-driven counterpart to random and heuristic selection, we train a Multiple Instance Learning (MIL) model with sequence-level image-understanding labels, treating each CMR sequence as a bag of slices. The trained model assigns task-specific scores to slices, and the top-$K$ are selected as model inputs. We use this protocol to test whether more label-relevant visual evidence improves MLLM performance. See details in Appendix~\ref{app:mil}.

\end{itemize}

Together, these protocols allow us to observe how evaluation results vary when a full 3D/4D CMR examination is represented by selected slices, helping us assess potential evaluation bias arising from current MLLM input constraints (see Section~\ref{sec:bottleneck_analysis}).

\subsection{Data Governance and Public Release}
\label{sec:data-governance}
All datasets were collected with approval from the Institutional Review Board (IRB) of Beijing Anzhen Hospital affiliated to Capital Medical University (2025216x). All CMR studies were de-identified prior to use. Due to institutional data-sharing constraints, only 100 of the 585 patients are permitted for public release. The released subset preserves the same multi-sequence structure, task definitions, and evaluation protocols as the full benchmark. Detailed statistics and analysis are in Appendix~\ref{app:public_subset}. The public-release package includes NIfTI studies, structured report references, diagnosis labels, and verified report-derived question--answer pairs. Data and code are available at \url{https://huggingface.co/datasets/kiwi12138/CardioLens} and \url{https://github.com/kiwi12138/CardioLens}.

\section{Experiments and Analysis}\label{sec:experiments_analysis}

In this section, we design experiments to answer {the following research questions:}
\textbf{(RQ1)} How do current MLLMs perform across the CMR interpretation workflow, from image understanding to report generation and disease diagnosis?
\textbf{(RQ2)} Does adapting high-dimensional 3D/4D CMR studies into sparse, MLLM-compatible inputs introduce evaluation bias?
\textbf{(RQ3)} Can clinical reasoning instructions improve the performance of MLLMs on our CardioLens, and how do they affect models' prediction behavior?

\subsection{Experimental Setup}
\label{sec:experimental_setup}
\textbf{Models.} We benchmark a diverse set of state-of-the-art MLLMs covering three axes: \textit{closed- vs.\ open-source}, \textit{general-purpose vs.\ medical-specialized}, and \textit{model scale} ($\sim$4B to $\sim$38B parameters). Closed-source systems include GPT-5~\cite{singh2025openai}, Gemini-3.1-Pro~\cite{googledeepmind2026gemini31pro}, and Claude-Opus-4.6~\cite{anthropic2026claudeopus46}. Open-source models include Qwen3-VL~\cite{bai2025qwen3}, InternVL3.5~\cite{wang2025internvl3}, Deepseek-VL2~\cite{wu2024deepseek} and Gemma3~\cite{gemmateam2025gemma3}; medical-specialized models include MedGemma~\cite{sellergren2025medgemma}, Lingshu~\cite{xu2025lingshu}, and other domain-adapted variants~\cite{liu2025fleming,jiang2025hulu,dai2025qoq}. Complete list is provided in Appendix~\ref{app:model-list}.

\textbf{Evaluation Protocol.} In the main benchmark evaluation, the MLLMs are evaluated in a strict \textbf{zero-shot} setting -- no task-specific fine-tuning, no in-context examples -- reflecting the intended deployment scenario of MLLMs as general-purpose clinical assistants. Each instance provides the selected slices and a task-specific textual prompt, and the model must produce a structured response.  Prompt templates are provided in Appendix~\ref{app:prompts}, and metric definitions are provided in Appendix~\ref{app:metrics}.

\begin{figure*}[t]
    \centering

    \begin{minipage}{0.61\textwidth}
        \centering
        \includegraphics[width=\linewidth]{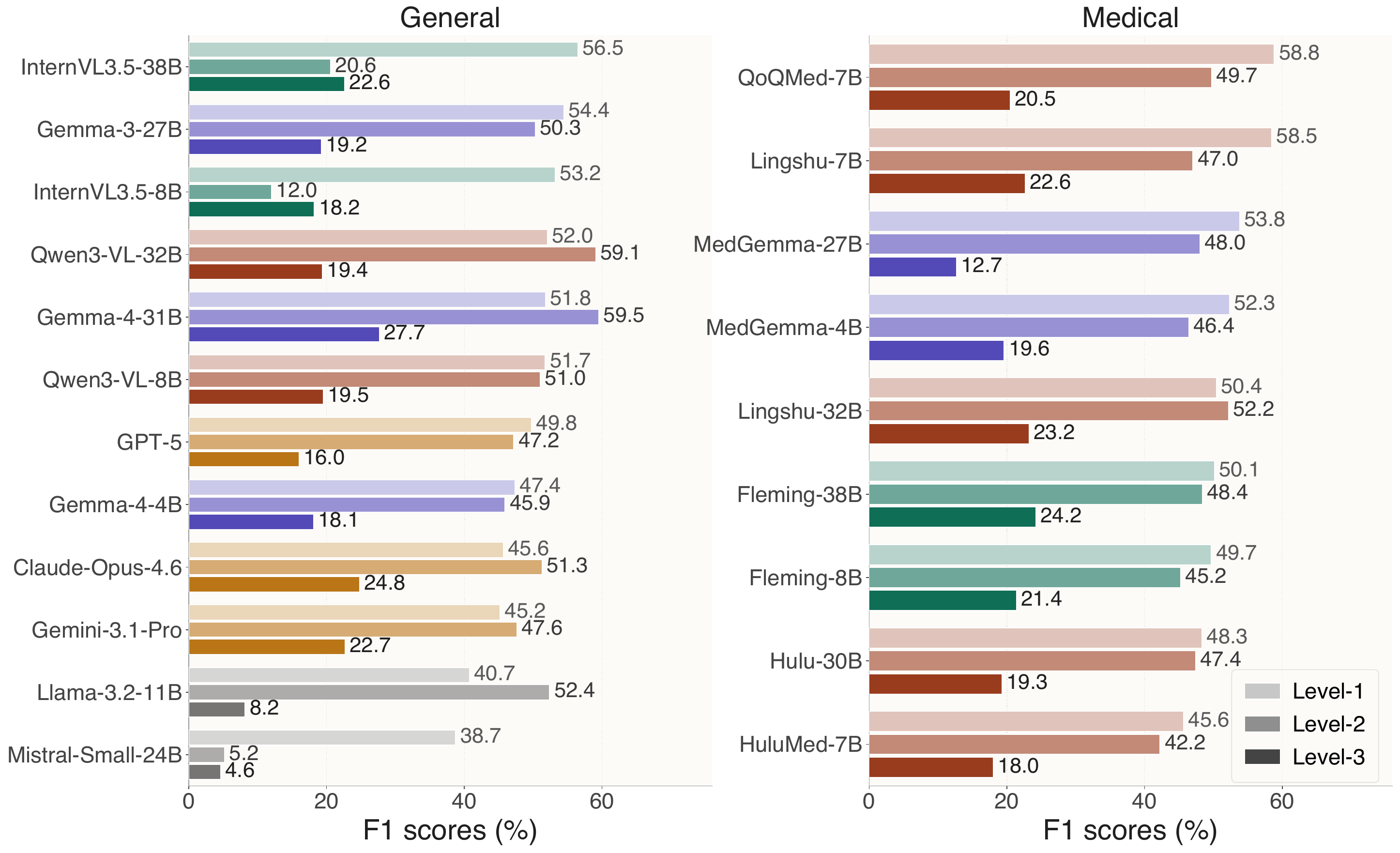}
    \end{minipage}
    \hfill
    \begin{minipage}{0.38\textwidth}
        \centering
        \includegraphics[width=\linewidth]{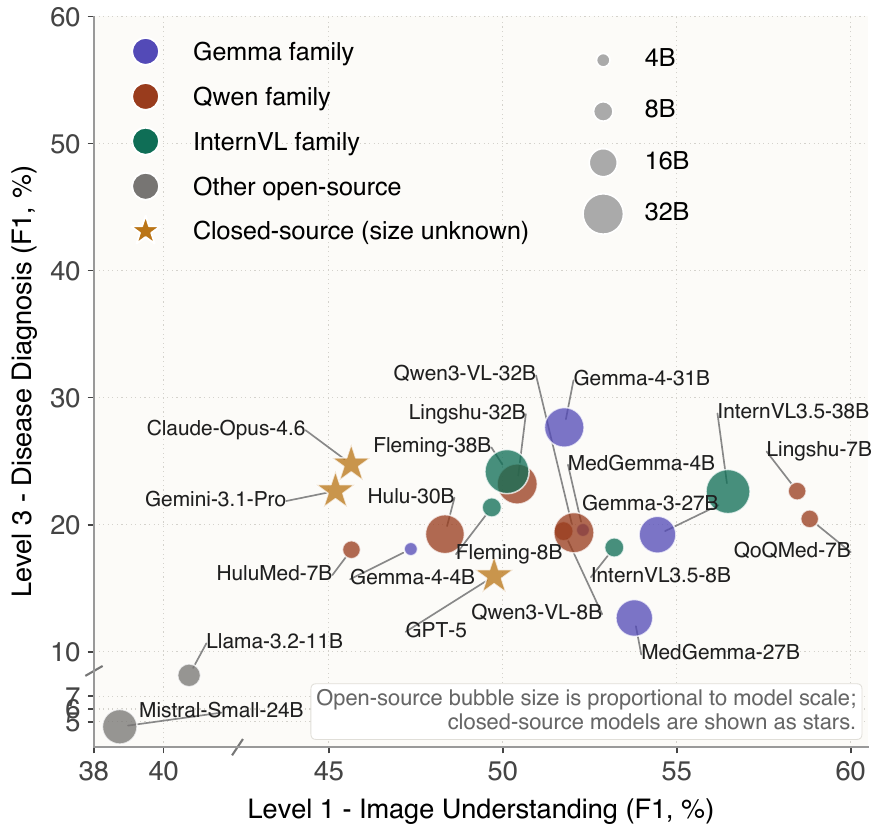}
    \end{minipage}
    \caption{\textbf{Three-tier performance landscape across evaluated MLLMs.} Values are averaged across the three slice-selection strategies.
    Performance consistently declines from image understanding to report generation and disease diagnosis, indicating that competence on isolated CMR findings does not reliably translate into higher-level clinical results. The right panel further highlights this disconnect: models with stronger Level 1 performance do not show a corresponding gain on Level 3, and most remain below 30\% for disease diagnosis.
    Open-source bubble size is proportional to parameter scale; closed-source models are shown as stars. }
    \label{fig:diag_report}
\end{figure*}

\subsection{RQ1. A Three-Layered Clinical Reality Gap}\label{sec:main_results}

We evaluate MLLMs along the full CardioLens pipeline. The overall results are summarized in Figure~\ref{fig:diag_report}. In the main text, we only report detailed image-understanding results in Table~\ref{tab:image_understanding}, while the full numerical breakdowns for report generation and disease diagnosis are provided in Appendix~\ref{app:more_detailed_results}.

\textbf{Competence Collapses Along the Clinical Pipeline.}
Figure~\ref{fig:diag_report} reveals a consistent degradation as evaluation moves from sequence-level finding recognition to clinically grounded synthesis and diagnosis. Taking QoQMed-7B as an illustrative example, the model achieves a moderate 58.8\% F1 on image understanding, but this apparent recognition ability does not reliably carry over to downstream clinical tasks. Its performance drops to 49.7\% on report generation, where multiple findings must be organized into a coherent clinical summary, and further declines to 20.5\% on disease diagnosis, where evidence must be integrated across sequences and mapped to patient-level disease categories.
This indicates that the breakdown begins before the final diagnostic stage: current MLLMs can partially recognize isolated CMR findings, but struggle to organize and integrate them into clinically consistent reports and patient-level diagnosis.
Additional oracle-report control further supports this interpretation: providing ground-truth structured clinical findings substantially improves Level-3 diagnosis across three models (Appendix~\ref{app:diagnostic_controls}).
Besides, we also observe that, within the same model family, larger models do not consistently achieve better performance: Lingshu-7B outperforms Lingshu-32B on image understanding (58.5\% vs. 50.4\% F1), MedGemma improves by less than 2\% F1 from 4B to 27B, and closed-source frontier models such as GPT-5, Gemini-3.1-Pro, and Claude-Opus-4.6 do not dominate the open-source $\leq 10$B medical models. These results suggest that parameter scaling does not prevent degradation along the clinical pipeline.

\begin{figure*}[t]
    \centering

    \begin{minipage}{0.48\textwidth}
        \centering
        \includegraphics[width=\linewidth]{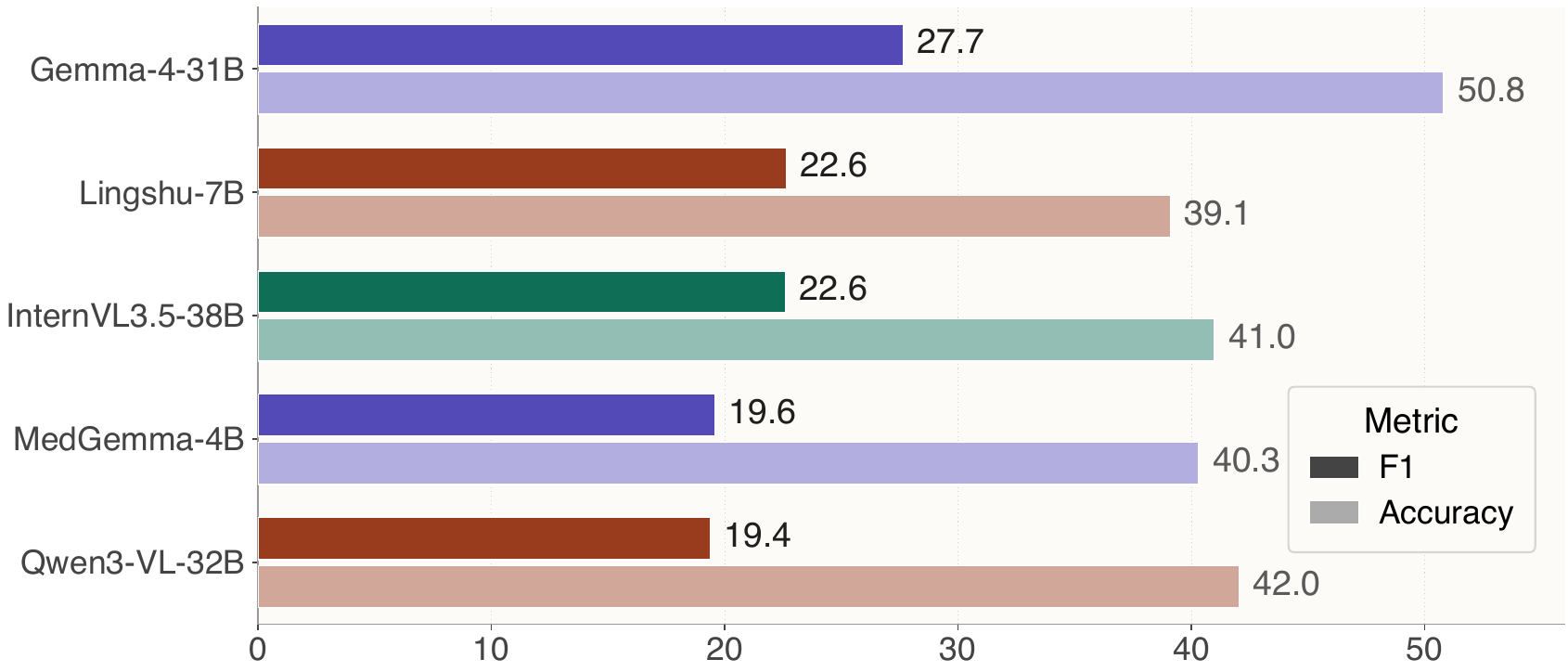}
    \end{minipage}
    \hfill
    \begin{minipage}{0.50\textwidth}
        \centering
        \includegraphics[width=\linewidth]{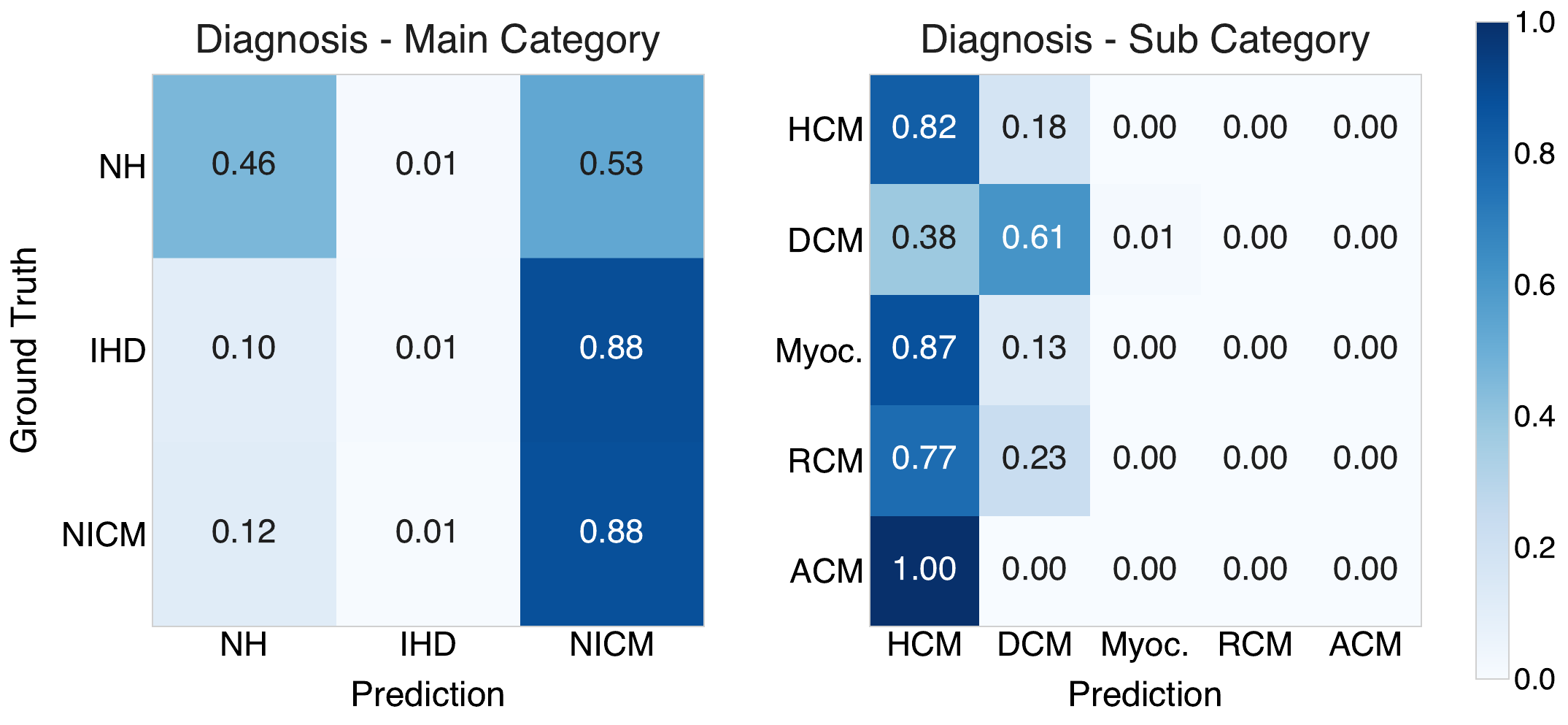}
    \end{minipage}

    \caption{\textbf{Diagnosis-stage failure analysis.}
 \textbf{Left}: Accuracy and F1 of representative open-source models on the disease diagnosis averaged across the three slice-selection strategies. \textbf{Right}: Confusion matrices for Gemma-4-31B on etiology classification and NICM subtyping, obtained from a single run under random slice selection. Predictions collapse onto the most prevalent category — NICM at the etiology level and HCM at the subtype level.}

    \label{fig:diagnosis}
\end{figure*}

\textbf{Category Collapse Drives Poor Clinical Performance.}
Aggregating metrics alone can obscure how current MLLMs fail: a model may achieve moderate accuracy while still failing to distinguish diseases in a meaningful way.
For example, as shown in the left panel of Figure~\ref{fig:diagnosis}, Gemma-4-31B achieves an F1 score of only 27.7\% on disease diagnosis, while its accuracy reaches 50.8\%, resulting in a gap of more than 20\%.
This gap suggests that the model can achieve seemingly reasonable accuracy by favoring common categories, rather than by making reliable diagnostic distinctions.
The confusion matrices in the right panel of Figure~\ref{fig:diagnosis} make this failure mode explicit.
In the etiology classification task, Gemma-4-31B predicts NICM -- the most prevalent category in the cohort -- for 88\% of IHD cases and 88\% of NICM cases, and even assigns 53\% of normal hearts to NICM. A similar collapse appears in the NICM subtyping task, where predictions are concentrated almost entirely on HCM, the most frequent NICM subtype. Specifically, 77\% of RCM cases, 87\% of myocarditis cases, and both ACM cases (2/2) are misclassified as HCM.
These patterns show that the model is not performing robust diagnosis, but instead defaults to frequent abnormal categories.
This behavior is clinically concerning: models frequently favor common disease labels even when category-relevant visual evidence is available, undermining reliable clinical decision-making.

\begin{table*}[t]
\centering
\small
\setlength{\tabcolsep}{2.8pt}
\renewcommand{\arraystretch}{1.25}
\caption{ Image understanding performance on CardioLens (Level 1). Accuracy, Precision, Recall, and F1 (\%) under three slice-selection strategies (Random / Heuristic / MIL). Values are mean±std over three independent inference runs (Heuristic is deterministic and reports a single run).}
\label{tab:image_understanding}
\newcommand{\std}[1]{\scriptsize$_{\pm #1}$}
\newcommand{\best}[1]{\textbf{#1}}
\newcommand{\second}[1]{\underline{#1}}
\resizebox{1.0\textwidth}{!}{
\begin{tabular}{llcccccccccccc}
\toprule
& \textbf{Model} &
\multicolumn{3}{c}{\textbf{Accuracy (\%)}} &
\multicolumn{3}{c}{\textbf{Precision (\%)}} &
\multicolumn{3}{c}{\textbf{Recall (\%)}} &
\multicolumn{3}{c}{\textbf{F1 Score (\%)}} \\

\cmidrule(lr){3-5}
\cmidrule(lr){6-8}
\cmidrule(lr){9-11}
\cmidrule(lr){12-14}

& & Random & Heuristic & MIL
& Random & Heuristic & MIL
& Random & Heuristic & MIL
& Random & Heuristic & MIL \\

\midrule

\multicolumn{14}{l}{\textit{Open-Source Models ($\leq$10B)}} \\
\midrule

\multirow{5}{*}{\rotatebox{90}{\scriptsize Medical}}
& MedGemma-4B
& 58.72\std{0.15} & 59.54\std{0.00} & 59.45\std{0.86}
& 41.83\std{0.05} & 43.28\std{0.00} & 43.39\std{0.70}
& \second{67.44}\std{0.16} & \second{67.48}\std{0.00} & \second{68.09}\std{0.43}
& 51.53\std{0.08} & 52.54\std{0.00} & 52.84\std{0.65} \\

& Lingshu-7B
& \best{65.22}\std{0.07} & \best{66.75}\std{0.00} & \best{67.52}\std{0.08}
& \second{50.76}\std{0.06} & \best{54.67}\std{0.00} & \best{55.08}\std{0.37}
& 63.43\std{0.05} & 65.58\std{0.00} & 65.98\std{0.17}
& \second{56.22}\std{0.04} & \best{59.34}\std{0.00} & \best{59.82}\std{0.15} \\

& Fleming-8B
& \second{64.72}\std{0.05} & \second{64.62}\std{0.00} & \second{62.96}\std{0.40}
& \best{53.18}\std{0.09} & \second{53.15}\std{0.00} & \second{51.78}\std{0.60}
& 47.93\std{0.09} & 47.71\std{0.00} & 46.55\std{0.54}
& 50.21\std{0.09} & 50.05\std{0.00} & 48.79\std{0.57} \\

& HuluMed-7B
& 50.60\std{0.25} & 51.44\std{0.00} & 52.58\std{1.35}
& 37.30\std{0.15} & 38.16\std{0.00} & 38.72\std{0.73}
& 57.16\std{0.20} & 57.87\std{0.00} & 59.22\std{1.00}
& 44.81\std{0.16} & 45.67\std{0.00} & 46.47\std{0.84} \\

& QoQMed-7B
& 58.96\std{0.11} & 59.18\std{0.00} & 59.31\std{0.67}
& 49.54\std{0.16} & 50.19\std{0.00} & 49.67\std{0.34}
& \best{74.00}\std{0.19} & \best{74.05}\std{0.00} & \best{72.92}\std{0.59}
& \best{58.72}\std{0.17} & \second{59.11}\std{0.00} & \second{58.63}\std{0.43} \\

\cmidrule(l){2-14}
\multirow{3}{*}{\rotatebox{90}{\scriptsize General}}
& Gemma-4-4B
& 49.59\std{0.26} & 55.23\std{0.00} & 57.69\std{1.60}
& 38.22\std{0.18} & 44.79\std{0.00} & 47.78\std{1.56}
& 47.25\std{0.22} & 53.41\std{0.00} & 55.61\std{1.80}
& 42.19\std{0.19} & 48.64\std{0.00} & 51.24\std{1.61} \\

& InternVL3.5-8B
& \best{60.84}\std{0.07} & \best{61.88}\std{0.00} & \best{61.95}\std{0.92}
& \best{49.27}\std{0.10} & \best{50.15}\std{0.00} & \second{51.48}\std{0.92}
& \best{56.27}\std{0.11} & \second{56.81}\std{0.00} & \second{56.53}\std{0.44}
& \best{52.51}\std{0.11} & \best{53.23}\std{0.00} & \second{53.87}\std{0.71} \\

& Qwen3-VL-8B
& \second{53.94}\std{0.18} & \second{59.02}\std{0.00} & \second{61.78}\std{1.08}
& \second{40.90}\std{0.20} & \second{48.04}\std{0.00} & \best{51.90}\std{1.37}
& \second{54.38}\std{0.23} & \best{59.13}\std{0.00} & \best{60.27}\std{0.51}
& \second{46.67}\std{0.21} & \second{52.91}\std{0.00} & \best{55.66}\std{0.69} \\

\midrule
\multicolumn{14}{l}{\textit{Open-Source Models ($>$10B)}} \\
\midrule

\multirow{4}{*}{\rotatebox{90}{\scriptsize Medical}}
& MedGemma-27B
& 63.06\std{0.09} & \second{65.24}\std{0.00} & \second{62.87}\std{0.40}
& 48.74\std{0.07} & \second{52.33}\std{0.00} & \second{50.52}\std{0.24}
& \best{58.24}\std{0.20} & \best{58.68}\std{0.00} & \best{58.84}\std{0.22}
& \best{52.69}\std{0.12} & \best{54.76}\std{0.00} & \best{53.89}\std{0.19} \\

& Hulu-30B
& 58.14\std{0.05} & 58.81\std{0.00} & 59.82\std{0.89}
& 43.18\std{0.06} & 43.74\std{0.00} & 44.53\std{0.72}
& \second{53.69}\std{0.10} & \second{53.45}\std{0.00} & \second{54.85}\std{0.63}
& 47.82\std{0.07} & 48.08\std{0.00} & 49.11\std{0.68} \\

& Lingshu-32B
& \best{65.74}\std{0.23} & \best{65.78}\std{0.00} & \best{65.45}\std{0.32}
& \best{54.63}\std{0.32} & \best{55.02}\std{0.00} & \best{54.37}\std{0.42}
& 47.28\std{0.26} & 47.70\std{0.00} & 47.17\std{0.36}
& 50.34\std{0.28} & \second{50.74}\std{0.00} & \second{50.16}\std{0.38} \\

& Fleming-38B
& \second{63.17}\std{0.17} & 62.68\std{0.00} & 61.98\std{0.42}
& \second{50.02}\std{0.15} & 49.23\std{0.00} & 48.96\std{0.77}
& 51.83\std{0.18} & 50.18\std{0.00} & 50.96\std{0.75}
& \second{50.85}\std{0.16} & 49.62\std{0.00} & 49.89\std{0.76} \\

\cmidrule(l){2-14}
\multirow{9}{*}{\rotatebox{90}{\scriptsize General}}
& Llama-3.2-11B
& 38.33\std{0.07} & 38.17\std{0.00} & 38.44\std{0.06}
& 31.42\std{0.10} & 31.01\std{0.00} & 31.20\std{0.19}
& 60.20\std{0.15} & 59.10\std{0.00} & 59.58\std{0.19}
& 41.01\std{0.11} & 40.45\std{0.00} & 40.74\std{0.06} \\

& Phi-4-14B
& 39.66\std{0.18} & 39.98\std{0.00} & 39.31\std{0.61}
& 29.09\std{0.11} & 29.19\std{0.00} & 29.03\std{0.26}
& 48.87\std{0.18} & 48.88\std{0.00} & 48.88\std{0.36}
& 36.40\std{0.14} & 36.49\std{0.00} & 36.37\std{0.30} \\

& DeepSeek-VL2-16B
& 46.49\std{0.27} & 46.98\std{0.00} & 46.58\std{0.92}
& 35.27\std{0.24} & 35.79\std{0.00} & 36.18\std{1.53}
& 56.98\std{0.37} & 58.02\std{0.00} & 59.50\std{1.57}
& 43.50\std{0.29} & 44.22\std{0.00} & 44.87\std{1.54} \\

& Mistral-Small-24B
& 39.32\std{0.35} & 39.15\std{0.00} & 39.08\std{5.16}
& 31.17\std{0.28} & 30.95\std{0.00} & 31.79\std{3.27}
& 52.38\std{0.30} & 52.57\std{0.00} & 54.14\std{1.26}
& 38.50\std{0.11} & 38.32\std{0.00} & 39.40\std{2.08} \\

& DeepSeek-VL2-27B
& 51.09\std{0.04} & 51.27\std{0.00} & 51.63\std{0.08}
& 37.88\std{0.01} & 38.22\std{0.00} & 39.03\std{0.10}
& 49.71\std{0.03} & 49.47\std{0.00} & 51.42\std{0.20}
& 42.87\std{0.01} & 42.94\std{0.00} & 44.26\std{0.10} \\

& Gemma-3-27B
& 56.03\std{0.13} & 57.10\std{0.00} & 59.59\std{0.76}
& 43.70\std{0.06} & \second{47.29}\std{0.00} & \second{48.83}\std{0.58}
& \best{64.16}\std{0.06} & \best{67.28}\std{0.00} & \best{67.34}\std{0.66}
& 51.77\std{0.06} & \second{55.21}\std{0.00} & \second{56.36}\std{0.61} \\

& Gemma-4-31B
& \second{61.83}\std{0.31} & \second{61.53}\std{0.00} & \second{62.56}\std{0.27}
& \second{46.62}\std{0.26} & 46.47\std{0.00} & 47.75\std{0.22}
& 57.86\std{0.31} & 57.26\std{0.00} & 58.12\std{0.47}
& 51.61\std{0.28} & 51.29\std{0.00} & 52.40\std{0.32} \\

& Qwen3-VL-32B
& 59.31\std{0.05} & 58.59\std{0.00} & 61.25\std{1.03}
& 45.98\std{0.09} & 45.51\std{0.00} & 48.03\std{0.64}
& \second{60.29}\std{0.11} & 58.52\std{0.00} & 60.44\std{0.66}
& \second{51.94}\std{0.10} & 50.91\std{0.00} & 53.28\std{0.63} \\

& InternVL3.5-38B
& \best{63.97}\std{0.28} & \best{64.72}\std{0.00} & \best{65.29}\std{1.30}
& \best{51.09}\std{0.18} & \best{52.67}\std{0.00} & \best{53.73}\std{0.94}
& 60.27\std{0.33} & \second{62.46}\std{0.00} & \second{60.98}\std{0.96}
& \best{55.25}\std{0.25} & \best{57.09}\std{0.00} & \best{57.08}\std{0.96} \\

\midrule
\multicolumn{14}{l}{\textit{Closed-Source Models}\textsuperscript{\textdagger}} \\
\midrule
& GPT-5
& \best{60.92}\std{0.55} & \best{62.11}{\std{0.00}} & \best{61.15}\std{0.98}
& \best{48.56}\std{1.04} & \best{49.97}{\std{0.00}} & \best{48.81}\std{0.99}
& \best{49.93}\std{1.04} & \best{52.03}{\std{0.00}} & \best{49.41}\std{1.46}
& \best{49.21}\std{1.04} & \best{50.95}{\std{0.00}} & \best{49.09}\std{1.17} \\

& Gemini-3.1-Pro
& \second{57.02}\std{0.86} & \second{58.47}{\std{0.00}} & \second{57.59}\std{0.85}
& \second{43.29}\std{0.58} & \second{44.88}{\std{0.00}} & \second{43.46}\std{1.04}
& 45.61\std{0.59} & 48.04{\std{0.00}} & 46.26\std{0.48}
& 44.41\std{0.53} & \second{46.39}{\std{0.00}} & 44.77\std{0.76} \\

& Claude-Opus-4.6
& 56.50\std{0.35} & 56.19{\std{0.00}} & 56.80\std{1.25}
& 42.15\std{0.35} & 42.52{\std{0.00}} & 42.81\std{1.27}
& \second{49.61}\std{0.31} & \second{49.08}{\std{0.00}} & \second{49.37}\std{1.61}
& \second{45.55}\std{0.33} & 45.54{\std{0.00}} & \second{45.84}\std{1.41} \\
\bottomrule
\end{tabular}
}
\scriptsize\textsuperscript{\textdagger}\textit{ Open-source: full 585-patient cohort. Closed-source: 100-patient public subset.}
\end{table*}

\textbf{Medical Knowledge Improves Abnormality Sensitivity but Not Clinical Reliability.}
We next examine whether additional medical knowledge can alleviate these clinical failures.
From Table~\ref{tab:image_understanding}, medical fine-tuning partially helps at the image-understanding level: among $\leq$ 10B open-source models, Lingshu-7B achieves 58.5\% F1, outperforming all closed-source frontier systems such as GPT-5 (49.8\% F1). However, this improvement does not translate into reliable clinical discrimination.
Several medically adapted models show large recall-over-precision gaps, including MedGemma-4B (68.1\% recall / 43.4\% precision) and QoQMed-7B (72.9\% recall / 49.7\% precision), whereas closed-source models exhibit more balanced precision--recall behavior.
This pattern suggests that
medical adaptation can increase the tendency to report possible CMR findings without ensuring that each prediction is visually supported.
We call this failure mode enumeration bias: medical-tuned models tend to over-predict plausible abnormalities without sufficient visual evidence. Therefore, additional medical knowledge can improve prediction coverage, but it does not resolve the deeper clinical failure mechanisms revealed by CardioLens.

\begin{table*}[t]
\centering
\small
\setlength{\tabcolsep}{4pt}
\renewcommand{\arraystretch}{1.2}
\caption{Effect of input size $K$ on classification performance with random selection. For each model, results are reported at $K \in \{8, 32, 128\}$. The trend column visualizes the trajectory across the three values, and the colored number gives $\Delta = K_{=128} - K_{=8}$. All values are in percentage (\%).}
\label{tab:bag_size}
\resizebox{1.0\textwidth}{!}{
\begin{tabular}{l ccc c ccc c ccc c ccc c}
\toprule
\textbf{Model} &
\multicolumn{4}{c}{\textbf{Accuracy (\%)}} &
\multicolumn{4}{c}{\textbf{Precision (\%)}} &
\multicolumn{4}{c}{\textbf{Recall (\%)}} &
\multicolumn{4}{c}{\textbf{F1 Score (\%)}} \\
\cmidrule(lr){2-5} \cmidrule(lr){6-9} \cmidrule(lr){10-13} \cmidrule(lr){14-17}
& $K{=}8$ & $K{=}32$ & $K{=}128$ & Trend
& $K{=}8$ & $K{=}32$ & $K{=}128$ & Trend
& $K{=}8$ & $K{=}32$ & $K{=}128$ & Trend
& $K{=}8$ & $K{=}32$ & $K{=}128$ & Trend \\
\midrule
MedGemma-4B
  & 58.79 & 57.79 & 56.78 & \spark{58.79}{57.79}{56.78}\,\dneg{2.00}
  & 41.82 & 40.79 & 40.85 & \spark{41.82}{40.79}{40.85}\,\dneg{0.97}
  & 67.54 & 70.28 & 70.95 & \spark{67.54}{70.28}{70.95}\,\dpos{3.42}
  & 51.55 & 51.56 & 51.80 & \spark{51.55}{51.56}{51.80}\,\dpos{0.25} \\
Lingshu-7B
  & 65.29 & 64.92 & 65.58 & \spark{65.29}{64.92}{65.58}\,\dpos{0.29}
  & 50.69 & 51.92 & 54.45 & \spark{50.69}{51.92}{54.45}\,\dpos{3.76}
  & 63.48 & 61.45 & 61.50 & \spark{63.48}{61.45}{61.50}\,\dneg{1.98}
  & 56.22 & 56.06 & 57.54 & \spark{56.22}{56.06}{57.54}\,\dpos{1.33} \\
Gemma-4-31B
  & 61.55 & 60.02 & 59.71 & \spark{61.55}{60.02}{59.71}\,\dneg{1.84}
  & 46.35 & 45.08 & 44.98 & \spark{46.35}{45.08}{44.98}\,\dneg{1.36}
  & 57.57 & 58.68 & 58.70 & \spark{57.57}{58.68}{58.70}\,\dpos{1.12}
  & 51.34 & 50.94 & 50.88 & \spark{51.34}{50.94}{50.88}\,\dneg{0.46} \\
Qwen3-VL-32B
  & 59.25 & 59.36 & 61.14 & \spark{59.25}{59.36}{61.14}\,\dpos{1.88}
  & 45.87 & 44.04 & 47.57 & \spark{45.87}{44.04}{47.57}\,\dpos{1.70}
  & 60.16 & 59.76 & 61.02 & \spark{60.16}{59.76}{61.02}\,\dpos{0.86}
  & 51.83 & 50.54 & 53.16 & \spark{51.83}{50.54}{53.16}\,\dpos{1.33} \\
InternVL3.5-38B
  & 64.29 & 65.35 & 65.48 & \spark{64.29}{65.35}{65.48}\,\dpos{1.19}
  & 51.30 & 52.14 & 52.18 & \spark{51.30}{52.14}{52.18}\,\dpos{0.88}
  & 60.65 & 61.22 & 61.77 & \spark{60.65}{61.22}{61.77}\,\dpos{1.13}
  & 55.53 & 56.26 & 56.47 & \spark{55.53}{56.26}{56.47}\,\dpos{0.94} \\
\bottomrule
\end{tabular}
}
\end{table*}

\begin{table*}[t]
\centering
\scriptsize
\setlength{\tabcolsep}{6pt}
\renewcommand{\arraystretch}{1.1}
\caption{Effect of reasoning prompts on image understanding performance with random selection. ``R'' denotes the reasoning-prompt condition; ``w/o R'' refers to the standard visual prompt, while ``w/ R'' indicates a reasoning-structured prompt that asks for an answer followed by a step-by-step justification. $\Delta = \text{w/ R} - \text{w/o R}$.}
\label{tab:reasoning_comparison_random}
\resizebox{1.0\textwidth}{!}{
\begin{tabular}{lccc|ccc|ccc|ccc}
\toprule
\multirow{2}{*}{\textbf{Model}} &
\multicolumn{3}{c|}{\textbf{Accuracy (\%)}} &
\multicolumn{3}{c|}{\textbf{Precision (\%)}} &
\multicolumn{3}{c|}{\textbf{Recall (\%)}} &
\multicolumn{3}{c}{\textbf{F1 Score (\%)}} \\
\cmidrule(lr){2-4} \cmidrule(lr){5-7} \cmidrule(lr){8-10} \cmidrule(lr){11-13}
& w/o R & w/ R & $\Delta$
& w/o R & w/ R & $\Delta$
& w/o R & w/ R & $\Delta$
& w/o R & w/ R & $\Delta$ \\
\midrule
MedGemma-4B
& 58.72 & 47.17 & \nneg{11.55}
& 41.83 & 37.76 & \nneg{4.07}
& 67.44 & 53.77 & \nneg{13.67}
& 51.53 & 44.24 & \nneg{7.29} \\

Lingshu-7B
& 65.22 & 64.95 & \nneg{0.27}
& 50.76 & 53.08 & \pos{2.32}
& 63.43 & 52.86 & \nneg{10.57}
& 56.22 & 52.95 & \nneg{3.27} \\

Gemma-4-31B
& 61.83 & 64.80 & \pos{2.97}
& 46.62 & 52.42 & \pos{5.80}
& 57.86 & 57.81 & \nneg{0.05}
& 51.61 & 54.95 & \pos{3.34} \\

Qwen3-VL-32B
& 59.31 & 49.19 & \nneg{10.12}
& 45.98 & 36.57 & \nneg{9.41}
& 60.29 & 60.26 & \nneg{0.03}
& 51.94 & 45.13 & \nneg{6.81} \\

InternVL3.5-38B
& 63.97 & 60.35 & \nneg{3.62}
& 51.09 & 46.38 & \nneg{4.71}
& 60.27 & 48.96 & \nneg{11.31}
& 55.25 & 47.53 & \nneg{7.72} \\
\bottomrule
\end{tabular}}
\end{table*}

\subsection{RQ2. Input Construction Alone Does Not Explain the Performance Gap}\label{sec:bottleneck_analysis}

A natural concern for CardioLens is whether converting high-dimensional 3D/4D CMR studies into sparse inputs unfairly biases the evaluation of clinical task performance. Since current MLLMs cannot directly process full CMR studies, their inputs must be constructed by selecting a limited number of slices from multi-view, multi-sequence, and temporally resolved imaging data. We therefore examine whether model performance is influenced by this input-construction step.

\textbf{Slice-selection Strategy.} We first compare Random, Heuristic, and MIL-based slice selection in Table~\ref{tab:image_understanding}. Across most models, the maximum F1 spread among the three strategies is $\leq 2\%$, suggesting that performance is relatively insensitive to how slices are selected. Smaller models, such as Gemma-4-4B and Qwen3-VL-8B, benefit more from informed selection, with gains up to 9\% F1, indicating that sparse input construction has a larger impact on models with limited capacity. However, these improvements remain insufficient to close the overall clinical performance gap.

\textbf{Slice Budget.} We next test whether providing more visual evidence alleviates the clinical gap. Table~\ref{tab:bag_size} ablates $K \in \{8, 32, 128\}$ under random selection across five representative models. Increasing $K$ from 8 to 128 changes F1 by at most 1.3\% for Lingshu-7B and Qwen3-VL-32B, while other models show flat or even slightly negative trends. Accuracy and recall often trade off rather than improve jointly. Thus, a $16\times$ expansion of the visual context does not produce meaningful performance gains.

These results show that the large clinical performance gap persists across the tested slice-selection strategies and budgets. Providing more slices or more targeted selections alone does not resolve it.

\subsection{RQ3. Reasoning Prompts Provide No Consistent Gains}\label{sec:reasoning_effect}

We examine whether explicit clinical reasoning instructions can improve MLLM performance on CMR tasks. Table~\ref{tab:reasoning_comparison_random} compares the standard visual prompt with a reasoning-structured prompt, using identical visual inputs but requiring a clinical justification after the answer (Appendix~\ref{app:prompts}). Overall, reasoning prompts do not produce consistent gains, with F1 decreasing for four of the five models. In particular, recall falls from 67.44\% to 53.77\% for MedGemma-4B, from 63.43\% to 52.86\% for Lingshu-7B, and from 60.27\% to 48.96\% for InternVL3.5-38B. At the same time, precision improves for some models, such as Lingshu-7B and Gemma-4-31B.
This pattern further exposes the instability of current MLLMs: their clinical predictions are not reliably grounded in CMR visual evidence. The recurring recall losses are consistent with a shift toward more conservative predictions rather than improved identification of abnormal findings.
A detailed case study is provided in Appendix~\ref{app:case_reasoning}.
Additional analyses of the clinical framework and output order are presented in Appendix~\ref{app:prompt_order_controls}, with neither consistently improving performance.

\section{Conclusion}
We introduced CardioLens, a leakage-resistant multi-sequence CMR testbed for evaluating Multimodal Large Language Models (MLLMs) beyond conventional medical VQA. By covering image understanding, report generation, and disease diagnosis, CardioLens shows that success on isolated, recognition-style benchmarks does not justify clinical readiness. Current MLLMs exhibit a clear clinical reality gap: they struggle to interpret distributed CMR findings, generate clinically faithful sequence-level reports, and make patient-level diagnoses from multi-sequence images. These failures reveal a structural mismatch between existing evaluation practice and clinically deployable interpretation. By aligning evaluation with real-world medical workflow, CardioLens provides a controlled testbed for redefining the clinical capability boundary of MLLMs.

\begin{ack}
This study was supported by grants from the National Key R\&D Program of China (2022YFE0209800), the National Natural Science Foundation of China (82271986, 82402256), and the Youth Fund of the National Natural Science Foundation of China (82603787).
\end{ack}

\bibliographystyle{plainnat}

\bibliography{reference}

\clearpage
\appendix

\section{Task and Sequence Specifications in CardioLens}
\label{app:cardiolens_task-details}

This appendix provides the complete per-subtask and per-sequence specifications deferred from Section~\ref{sec:task_definition}. Section~\ref{app:sequence-details} details the clinical role of each CMR sequence. Section~\ref{app:image-understanding-details} enumerates the 32 image understanding subtasks with full class definitions and clinical rationale. Section~\ref{app:diagnosis-subtasks} describes the disease diagnosis tasks and their clinically grounded multi-sequence phenotypes.

\subsection{Per-Sequence Clinical Roles}
\label{app:sequence-details}

\paragraph{Cine sequence.} Cine MRI acquires 4D data (three spatial dimensions plus time), capturing complete cardiac motion throughout the cardiac cycle. It serves as the foundational sequence for assessing ventricular wall motion, systolic and diastolic function, and valvular activity. Each patient study typically includes a short-axis (SAX) stack providing cross-sectional coverage from base to apex, together with long-axis views -- principally 2 chamber(CH), 3CH, and 4CH orientations -- each capturing complementary anatomical information. The SAX stack is the primary substrate for segmental wall thickness and motion assessment, while 3CH and 4CH long-axis views are specifically required for evaluating valvular activity, left ventricular outflow tract (LVOT) pathology, and whole-chamber dynamics. Because a single Cine view contributes tens to hundreds of slices (spatial $\times$ temporal), Cine is both the richest and the most view-sensitive sequence in the benchmark.

\paragraph{LGE sequence.} Late gadolinium enhancement imaging reveals the distribution of myocardial fibrosis and scar tissue through delayed contrast accumulation, serving as the gold standard for differentiating ischemic from non-ischemic cardiomyopathy. Each LGE study provides a SAX stack for transmural and circumferential characterization of enhancement patterns, together with long-axis views -- principally 4CH and 2CH. Different LGE sub-findings (e.g., layer vs.\ segment vs.\ circumferential region) are best assessed from different views.

\paragraph{Perfusion sequence.} First-pass perfusion imaging dynamically captures contrast agent transit through the myocardium, enabling assessment of myocardial microvascular perfusion and serving as a critical tool for functional evaluation of ischemic heart disease. Unlike Cine and LGE, perfusion studies typically consist of multiple 2D+t acquisitions, commonly including short-axis (SAX) and four-chamber (4CH) views rather than a single static stack. The large slice count relative to file count reflects the combination of multiple view acquisitions and the dense temporal sampling inherent to first-pass imaging.

\paragraph{T2-weighted sequence.} T2-weighted imaging is highly sensitive to tissue water content, primarily used to detect myocardial edema and acute inflammatory responses. It provides irreplaceable diagnostic value in the early detection of myocarditis and acute myocardial infarction. In CardioLens, T2 is acquired with fat suppression (T2-STIR) to suppress epicardial fat signal and enhance edema conspicuity, predominantly as a SAX stack, with 4CH available in a subset of patients.

\subsection{Image Understanding Subtasks}
\label{app:image-understanding-details}

We define 32 image understanding subtasks by structuring recurrent, clinically meaningful observations commonly described in routine radiology reports into standardized prediction targets. Each task is formulated as either single-label (exactly one category applies) or multi-label (multiple findings may co-exist) classification. The multi-label formulation matches the clinical reality that diagnostic observations at a given anatomical location can be multifaceted (e.g., different myocardial segments may simultaneously exhibit distinct abnormalities).

\subsubsection{Cine Sequence (11 subtasks)}

The Cine sequence is central to cardiac functional assessment. These tasks require the model to extract comprehensive information about myocardial motion, ventricular function, and valvular status from dynamic 4D sequences, directly challenging the model's temporal understanding capability -- since a single static frame is insufficient for complete functional evaluation.

\begin{itemize}[leftmargin=*]
\item \textbf{Myocardial wall thickness}: Multi-label, 5 classes -- normal, thinning, heterogeneous thickness, thickening, and bulging. Thinning is commonly observed in dilated cardiomyopathy or myocardial atrophy. Thickening may suggest hypertrophic cardiomyopathy (HCM) or hypertensive heart disease; aneurysmal bulging indicates ventricular aneurysm formation. Different segments may simultaneously exhibit distinct thickness abnormalities.

\item \textbf{Wall motion coordination}: Single-label, 2 classes -- coordinated and uncoordinated. Determines whether wall segments contract synchronously during systole. Uncoordination may be observed in conduction abnormalities, cardiomyopathy, or post-infarction remodeling, requiring the model to understand temporal synchronization relationships across multiple segments.

\item {\textbf{Wall motion amplitude}: Multi-label, 5 classes -- normal, reduced, absent, paradoxical motion, and enhanced. This task assesses regional LV contraction amplitude on cine MRI, ranging from preserved motion to reduced or absent motion, paradoxical systolic displacement, and enhanced or compensatory contraction.}

\item \textbf{Left ventricular systolic function}: Single-label, 3 classes -- normal, reduced, and enhanced. Reduced systolic function is the cardinal manifestation of heart failure; enhanced systolic function may be observed in early-stage HCM or hyperdynamic circulatory states.

\item \textbf{Diastolic function}: Single-label, 2 classes -- normal and restricted. Restricted diastolic function indicates reduced ventricular compliance, commonly seen in restrictive cardiomyopathy, myocardial fibrosis, and constrictive pericarditis.

\item \textbf{Mitral valve regurgitation}: Single-label, 2 classes -- regurgitation, no regurgitation. Mitral regurgitation may result from ventricular dilation or primary valvular disease.

\item \textbf{Tricuspid valve regurgitation}: Single-label, 2 classes -- regurgitation, no regurgitation. Reflects right ventricular function and pulmonary arterial pressure.

\item \textbf{Aortic valve regurgitation }: Single-label, 2 classes -- regurgitation, no regurgitation. Critical for diagnosing aortic valvular disease and aortic root pathology.

\item \textbf{Special signs}: Multi-label, 6 classes -- SAM sign positive, LVOT obstruction, ``ace of spades'' sign, aortic stenosis, pulmonary artery dilation, and no special signs present. These findings carry important diagnostic value: SAM sign and LVOT obstruction are commonly associated with obstructive HCM; the ``ace of spades'' sign is classically associated with apical HCM; aortic stenosis identification directly impacts valve intervention planning and clinical management.

\item \textbf{Pericardial effusion}: Single-label, 2 classes -- present, absent. Routinely assessed in CMR examinations; can accompany a wide range of disease states.

\item \textbf{Pleural effusion}: Single-label, 2 classes -- present, absent. Often indicates heart failure or systemic disease; important incidental finding.
\end{itemize}

\subsubsection{LGE Sequence (11 subtasks)}

LGE is the gold standard for myocardial tissue characterization. The interpretive logic is inherently hierarchical: the radiologist first determines whether abnormal enhancement is present; if so, subsequent analysis characterizes signal nature, affected segments and regions, morphological distribution pattern, and involved myocardial layers. Our subtask design directly mirrors this structured workflow.

\begin{itemize}[leftmargin=*]
\item \textbf{Enhancement status}: Single-label, 2 classes -- no abnormal enhancement, abnormal enhancement present. The primary gating question for all downstream LGE subtasks; abnormal enhancement signifies myocardial fibrosis, necrosis, or infiltrative disease.

\item \textbf{Abnormal signal characteristics}: Single-label, 2 classes -- High, mixed. Pure high signal typically reflects relatively homogeneous fibrosis or scar. Mixed signal refers to hyperenhancement containing internal low-signal components, which may indicate heterogeneous tissue injury such as microvascular obstruction and can be associated with more severe myocardial damage.

\item \textbf{High/low signal segments}: Multi-label. High-signal segments have 5 classes -- basal, mid-ventricular, apical, apex, and papillary muscle; low-signal segments retain 4 classes -- basal, mid-ventricular, apical, apex. These tasks describe the longitudinal extent of enhancement, with papillary muscle involvement additionally included for high signal.

\item \textbf{High/low signal regions}: Multi-label, 6 classes -- anterior, anteroseptal, inferoseptal, inferior, inferolateral, anterolateral walls. Circumferential localization; correspondence with coronary artery perfusion territories distinguishes ischemic from non-ischemic etiologies.

\item \textbf{High/low signal pattern}: Multi-label, 5 classes -- diffuse, linear, patchy, transmural, speckled. Transmural enhancement suggests coronary-related ischemic injury; mid-wall linear enhancement characterizes Dilated Cardiomyopathy (DCM); subepicardial speckled enhancement is typical of myocarditis.

\item \textbf{High/low signal layer}: Multi-label, 5 classes -- subendocardial, mid-myocardial, {subepicardial}, transmural, papillary muscle. Layer involvement is among the most important discriminators for differential diagnosis: subendocardial-to-transmural enhancement in a coronary perfusion territory is the hallmark of ischemic injury; mid-myocardial enhancement suggests non-ischemic etiologies, classically DCM, sarcoidosis, or chronic myocarditis.

\item \textbf{Special findings}: Multi-label, 4 classes -- No special findings, pericardial abnormality, hyperintensity at interventricular septal insertion points, right ventricular hyperintensity. Septal insertion point hyperintensity indirectly signals pulmonary hypertension or right ventricular pressure overload; right ventricular hyperintensity may indicate arrhythmogenic cardiomyopathy (ACM) and HCM.
\end{itemize}

\subsubsection{T2-Weighted Sequence (5 subtasks)}

The clinical value of T2-weighted imaging lies in detecting myocardial edema and acute inflammation. The interpretive workflow parallels LGE: initial assessment of signal characteristics, followed by spatial localization and distribution pattern characterization.

\begin{itemize}[leftmargin=*]

\item \textbf{T2 signal status}: Single-label, 4 classes -- normal, high, low, and mixed.
High T2 signal generally indicates increased myocardial water content and is commonly associated with edema or acute inflammatory injury, whereas low T2 signal may reflect shortened relaxation effects related to iron deposition, hemorrhagic products, or chronic residual changes. Mixed T2 signal denotes a heterogeneous pattern in which high- and low-signal components coexist within the myocardium.

\item \textbf{Abnormal signal segments}: Multi-label, 4 classes -- basal, mid-ventricular, apical, apex. Longitudinal distribution of T2 abnormalities.

\item \textbf{Abnormal signal regions}: Multi-label, 6 classes (same as LGE circumferential scheme). Circumferential localization of T2 abnormalities.

\item \textbf{Signal distribution pattern}: Multi-label, 5 classes (same as LGE pattern scheme). Diffuse edema suggests systemic involvement or widespread viral myocarditis; segmental distribution indicates acute injury within a specific coronary territory.

\item \textbf{Affected myocardial layer}: Multi-label, 5 classes (same as LGE layer scheme). Layer distribution of T2 abnormalities carries etiological significance analogous to LGE.
\end{itemize}

\subsubsection{Perfusion Sequence (5 subtasks)}

Perfusion imaging assesses myocardial microvascular functional status. Interpretation centers on the presence, spatial distribution, and severity of perfusion abnormalities.

\begin{itemize}[leftmargin=*]
\item \textbf{Perfusion status}: Single-label, 2 classes -- normal, abnormal. Gating criterion for subsequent detailed analysis.

\item \textbf{Perfusion signal characteristics}: Single-label, 3 classes -- reduced perfusion, delayed perfusion, and perfusion defect.
This task characterizes abnormal first-pass myocardial contrast dynamics. Reduced perfusion denotes decreased myocardial enhancement, delayed perfusion captures a temporal delay in contrast wash-in, and perfusion defect refers to a focal region of absent or markedly reduced contrast uptake. These patterns may reflect impaired myocardial blood flow, reduced perfusion reserve, or microvascular dysfunction.

\item \textbf{Abnormal perfusion segments}: Multi-label, 4 classes -- basal, mid-ventricular, apical, apex.

\item \textbf{Abnormal perfusion regions}: Multi-label, 6 classes (same circumferential scheme as LGE/T2). The correspondence between perfusion abnormality regions and coronary perfusion territories is a cornerstone of coronary artery disease diagnosis.

\item \textbf{Affected myocardial layer}: Multi-label, 5 classes (same layer scheme as LGE/T2). Subendocardial perfusion abnormality is an early and sensitive marker of ischemia; transmural defect indicates severe supply-demand mismatch.
\end{itemize}

\subsection{Diagnosis Subtasks}
\label{app:diagnosis-subtasks}

Unlike image-understanding tasks that focus on a single sequence, Level-3 diagnosis maps multi-sequence CMR images directly to patient-level disease categories. As an end-to-end task, it involves recognizing and organizing clinical findings and using complementary evidence across sequences to reach a diagnosis.
The ground-truth labels are aligned with clinical discharge diagnoses rather than purely imaging-derived interpretations, reflecting real-world clinical decision-making.

Etiology classification  is a three-class classification task (Normal Heart (NH), Ischemic Heart Disease (IHD), Non-Ischemic Cardiomyopathy (NICM)), corresponding to the most fundamental differential diagnostic framework in CMR practice. It captures the primary separation between the absence of structural disease, ischemic etiologies, and non-ischemic myocardial disorders.

NICM subtyping task further subcategorizes non-ischemic cardiomyopathy (NICM) into five phenotypes: Hypertrophic Cardiomyopathy (HCM), Dilated Cardiomyopathy (DCM), Restrictive Cardiomyopathy (RCM), Arrhythmogenic Cardiomyopathy (ACM), and Myocarditis. Among these, RCM and ACM are relatively uncommon, with only 13 and 2 cases, respectively, in our cohort. Each subtype is characterized by a distinct multi-sequence imaging phenotype, including patterns of myocardial hypertrophy or dilation, wall motion abnormalities, T2 signal abnormalities, and LGE distribution patterns.

\begin{itemize}[leftmargin=*]

\item \textbf{Hypertrophic Cardiomyopathy (HCM)}: asymmetric or focal myocardial hypertrophy, usually preserved systolic function, and non-ischemic LGE, often involving hypertrophied segments or right ventricular insertion points. Additional phenotype-specific signs may include systolic anterior motion (SAM) in obstructive HCM and the ``ace-of-spades'' configuration in apical HCM.

\item \textbf{Dilated Cardiomyopathy (DCM)}: left ventricular or biventricular chamber dilation, global wall motion reduction, reduced systolic function, and non-ischemic mid-wall linear LGE, particularly in the interventricular septum.

\item \textbf{Restrictive Cardiomyopathy (RCM)}: restrictive ventricular filling or impaired diastolic function, typically accompanied by biatrial enlargement, with normal or near-normal ventricular cavity size and relatively preserved systolic function.

\item \textbf{Arrhythmogenic Cardiomyopathy (ACM)}: right ventricular or biventricular dilation, regional wall motion abnormalities, and myocardial fibrofatty replacement or fibrosis, often affecting the right ventricle but potentially involving the left ventricle.

\item \textbf{Myocarditis}: myocardial edema reflected by T2 hyperintensity or elevated T2-based markers, often accompanied by non-ischemic injury patterns such as subepicardial or mid-myocardial LGE.

\end{itemize}

\clearpage

\section{Heuristic Slice Selection Details}
\label{app:heuristic}

\begin{table}[htbp]
\centering
\small
\caption{Task-to-view assignments under the heuristic protocol. Each subtask is routed to the view conventionally used in clinical reading practice.}
\label{tab:task_to_view}
\setlength{\tabcolsep}{4pt}
\renewcommand{\arraystretch}{1.12}

\begin{tabular}{llp{8.2cm}}
\toprule
Sequence & View & Subtasks \\
\midrule

\multirow{3}{*}{Cine}
& SAX
& \texttt{thickening} \\

& 4CH
& \texttt{motion\_coordination}, \texttt{motion\_amplitude},
\texttt{systolic\_function}, \texttt{diastolic\_function},
\texttt{pericardial\_effusion}, \texttt{pleural\_effusion},
\texttt{mitral\_valve}, \texttt{tricuspid\_valve} \\

& 3CH
& \texttt{aortic\_valve}, \texttt{special\_signs} \\

\midrule

\multirow{2}{*}{LGE}
& 4CH
& \texttt{high\_seg}, \texttt{low\_seg}, \texttt{special\_findings} \\

& SAX
& \texttt{enhancement\_status}, \texttt{abnormal\_signal},
\texttt{high/low\_region}, \texttt{high/low\_pattern},
\texttt{high/low\_layer} \\

\midrule

T2
& SAX
& all five T2 subtasks \\

Perfusion
& Dynamic
& all five perfusion subtasks \\

\bottomrule
\end{tabular}
\end{table}

This appendix details the implementation of the heuristic slice selection
protocol described in Section~\ref{sec:slice_selection}. We first describe the
overall extraction framework (Section~\ref{app:heuristic:framework}), then give
per-sequence allocation rules (Section~\ref{app:heuristic:cine}--\ref{app:heuristic:perfusion}),
and finally describe how the protocol adapts to small and large slice
budgets (Section~\ref{app:heuristic_adapting_to_budget}).

\subsection{Overall Framework}
\label{app:heuristic:framework}

Given a patient case, a target clinical subtask, and a per-sequence slice budget $K$,
the heuristic protocol produces an ordered list of $K$ 2D grayscale PNG slices
that will be fed to the MLLM as the visual context for that question. The
extraction proceeds in two steps:
\begin{enumerate}[leftmargin=*, itemsep=1pt]
    \item \textbf{View routing.} The subtask is mapped to one or more
    canonical views (sequence, orientation) according to
    Table~\ref{tab:task_to_view}. This mapping was elicited from practicing
    cardiac radiologists and reflects how each subtask is read in clinical
    practice (e.g., wall thickening from Cine SAX, valvular motion from
    Cine 4CH, scar localization from LGE SAX).
    \item \textbf{Per-view slice extraction.} For each routed view, we apply a
    view-specific extraction rule that exploits the known acquisition
    structure of that sequence (spatial stacks, cardiac-cycle frames, or
    2D+t dynamic series). The rules are detailed per sequence below.

\end{enumerate}

\paragraph{Storage.} The spatial and temporal organization of the acquisition images is retained in the stored NIfTI arrays. Individual acquisitions are stored as 3D NIfTI volumes
of shape $(H, W, Z)$. For Cine sequences, the $Z$ axis interleaves $N_s$
spatial slices with $T$ cardiac-phase frames (so $Z = N_s \cdot T$ with
typically $T \in \{25, 30\}$); frames $[kT, (k{+}1)T{-}1]$ correspond to
the $k$-th spatial slice imaged over one full cardiac cycle. For LGE and
T2, $Z$ is purely spatial. For Perfusion, the stored images are indexed over the first-pass dynamics, with spatial-plane coverage varying across acquisitions.

\subsection{Cine Sequence}
\label{app:heuristic:cine}

Cine acquisitions are the richest sequence in the benchmark because they
carry both spatial and temporal structure: the same anatomy is imaged
across an entire cardiac cycle. The extraction rules differ by view
as different views are used for fundamentally different
clinical questions.

\paragraph{Cine SAX.} We recover the spatial structure by detecting the temporal period from the total frame count $Z$: if $Z \bmod 25 = 0$ we use $T = 25$; else if $Z \bmod 30 = 0$ we use $T = 30$. This yields $N_s = Z / T$ spatial slices. From each spatial slice, only the first temporal frame (end-diastolic phase) is retained, and $K_{\mathrm{sax}}$ slices are uniformly subsampled from these $N_s$ candidates. This ensures consistent anatomical coverage from base to apex at end-diastole, where myocardial wall thickness is most reliably measured due to maximal ventricular filling, following standard clinical practice for structural assessment.
\paragraph{Cine 4CH and 3CH.} Long-axis Cine acquisitions typically contain
\emph{multiple spatial slices} (commonly 3
slices for 2CH and 4CH views while 1 for 3CH view), each imaged over one full cardiac cycle of
$T \in \{25, 30\}$ frames. To obtain a clinically representative view, we
extract the \emph{central} spatial slice -- conventionally the anatomically
best-aligned plane through the mitral/aortic annulus -- and uniformly
sample $K_{\mathrm{long}}$ frames from its full cardiac cycle.
This yields $K_{\mathrm{long}}$ frames spanning a complete systole--diastole
cycle through the most diagnostically representative long-axis plane.

\subsection{LGE Sequence}
\label{app:heuristic:lge}

LGE volumes are purely 3D (without an explicit temporal axis), which simplifies extraction compared to Cine but introduces a different design consideration. The SAX stack provides contiguous ventricular coverage and serves as the main source for regional assessment, while long-axis views (4CH and 2CH, when available) offer complementary perspectives for anatomical localization. Our allocation is guided by coverage and complementarity.

\paragraph{LGE SAX.} For subtasks routed to SAX, we uniformly sample
$K$ slices along the spatial dimension, yielding anatomical coverage from
base to apex.

\paragraph{LGE 4CH.} For subtasks routed to 4CH, all available 4CH slices
are used when the count is $\leq K_{\mathrm{4ch}}$; otherwise
$K_{\mathrm{4ch}}$ slices are uniformly subsampled. The 4CH budget is
capped rather than scaled with $K$ because 4CH is used primarily for
global orientation, with detailed regional assessment deferred to SAX.

\subsection{T2-Weighted Sequence}
\label{app:heuristic:t2}

For T2 we sample from the SAX acquisition when available, as it provides
comprehensive spatial coverage for detecting regional edema. $K$ slices
are uniformly extracted along the spatial dimension. When T2 SAX is
unavailable for a patient, we fall back to T2 4CH rather than dropping
the case.

\subsection{Perfusion Sequence}
\label{app:heuristic:perfusion}

Each stored Perfusion acquisition consists of one or more spatial planes, each represented as a 2D+t dynamic image sequence. $K$ slices are uniformly extracted along the temporal dimension,
capturing the first-pass contrast dynamics from baseline through peak
myocardial enhancement to washout.

\subsection{Adapting to the Slice Budget $K$}
\label{app:heuristic_adapting_to_budget}

The rules above describe \emph{which candidate slices are considered} for
each view; the total budget $K$ determines how these candidates are
allocated.

\paragraph{Small-$K$ regime ($K = 8$).}
At $K=8$, the protocol uses fixed view allocations determined by the clinical task. For Level-1 questions, the budget is assigned to the view specified in Table~\ref{tab:task_to_view}. For example, motion questions use eight frames uniformly sampled across the cardiac cycle from the central 4CH slice.

\paragraph{Large-$K$ regime ($K = 32/128$).}
At larger budgets, fixed per-view splits become wasteful: a view's
candidate pool can be exhausted (e.g., $K = 128$ exceeds a 25-frame cardiac cycle) while other views still have untapped diagnostic content.
We therefore use a \emph{cascading multi-phase allocation} that fills
slices in order of diagnostic priority and only moves to the next source
when the current one is exhausted. For Cine SAX, the cascade is:
\begin{enumerate}[leftmargin=*, itemsep=1pt]
    \item \textbf{Phase 1 (spatial coverage):} the end-diastolic frame
    from every spatial slice ($N_s$ frames, matching the small-$K$ rule).
    \item \textbf{Phase 2 (key-layer dynamics):} additional temporal
    frames from three clinically key layers -- basal,
    mid, and apical -- sampled uniformly per layer. These layers are chosen to match
    the conventional three-slice reporting convention (basal, mid, apical
    short-axis) used in both clinical reads and AHA 17-segment analysis.
    \item \textbf{Phase 3 (remaining-layer dynamics):} if the budget is
    still not filled, temporal frames from the remaining spatial layers,
    uniformly distributed across layers and cardiac phases.
    \item \textbf{Phase 4 (cross-view fallback):} if all Cine SAX content
    is exhausted and the budget is still not filled, we draw additional
    frames from Cine 4CH, starting with the central spatial layer's
    cardiac cycle and expanding outward.
\end{enumerate}
Cine 3CH and Cine 4CH use analogous cascades, falling back to each other
as the final phase. LGE switches from the fixed small-$K$ split to an \emph{all-long-axis-first} policy: 4CH and 2CH (when available) are taken in full, and SAX fills the remainder uniformly. In practice, for $K \geq 32$, the entire LGE volume is typically covered (SAX $\approx 8$--$12$ slices; 4CH/2CH $\approx 3$ each), so the allocation becomes effectively exhaustive rather than selective. T2 and Perfusion retain their sequence-specific spatial and temporal sampling rules.

\paragraph{View unavailability.} Across all budgets, if a required view
is entirely missing for a patient, the corresponding budget is redirected
to the next view in the cascade (e.g., missing 3CH $\to$ extra 4CH;
missing T2 SAX $\to$ T2 4CH).

\clearpage

\section{Data-driven MIL Framework}
\label{app:mil}

This section provides a complete description of the Data-driven Multiple Instance Learning (MIL) framework introduced in Section~\ref{sec:slice_selection}, which serves as the backbone of our data-driven slice selection protocol. We detail the model architecture (Section~\ref{app:mil:arch}), training objective (Section~\ref{app:mil:obj}), implementation specifics (Section~\ref{app:mil:impl}), and empirical validation of the learned attention (Section~\ref{app:mil:val}). The purpose of this framework is \emph{not} to serve as a diagnostic model in itself, but to learn view selection under multi-task supervision, thereby providing a data-driven comparison against which MLLM performance can be benchmarked.

\subsection{Architecture}
\label{app:mil:arch}

\subsubsection{Problem Formulation}

Let $\mathcal{S} = \{s_1, \ldots, s_N\}$ denote a bag of $N$ 2D slices from a single cardiac MRI sequence of a given patient, where each $s_i \in \mathbb{R}^{H \times W}$ is a single-channel image. Associated with the sequence is a set of clinical tasks $\mathcal{T} = \{t_1, \ldots, t_M\}$, each with a label $y_j$ and a validity mask $m_j \in \{0, 1\}$ indicating whether the label is available for the current patient (labels may be partially missing due to clinical reporting conventions). The objective is to jointly (i) learn per-slice, per-task importance weights $w_{j,i} \in [0, 1]$ and (ii) predict $\hat{y}_j$ for all valid tasks, using only bag-level supervision.

\subsubsection{Shared Slice Encoder}

All slices, across all four sequences (Cine, LGE, T2, Perfusion), are encoded by a single shared backbone $f_\theta$, instantiated as a ResNet-18~\cite{he2016deep} pretrained on ImageNet. The first convolutional layer is replaced with a $7 \times 7$ kernel accepting single-channel input. Each slice is encoded independently:
\begin{equation}
\mathbf{h}_i = f_\theta(s_i) \in \mathbb{R}^D, \quad i = 1, \ldots, N,
\end{equation}
with $D = 512$. The per-bag feature matrix is $\mathbf{H} = [\mathbf{h}_1, \ldots, \mathbf{h}_N]^\top \in \mathbb{R}^{N \times D}$.

\subsubsection{Learnable Positional Encoding}

Because cardiac MRI slices carry intrinsic spatial (base$\rightarrow$apex) and temporal (cardiac phase) ordering that is informative for diagnosis, we inject ordinal information via a learnable positional encoding $\mathbf{P} \in \mathbb{R}^{N_{\max} \times D}$ (with $N_{\max} = 1000$):
\begin{equation}
\tilde{\mathbf{H}} = \mathbf{H} + \mathbf{P}_{1:N}.
\end{equation}
$\mathbf{P}$ is initialized from $\mathcal{N}(0, 1)$ and trained jointly with the rest of the model.

\subsubsection{Task-aware Transformer Attention}

For each clinical task $t_j$, we instantiate an independent attention module $\mathcal{A}_j$ comprising three stages.

\paragraph{Stage 1  --  Self-Attention Among Slices.} A multi-head self-attention (MHSA) layer contextualizes each slice feature with respect to the global bag:
\begin{equation}
\mathbf{H}^{\text{sa}} = \text{LayerNorm}\!\left(\tilde{\mathbf{H}} + \text{Dropout}\!\left(\text{MHSA}(\tilde{\mathbf{H}}, \tilde{\mathbf{H}}, \tilde{\mathbf{H}})\right)\right),
\end{equation}
with $n_h = 4$ heads and dropout $p = 0.1$. This step captures inter-slice dependencies essential for cardiac prediction (e.g., ``a slice shows basal hypertrophy \emph{relative to} the mid-ventricular slices'').

\paragraph{Stage 2  --  Task Query Cross-Attention.} Each task maintains an independent learnable query vector $\mathbf{q}_j \in \mathbb{R}^{1 \times D}$, initialized from $\mathcal{N}(0, 1)$. The query attends over the contextualized bag:
\begin{equation}
\mathbf{c}_j = \text{CrossAttn}(\mathbf{q}_j, \mathbf{H}^{\text{sa}}, \mathbf{H}^{\text{sa}}),
\end{equation}
followed by a residual feed-forward block:
\begin{equation}
\mathbf{r}_j = \text{LayerNorm}\!\left(\mathbf{c}_j + \text{Dropout}(\text{FFN}(\mathbf{c}_j))\right),
\end{equation}
where $\text{FFN}(x) = W_2 \, \text{ReLU}(W_1 x)$ with hidden dimension $2D$. The resulting $\mathbf{r}_j$ encodes the task-conditioned global summary.

\paragraph{Stage 3  --  Slice Importance Scoring.} The task representation is broadcast and additively fused with the contextualized slice features to form task-enhanced per-slice representations:
\begin{equation}
\hat{\mathbf{H}}_j = \mathbf{H}^{\text{sa}} + \mathbf{r}_j \in \mathbb{R}^{N \times D}.
\end{equation}
A lightweight two-layer MLP scorer $\phi_j: \mathbb{R}^D \rightarrow \mathbb{R}$ produces scalar importance logits, which are passed through an element-wise sigmoid rather than softmax:
\begin{equation}
w_{j,i} = \sigma\!\left(\phi_j(\hat{\mathbf{h}}_{j,i})\right) \in (0, 1).
\end{equation}

\paragraph{Why sigmoid rather than softmax?}
Softmax normalizes slice weights onto a probability simplex, introducing a competitive allocation in which increasing the weight of one slice necessarily reduces the relative mass assigned to others. This assumption is poorly aligned with cardiac MRI, where diagnostic evidence is often sparse yet distributed across multiple non-adjacent slices, views, or cardiac phases (e.g., fibrosis may appear in both basal and apical segments). Sigmoid activation instead assigns each slice an independent relevance score, allowing multiple diagnostically informative slices to receive high weights simultaneously.

\subsubsection{Weighted Aggregation and Classification}

The bag-level representation for task $t_j$ is a weighted sum over the task-enhanced slice features:
\begin{equation}
\mathbf{z}_j = \sum_{i=1}^{N} w_{j,i} \, \hat{\mathbf{h}}_{j,i}.
\end{equation}
A task-specific two-layer MLP head $g_j$ produces the final prediction $\hat{y}_j = g_j(\mathbf{z}_j)$. Single-label tasks are optimized with cross-entropy loss; multi-label tasks with per-class binary cross-entropy.

\subsection{Training Objective}
\label{app:mil:obj}

\subsubsection{Sparsity Regularization Toward Target-$K$ Selection}

To encourage the model to concentrate attention on  $K$ informative slices -- matching the downstream MLLM slice budget -- we impose two complementary regularizers on each task's weight vector $\mathbf{w}_j$:
\begin{align}
\mathcal{L}^{(j)}_{\text{sum}} &= \left| \sum_{i=1}^{N} w_{j,i} - K \right|, \\
\mathcal{L}^{(j)}_{\text{bin}} &= \sum_{i=1}^{N} w_{j,i}(1 - w_{j,i}).
\end{align}
Together, these two terms softly enforce the constraint ``select roughly $K$ slices with near-binary confidence,'' while keeping the entire pipeline differentiable.

\subsubsection{Full Objective}

Let $\mathcal{T}_{\text{valid}} = \{j : m_j = 1\}$ denote the set of tasks with available labels in the current bag. The total loss is:
\begin{equation}
\mathcal{L} = \mathcal{L}_{\text{task}} + \frac{\lambda_1}{|\mathcal{T}_{\text{valid}}|} \sum_{j \in \mathcal{T}_{\text{valid}}} \mathcal{L}^{(j)}_{\text{sum}} + \frac{\lambda_2}{|\mathcal{T}_{\text{valid}}|} \sum_{j \in \mathcal{T}_{\text{valid}}} \mathcal{L}^{(j)}_{\text{bin}}.
\end{equation}
We set $\lambda_1 = \lambda_2 = 1.0$.

\subsection{Training Data and Splits}
\label{app:mil:data}

\subsubsection{Data Source and Relation to the Benchmark}

The MIL framework is trained on the \emph{same} patient cohort and \emph{same} structured labels as the CardioLens image understanding tasks (Section~\ref{app:image-understanding-details}). Specifically, the 32 sequence-level classification tasks used by the MIL (11 Cine, 11 LGE, 5 T2, 5 Perfusion) are identical to the 32 image understanding subtasks in the benchmark. Importantly, we clarify the methodological status of this framework:

\begin{itemize}
\item The MIL is \textbf{not a diagnostic model under evaluation}, but a \emph{supervised input-selection control}. It uses task-label supervision to learn per-slice importance scores for the data-driven slice-selection protocol in Section~\ref{sec:slice_selection}.
\item We therefore do \emph{not} maintain a held-out MIL test set. The MIL's learned attention weights are applied to the full benchmark cohort to extract the slice subset supplied to the downstream MLLMs; MLLM evaluation itself is zero-shot and does not use any MIL-trained predictor.
\end{itemize}

\begin{algorithm}[h]
\caption{Data-driven MIL Training (one epoch)}
\label{alg:mil-train}
\begin{algorithmic}[1]
\Require Training set $\mathcal{D} = \{(\mathcal{S}^{(p)}_s, \mathbf{y}^{(p)}_s, \mathbf{m}^{(p)}_s)\}$ indexed by patient $p$ and sequence $s \in \{\text{Cine, LGE, T2, Perfusion}\}$
\Require Shared encoder $f_\theta$, positional encoding $\mathbf{P}$, per-task attention modules $\{\mathcal{A}_j\}$, classifiers $\{g_j\}$
\Require Target selection $K$, sparsity weights $\lambda_1, \lambda_2$, accumulation steps $A$, learning rate $\eta$
\State $\text{grad\_count} \gets 0$
\For{each patient bag $p$ in $\mathcal{D}$}
    \State Sample sequence $s \sim \text{Uniform}(\text{available sequences of } p)$
    \State Load slices $\mathcal{S} = \{s_1, \ldots, s_N\}$, labels $\{y_j\}$, masks $\{m_j\}$ for sequence $s$
    \If{$\sum_j m_j = 0$} \Comment{skip bags with no valid labels}
        \State \textbf{continue}
    \EndIf
    \State \textbf{// Shared feature extraction}
    \State $\mathbf{h}_i \gets f_\theta(s_i)$ for $i = 1, \ldots, N$
    \State $\tilde{\mathbf{H}} \gets [\mathbf{h}_1; \ldots; \mathbf{h}_N] + \mathbf{P}_{1:N}$
    \State \textbf{// Self-attention contextualization}
    \State $\mathbf{H}^{\text{sa}} \gets \text{LayerNorm}(\tilde{\mathbf{H}} + \text{MHSA}(\tilde{\mathbf{H}}, \tilde{\mathbf{H}}, \tilde{\mathbf{H}}))$
    \State $\mathcal{L}_{\text{task}} \gets 0, \; \mathcal{L}_{\text{sum}} \gets 0, \; \mathcal{L}_{\text{bin}} \gets 0$
    \State $\mathcal{T}_{\text{valid}} \gets \{j : m_j = 1\}$
    \For{each valid task $j \in \mathcal{T}_{\text{valid}}$}
        \State \textbf{// Task query cross-attention}
        \State $\mathbf{c}_j \gets \text{CrossAttn}(\mathbf{q}_j, \mathbf{H}^{\text{sa}}, \mathbf{H}^{\text{sa}})$
        \State $\mathbf{r}_j \gets \text{LayerNorm}(\mathbf{c}_j + \text{FFN}(\mathbf{c}_j))$
        \State \textbf{// Slice importance scoring}
        \State $\hat{\mathbf{H}}_j \gets \mathbf{H}^{\text{sa}} + \mathbf{r}_j$
        \State $w_{j,i} \gets \sigma(\phi_j(\hat{\mathbf{h}}_{j,i}))$ for $i = 1, \ldots, N$
        \State \textbf{// Weighted aggregation and prediction}
        \State $\mathbf{z}_j \gets \sum_{i=1}^N w_{j,i} \, \hat{\mathbf{h}}_{j,i}$
        \State $\hat{y}_j \gets g_j(\mathbf{z}_j)$
        \State \textbf{// Accumulate losses}
        \State $\mathcal{L}_{\text{task}} \mathrel{+}= \ell_{\text{task},j}(\hat{y}_j, y_j)$ \Comment{CE for single-label, BCE for multi-label}
        \State $\mathcal{L}_{\text{sum}} \mathrel{+}= |\sum_i w_{j,i} - K|$
        \State $\mathcal{L}_{\text{bin}} \mathrel{+}= \sum_i w_{j,i}(1 - w_{j,i})$
    \EndFor
    \State $\mathcal{L} \gets \mathcal{L}_{\text{task}} + \dfrac{\lambda_1}{|\mathcal{T}_{\text{valid}}|} \mathcal{L}_{\text{sum}} + \dfrac{\lambda_2}{|\mathcal{T}_{\text{valid}}|} \mathcal{L}_{\text{bin}}$
    \State $\mathcal{L} \gets \mathcal{L} / A$
    \State Backpropagate $\mathcal{L}$
    \State $\text{grad\_count} \mathrel{+}= 1$
    \If{$\text{grad\_count} \bmod A = 0$}
        \State Clip gradients to max-norm 1.0
        \State Update parameters: $\theta \gets \theta - \eta \nabla_\theta \mathcal{L}$
        \State Zero gradients
    \EndIf
\EndFor
\end{algorithmic}
\end{algorithm}

\begin{algorithm}[h]
\caption{Importance-Based Slice Selection (Inference)}
\label{alg:mil-infer}
\begin{algorithmic}[1]
\Require Trained MIL model ($f_\theta$, $\mathbf{P}$, $\{\mathcal{A}_j\}$)
\Require Patient $p$ with sequence $s$ of $N$ slices, sequence-associated tasks $\mathcal{T}_s$
\Require Target budget $K$, mode $\in \{\textsc{per-task}, \textsc{task-averaged}\}$
\Ensure Per-task selections $\{\mathcal{I}^*_j\}_{j \in \mathcal{T}_s}$, or shared selection $\mathcal{I}^*$
\State $\mathbf{h}_i \gets f_\theta(s_i)$ for $i = 1, \ldots, N$
\State $\tilde{\mathbf{H}} \gets [\mathbf{h}_1; \ldots; \mathbf{h}_N] + \mathbf{P}_{1:N}$
\State $\mathbf{H}^{\text{sa}} \gets \text{LayerNorm}(\tilde{\mathbf{H}} + \text{MHSA}(\tilde{\mathbf{H}}, \tilde{\mathbf{H}}, \tilde{\mathbf{H}}))$
\For{each task $j \in \mathcal{T}_s$}
    \State $\mathbf{c}_j \gets \text{CrossAttn}(\mathbf{q}_j, \mathbf{H}^{\text{sa}}, \mathbf{H}^{\text{sa}})$
    \State $\mathbf{r}_j \gets \text{LayerNorm}(\mathbf{c}_j + \text{FFN}(\mathbf{c}_j))$
    \State $w_{j,i} \gets \sigma(\phi_j(\mathbf{H}^{\text{sa}}_i + \mathbf{r}_j))$ for $i = 1, \ldots, N$
\EndFor
\If{mode $=$ \textsc{per-task}} \Comment{Image understanding}
    \State $\mathcal{I}^*_j \gets \text{TopK}(\{w_{j,i}\}_{i=1}^N, K)$ for each $j \in \mathcal{T}_s$
    \State \Return $\{\mathcal{I}^*_j\}_{j \in \mathcal{T}_s}$
\Else \Comment{Report generation / disease diagnosis}
    \State $\bar{w}_i \gets \tfrac{1}{|\mathcal{T}_s|} \sum_{j \in \mathcal{T}_s} w_{j,i}$ for $i = 1, \ldots, N$
    \State $\mathcal{I}^* \gets \text{TopK}(\{\bar{w}_i\}_{i=1}^N, K)$
    \State \Return $\mathcal{I}^*$
\EndIf
\end{algorithmic}
\end{algorithm}

\subsubsection{Label Construction}

All 32 task labels are parsed from the same structured clinical reports that underlie the benchmark's image understanding QA pairs.
Single-label tasks produce a one-hot vector; multi-label tasks produce a binary vector over the task's choice set.

\paragraph{Label granularity.} Labels are \emph{sequence-level}, not slice-level. That is, for a given patient and a given sequence, a single set of labels applies to the entire bag of slices from that sequence. This reflects the reality that clinical reports describe findings at the level of anatomical regions and sequences, not individual slices. Consequently, the MIL must learn slice importance purely from the aggregate supervision signal -- no slice-level ground truth is ever provided.

\paragraph{Validity masks.} Not every task is annotated for every patient: some tasks are conditional (e.g., \textit{lge\_high\_region} is meaningful only when abnormal enhancement is present), and some findings are not documented in the report. For each task $t_j$ and each patient, a binary validity mask $m_j$ indicates whether a label is available. Tasks with $m_j = 0$ are excluded from both the supervised loss and the sparsity regularization for that bag.

\subsubsection{Data Augmentation and Class Balancing}
We do not apply data augmentation in the current implementation. To address class imbalance, we adopt a {patient-level oversampling} strategy during training, where patients associated with under-represented classes are sampled more frequently. This increases exposure to rare findings without modifying the original labels or introducing synthetic data. A moderate oversampling ratio (0.5) is used to partially balance the class distribution while avoiding overfitting.

\subsubsection{Bag Construction}
Each training example is a (patient, sequence) bag containing all slices from the sequence's task-relevant views, concatenated in a fixed (view, slice-index) order. The raw bag size $N$ is highly variable across the cohort, spanning
$[3,1500]$ depending on the sequence and the number of available views. To bound GPU memory, bags exceeding 512 slices are uniformly subsampled along the slice index down to 512 before being fed to the model. Because $N$
remains variable even after capping, we set the batch size to one bag and simulate a larger effective batch via gradient accumulation.

\subsubsection{Splits}
\label{app:mil_data_split}
We randomly partition the cohort at the {patient level} into training and validation subsets with an 80\%/20\% ratio. The validation split serves for \emph{model selection}. We retain the checkpoint that achieves the best validation metric as the final model for importance extraction.

\subsection{Algorithm Summary}
\label{app:mil:algo}

We summarize the full training and inference procedure of the Data-driven MIL framework in Algorithms~\ref{alg:mil-train} and~\ref{alg:mil-infer}, respectively. The training loop (Algorithm~\ref{alg:mil-train}) processes one patient bag at a time, randomly sampling an available sequence per step to handle heterogeneous sequence availability across patients. The inference procedure (Algorithm~\ref{alg:mil-infer}) extracts task-level importance scores and selects the top-$K$ slices that are subsequently passed to the downstream MLLM as the importance-based protocol described in Section~\ref{sec:slice_selection}.

\subsection{Implementation Details}
\label{app:mil:impl}
Hyperparameters are summarized in Table~\ref{tab:mil:config}.

\begin{table}[h]
\centering
\small
\caption{MIL framework training configuration.}
\label{tab:mil:config}
\begin{tabular}{ll}
\toprule
Hyperparameter & Value \\
\midrule
Backbone & ResNet-18 (ImageNet pretrained, 1-channel adapted) \\
Feature dimension $D$ & 512 \\
Attention heads $n_h$ & 4 \\
Target selection $K$ & 8 \\
Positional encoding & Learnable, max length 1000 \\
Dropout & 0.1 \\
Optimizer & Adam \\
Learning rate & $1 \times 10^{-5}$ \\
Batch size & 1 patient-sequence bag \\
Gradient accumulation & 64 steps \\
Gradient clipping & max-norm 1.0 \\
Training epochs & 50 \\
Sparsity weights $\lambda_1, \lambda_2$ & 1.0, 1.0 \\
Hardware & Single NVIDIA 4090 GPU ($\sim$24\,GB) \\
\bottomrule
\end{tabular}
\end{table}

\subsection{Model Behavior and Validation}
\label{app:mil:val}

We evaluate the MIL model's task performance, weight distributions, and component ablations before using its learned scores for MLLM slice selection.

\subsubsection{Task Prediction Performance}

Table~\ref{tab:mil:perf} reports the MIL model's own task accuracy on the cohort used for slice selection. Its overall accuracy of 83.6\% demonstrates the framework extracts clinically meaningful signal on this cohort. The MIL is a fully supervised specialist with direct task-label access, and is therefore not directly comparable to zero-shot MLLMs.

\begin{table}[h]
\centering
\small
\caption{MIL task prediction performance on the slice-selection cohort (mean $\pm$ std over three independent runs). Accuracy, Precision, Recall, and F1 are micro-averaged over all QA pairs; Exact-match reports the fraction of samples where all predicted labels match the ground truth set.}
\label{tab:mil:perf}
\resizebox{\linewidth}{!}{
\begin{tabular}{lccccc}
\toprule
Sequence & Accuracy & Precision & Recall & F1 & Exact-match \\
\midrule
Cine       & 0.831 $\pm$ 0.010 & 0.780 $\pm$ 0.012 & 0.791 $\pm$ 0.010 & 0.786 $\pm$ 0.010 & 0.778 $\pm$ 0.011 \\
LGE        & 0.875 $\pm$ 0.003 & 0.816 $\pm$ 0.015 & 0.854 $\pm$ 0.018 & 0.834 $\pm$ 0.004 & 0.682 $\pm$ 0.009 \\
T2         & 0.715 $\pm$ 0.059 & 0.657 $\pm$ 0.047 & 0.722 $\pm$ 0.067 & 0.688 $\pm$ 0.052 & 0.575 $\pm$ 0.066 \\
Perfusion  & 0.871 $\pm$ 0.005 & 0.840 $\pm$ 0.013 & 0.845 $\pm$ 0.007 & 0.842 $\pm$ 0.008 & 0.794 $\pm$ 0.008 \\
\midrule
\textbf{Overall} & \textbf{0.836 $\pm$ 0.006} & \textbf{0.786 $\pm$ 0.004} & \textbf{0.812 $\pm$ 0.014} & \textbf{0.798 $\pm$ 0.004} & \textbf{0.734 $\pm$ 0.002} \\
\bottomrule
\end{tabular}}
\end{table}

\subsubsection{Sparsity Regularization Behavior}

Table~\ref{tab:mil:weights_horizontal} reports the empirical behavior of the learned attention weights across runs.
The average weight sum is 8.10, close to the target $K=8$ and consistent with the intended budget constraint imposed by $\mathcal{L}_{\text{sum}}$.
The large spread between max ($\approx 0.69$) and min ($\approx 0.01$) values
shows that the model assigns differentiated importance scores to the candidate slices.
\begin{table}[htbp]
\centering
\small
\caption{Learned attention weight statistics averaged across three runs.}
\label{tab:mil:weights_horizontal}
\begin{tabular}{lcccc}
\toprule
Statistic & Weight sum $\sum_i w_i$ & Weight mean $\bar{w}_i$ & Weight max & Weight min \\
\midrule
Mean      & 8.10                  & 0.179                  & 0.685      & 0.014      \\
\bottomrule
\end{tabular}
\end{table}

\subsubsection{Component Ablation}
\label{app:mil:ablation}
To examine the contribution of each architectural component, we perform four ablations of the MIL model. As shown in Table~\ref{tab:mil:ablation}, removing self-attention causes the largest F1 drop ($-0.089$), showing that inter-slice context contributes to task prediction. Positional information is also clearly beneficial: removing it entirely drops F1 by $0.057$, while merely replacing the learnable PE with a fixed sinusoidal variant incurs a smaller drop ($-0.022$), suggesting that most of the gain comes from injecting position at all, with learnability providing an additional refinement. Finally, replacing sigmoid with softmax leads to a modest overall drop ($-0.020$ F1). This suggests
that enforcing a normalized, competitive attention distribution is suboptimal for CMR slice selection.

\begin{table}[h]
\centering
\small
\caption{Ablation study of the Data-driven MIL model on the validation set. We report overall performance across all MRI sequences and QA pairs. Values in parentheses denote the absolute performance drop compared to the full model.}
\label{tab:mil:ablation}
\resizebox{1.0\textwidth}{!}{
\begin{tabular}{lccccc}
\toprule
Variant & Accuracy & Precision & Recall & F1 & Exact-match \\
\midrule
\textbf{Full model}           & \textbf{0.836} & \textbf{0.786} & \textbf{0.812} & \textbf{0.798} & \textbf{0.734} \\
\midrule
w/o self-attention            & 0.802 \scriptsize{($-0.034$)} & 0.697 \scriptsize{($-0.089$)} & 0.722 \scriptsize{($-0.090$)} & 0.709 \scriptsize{($-0.089$)} & 0.675 \scriptsize{($-0.059$)} \\
w/o positional encoding       & 0.795 \scriptsize{($-0.041$)} & 0.730 \scriptsize{($-0.056$)} & 0.753 \scriptsize{($-0.059$)} & 0.741 \scriptsize{($-0.057$)} & 0.671 \scriptsize{($-0.063$)} \\
w/o learnable PE (fixed sinusoidal) & 0.821 \scriptsize{($-0.015$)} & 0.775 \scriptsize{($-0.011$)} & 0.777 \scriptsize{($-0.035$)} & 0.776 \scriptsize{($-0.022$)} & 0.717 \scriptsize{($-0.017$)} \\
sigmoid $\rightarrow$ softmax & 0.807 \scriptsize{($-0.029$)} & 0.775 \scriptsize{($-0.011$)} & 0.780 \scriptsize{($-0.032$)} & 0.778 \scriptsize{($-0.020$)} & 0.720 \scriptsize{($-0.014$)} \\
\bottomrule
\end{tabular}
}
\end{table}

\clearpage

\section{Slice Selection Usage Across Task Tiers}
\label{sec:slice-selection-across-task-tiers}

The three slice selection protocols introduced in
Section \ref{sec:slice_selection} were
defined for the \emph{image understanding} setting, where each question is
tied to a single sequence and a single clinical subtask. Applying them to
the other two task tiers requires adapting to (i) multi-subtask aggregation
within a sequence (report generation) and (ii) multi-sequence concatenation
across sequences (disease diagnosis). This section specifies those
adaptations; the protocols themselves are unchanged.

\paragraph{Image understanding (baseline case).}
Each image understanding question concerns a single sequence and a single
subtask. The $K$ slices selected by the active protocol for that
(sequence, subtask) pair are supplied directly as the visual input, along
with the question and answer options.

\paragraph{Report generation (per-sequence, multi-subtask).}
Report generation is still per-sequence -- one report for each evaluated patient--sequence pair -- but a single report covers \emph{all} subtasks defined for that
sequence. The three protocols therefore need to produce a single $K$-slice
input shared across subtasks, rather than a separate selection per
subtask. The adaptation differs per protocol:
\begin{itemize}[leftmargin=*]
    \item \textbf{Random.} The protocol is already subtask-agnostic: $K$
    slices are sampled uniformly at random from the pool of all available
    slices across all views of the sequence. No change from the image
    understanding setting.
    \item \textbf{Heuristic.}  We
    partition the $K$-slice budget across the sequence's canonical views: at the default $K=8$, Cine uses
$(K_{\mathrm{sax}}, K_{\mathrm{4ch}}, K_{\mathrm{3ch}}) = (4, 2, 2)$, with
the 4 SAX slices drawn as end-diastolic (ED) frame uniformly spanning
base to apex, and the 2 frames in each of 4CH and 3CH covering one ED and
one end-systolic (ES) frame; LGE $(K_{\mathrm{4ch}}, K_{\mathrm{sax}}) = (3, K{-}3)$;
    T2 draws all $K$ slices from SAX; Perfusion draws all $K$ frames from
    the dynamic series. Missing views have their budget redistributed to
    the remaining available views to preserve the total budget of $K$.
    \item \textbf{MIL.} The trained MIL model produces
    task-specific attention weights $\{w_{j,i}\}$ for every subtask
    $t_j$ associated with the sequence. Since report generation needs
    a single slice set shared across subtasks, we fuse the per-subtask
    weights into a single \emph{task-averaged importance score} per slice:
    \begin{equation}
    \bar{w}_i \;=\; \frac{1}{|\mathcal{T}_s|} \sum_{j \in \mathcal{T}_s} w_{j,i},
    \qquad i = 1, \ldots, N,
    \end{equation}
    where $\mathcal{T}_s$ is the set of subtasks defined for sequence $s$
    and $N$ is the total number of slices in the patient's bag. The $K$
    slices with the highest $\bar{w}_i$ form the MLLM input.
\end{itemize}
All three adaptations produce a single $K$-slice input per
(patient, sequence), which is then paired with the sequence's report
prompt.

\paragraph{Disease diagnosis (cross-sequence concatenation).}
Diagnosis requires integrating information across all four sequences
simultaneously. For each diagnosis question, we reuse the $K$-slice
selection already computed for each sequence's report generation task
under the active protocol, and concatenate them into a single
$4K$-slice input. At the default $K = 8$ this yields $4K = 32$ slices
per patient.

\clearpage

\section{Experimental Setup Details}
\label{app:experiments}

\subsection{Complete Model List}
\label{app:model-list}

Table~\ref{tab:model-list} summarizes the 24 MLLMs in the main benchmark evaluation, grouped by openness, specialization, and scale.

\begin{table}[h]
\centering
\small
\caption{Complete list of evaluated MLLMs.}
\label{tab:model-list}
\begin{tabular}{@{}p{0.18\linewidth}p{0.08\linewidth}p{0.66\linewidth}@{}}
\toprule
Category & Scale & Models \\
\midrule

\multirow{3}{=}{Closed-source}
&  --  & GPT-5~\cite{singh2025openai} \\
&  --  & Gemini-3.1-Pro~\cite{googledeepmind2026gemini31pro} \\
&  --  & Claude-Opus-4.6~\cite{anthropic2026claudeopus46} \\

\midrule
\multirow{2}{=}{Open-source, general}
& $\leq$10B
& Gemma-4-4B~\cite{google_gemma4_31b_it_hf}, InternVL3.5-8B~\cite{wang2025internvl3}, Qwen3-VL-8B~\cite{bai2025qwen3} \\

& $>$10B
& Llama-3.2-11B~\cite{grattafiori2024llama}, Phi-4-14B~\cite{abdin2024phi}, DeepSeek-VL2-16B~\cite{wu2024deepseek}, Mistral-Small-24B~\cite{mistral_small_31_24b_instruct_2503}, DeepSeek-VL2-27B~\cite{wu2024deepseek}, Gemma-3-27B~\cite{gemmateam2025gemma3}, Gemma-4-31B~\cite{google_gemma4_31b_it_hf}, Qwen3-VL-32B~\cite{bai2025qwen3}, InternVL3.5-38B~\cite{wang2025internvl3} \\

\midrule
\multirow{2}{=}{Open-source, medical}
& $\leq$10B
& MedGemma-4B~\cite{sellergren2025medgemma}, Lingshu-7B~\cite{xu2025lingshu}, Fleming-8B~\cite{liu2025fleming}, HuluMed-7B~\cite{jiang2025hulu}, QoQMed-7B~\cite{dai2025qoq} \\

& $>$10B
& MedGemma-27B~\cite{sellergren2025medgemma}, Hulu-30B~\cite{jiang2025hulu}, Lingshu-32B~\cite{xu2025lingshu}, Fleming-38B~\cite{liu2025fleming} \\

\bottomrule
\end{tabular}
\end{table}

\paragraph{Computational Resources.}
Open-source MLLM inference was conducted on NVIDIA H100 GPUs
(80\,GB memory per GPU), while closed-source models were
evaluated through their official APIs.
The computational resources for MIL training are detailed
in Appendix~\ref{app:mil:impl}.

\subsection{Prompt Templates}
\label{app:prompts}

All evaluation prompts follow a standardized template per task tier. Each
prompt establishes the assistant as a cardiac MRI VLM, describes the task,
and enforces strict output formatting rules to enable deterministic parsing
and fair comparison across models. In standard classification prompts, we forbid free-form reasoning and extra characters so that models return the answer alone. Complete
templates are given below.

\paragraph{Image Understanding.}
For each image understanding question, the prompt specifies (i) the imaging
sequence, (ii) the clinical subtask, (iii) the answer options
(single-select or multi-select depending on the subtask), and (iv) the
output format. Template and one concrete instantiation
(\texttt{wall\_thickness}, multi-select) shown below; all other subtasks
follow the same structure with different options and single/multi-select
rules.
\begin{tcolorbox}[
  colback=gray!10,
  colframe=gray!50,
  boxrule=0.5pt,
  arc=2pt,
  left=6pt, right=6pt, top=4pt, bottom=4pt,
  breakable
]
\small
\begin{verbatim}
You are a Vision-Language Model (VLM) for cardiac {sequence} MRI.
Task: using the provided image frames, answer the {subtask_description}
({single-select | multi-select}).

Strict rules:
- Select only from {option_letters}; one or multiple choices allowed.
  (or: exactly one choice)
- Output MUST be letters only; use English commas for multiple selections.
- Do NOT output explanations, reasoning, descriptions, confidence, or any
  extra characters (commas allowed).
- Even if uncertain, you must output the most likely choice(s); do not
  answer "cannot determine/uncertain".

Options:
{options}
\end{verbatim}
\end{tcolorbox}
\noindent\textit{Concrete example (Cine, \texttt{wall\_thickness}, multi-select):}
\begin{tcolorbox}[
  colback=gray!10,
  colframe=gray!50,
  boxrule=0.5pt,
  arc=2pt,
  left=6pt, right=6pt, top=4pt, bottom=4pt,
  breakable
]
\small
\begin{verbatim}
You are a Vision-Language Model (VLM) for cardiac cine MRI.
Task: using the provided image frames, answer the LV wall thickness
characteristic (multi-select).

Strict rules:
- Select only from A/B/C/D/E; one or multiple choices allowed.
- Output MUST be letters only; use English commas for multiple selections.
- Do NOT output explanations, reasoning, descriptions, confidence, or any
  extra characters (commas allowed).
- Even if uncertain, you must output the most likely choice(s); do not
  answer "cannot determine/uncertain".

Options:
A. Normal
B. Thinned
C. Heterogeneous thickness
D. Thickened
E. Bulging
\end{verbatim}
\end{tcolorbox}

\paragraph{Reasoning Template (Image Understanding).}
We extend the standard image understanding prompt
with a field-specific reasoning template that elicits a structured
justification \emph{after} the model gives its answer.
Concretely, the ``no explanation'' rule is lifted and replaced with a
two-line output contract: line one contains the option letter(s), and
line two begins with \texttt{Reason:} followed by a short justification.
The reasoning framework itself enumerates the visual evidence a
radiologist would cross-check for that specific subtask -- e.g.,
segment-wise thickness for \texttt{wall\_thickness}, jet detection in
the appropriate cardiac phase for valvular regurgitation, or temporal
persistence checks for perfusion abnormalities. Each
image understanding subtask across cine, LGE, T2, and perfusion
sequences has its own framework; the template for the
\texttt{wall\_thickness} subtask used above is shown below. This template corresponds to the ``w/ R'' condition in Table~\ref{tab:reasoning_comparison_random}. Variants without the clinical framework or with the rationale generated before the answer are evaluated in Appendix~\ref{app:prompt_order_controls}.

\begin{tcolorbox}[
  colback=gray!10,
  colframe=gray!50,
  boxrule=0.5pt,
  arc=2pt,
  left=6pt, right=6pt, top=4pt, bottom=4pt,
  breakable
]
\small
\begin{verbatim}
- After providing your answer, you must provide a reason for your choice.
- Output format:
  - Line 1: Answer letter(s) only (e.g., A or A,B)
  - Line 2: "Reason: [your explanation]"

When providing your reason, please follow this analysis framework:
Your reason should include:
1. Wall thickness assessment: Describe the overall thickness of the LV
   myocardium across different segments (anterior, inferior, septal,
   lateral).
2. Uniformity analysis: Note whether thickness is uniform or
   heterogeneous across segments.
3. Specific observations: Identify any segments showing thinning,
   thickening, or bulging.
4. Frame-by-frame consistency: Describe if thickness characteristics
   remain consistent across cardiac cycle frames.
5. Comparison to normal: Compare observed thickness patterns to what
   would be expected in normal myocardium.

Example structure: "The LV myocardium shows [uniform/heterogeneous]
thickness. [Specific segment observations]. [Comparison across frames].
This pattern indicates [normal/thinned/thickened/bulging]
characteristics."
\end{verbatim}
\end{tcolorbox}

\paragraph{Report Generation.}
Report generation prompts request a single coherent paragraph covering all
clinical fields required for the target sequence. We constrain the output
format (single paragraph, clinical style) and require that every listed
field be addressed, so that evaluation metrics can score field-wise
coverage without parser ambiguity.
\begin{tcolorbox}[
  colback=gray!10,
  colframe=gray!50,
  boxrule=0.5pt,
  arc=2pt,
  left=6pt, right=6pt, top=4pt, bottom=4pt,
  breakable
]
\small
\begin{verbatim}
You are a Vision-Language Model (VLM) for cardiac cine MRI.
Task: Using the provided image frames, generate a structured clinical
report as a single coherent paragraph.

Your report must include:
- LV wall thickness and morphology
- LV Wall motion coordination and amplitude
- LV systolic and diastolic function
- Valve assessment: mitral, tricuspid, and aortic regurgitation
- Pericardial and pleural effusion status
- Any additional findings (e.g., Positive SAM sign / LVOT obstruction /
  Spade sign / Aortic stenosis / Pulmonary artery dilatation)

Strict rules:
- Write as a single coherent paragraph in clinical report style.
- Base all findings strictly on visual image evidence.
- Cover all listed items; do not omit any category.
\end{verbatim}
\end{tcolorbox}
\noindent LGE, T2, and Perfusion use the same template with the field list
replaced by the corresponding sequence-specific items (scar location and
pattern for LGE, edema distribution for T2, perfusion defect location and
extent for Perfusion).

\paragraph{Disease Diagnosis.}
Diagnosis prompts take all four cardiac sequences jointly as visual input
and require a single categorical answer. The instruction explicitly directs
the model to integrate morphology, wall motion, enhancement patterns, and
overall function before answering -- mirroring how a radiologist
cross-references sequences in clinical practice.

\begin{tcolorbox}[
  colback=gray!10,
  colframe=gray!50,
  boxrule=0.5pt,
  arc=2pt,
  left=6pt, right=6pt, top=4pt, bottom=4pt,
  breakable
]
\small
\begin{verbatim}
You are a Vision-Language Model (VLM) for cardiac MRI diagnostic
classification.
Task: Using the provided cardiac MRI image frames, classify this case
into the most appropriate diagnostic category.

Choose exactly one from the following options:
A. Normal Heart
B. Ischemic Heart Disease
C. Non-Ischemic Cardiomyopathy

Strict rules:
- Consider LV morphology, wall motion, enhancement patterns, and overall
  cardiac function.
- Base all findings strictly on visual image evidence.
- Provide only the letter and label of the selected option.
  No explanations,reasoning, descriptions, confidence, or extra symbols.
- Even if uncertain, choose the most likely answer.
\end{verbatim}
\end{tcolorbox}

\subsection{Evaluation Metric Details}
\label{app:metrics}

\paragraph{Image Understanding.}
For each subtask, \textit{accuracy} is defined by question type.
For \textbf{single-label} questions, accuracy is the fraction of instances where the predicted label exactly matches the ground truth (\textit{exact match}).
For \textbf{multi-label} questions, we treat each candidate option as an independent binary label: given the option set $\mathcal{U}$, the ground-truth set $Y$, and the predicted set $\hat{Y}$, the per-instance accuracy is
$
\mathrm{Acc} = \frac{|\{o\in\mathcal{U} : \mathbf{1}[o\in\hat{Y}] = \mathbf{1}[o\in Y]\}|}{|\mathcal{U}|},
$
i.e.\ the fraction of options correctly included or excluded.
For example, with options $\{A,B,C,D\}$, $Y=\{A,C\}$ and $\hat{Y}=\{A,D\}$, options $A$ and $B$ are correct while $C$ and $D$ are wrong, giving $\mathrm{Acc}=2/4=0.5$.

\textit{Precision}, \textit{recall}, and \textit{F1} are computed by \textbf{micro-averaging} across all instances in the subtask: true positives, false positives, and false negatives are accumulated across all instances, then combined into $P$, $R$, and $F_1 = 2PR/(P+R)$.

\paragraph{Report Generation.}
\label{app:report-generation-metrics}
Report generation evaluates whether a model can synthesize visual evidence from a single CMR sequence into a structured and clinically meaningful report. Since CMR reports require accurate clinical findings, abnormality status, anatomical localization, and tissue-level attributes, we use five complementary metrics: Clinical Findings F1, Checklist Graph-F1, Clinical BERTScore, ROUGE-L, and CIDEr-lite.

\smallskip
\noindent\textit{Sequence-specific Checklist Extraction.}
Before metric computation, both ground-truth and generated reports are parsed into a unified sequence-specific checklist representation. Cine focuses on morphology, motion, and function; LGE on delayed enhancement and its distribution; Perfusion on perfusion defects; and T2 on edema, inflammation, and abnormal signal. The parser extracts finding status, labels, and attributes from natural-language reports, normalizes synonymous expressions, and handles negation before assigning positive findings. The checklist scope is summarized in Table~\ref{tab:report_generation_checklist_scope}.

\begin{table}[h]
\centering
\caption{Sequence-specific checklist scope for report generation evaluation.}
\label{tab:report_generation_checklist_scope}
\begin{tabularx}{\linewidth}{lX}
\toprule
\textbf{Sequence} & \textbf{Extracted findings and attributes} \\
\midrule
Cine & LV wall morphology, wall motion, systolic/diastolic function, valvular regurgitation, pericardial/pleural effusion, SAM sign, and LVOT obstruction. \\
LGE & Delayed enhancement status, signal characteristics, high-/low-signal abnormality locations, pericardial abnormality, and abnormal signals at septal insertion points or the right ventricle. \\
Perfusion & Perfusion status, reduced/delayed perfusion, perfusion defect, affected segment, and myocardial layer. \\
T2 & High-/low-/mixed-signal abnormality, edema/inflammation-related findings, location, myocardial layer, and distribution pattern. \\
\bottomrule
\end{tabularx}
\end{table}

Localization attributes are normalized into four categories: segment, wall region, myocardial layer, and distribution pattern.

\smallskip
\noindent\textit{Clinical Findings F1.}
Clinical Findings F1 measures finding-level factual consistency between the generated and ground-truth reports. After checklist extraction, each report is represented as a set of clinical atoms \(A\), including status and label facts, e.g.,
\begin{tcolorbox}[
  colback=gray!10,
  colframe=gray!50,
  boxrule=0.5pt,
  arc=2pt,
  left=6pt, right=6pt, top=4pt, bottom=4pt,
  breakable
]
\small
\begin{verbatim}
lv_systolic_function:status=normal
perfusion_status:status=abnormal
cine_additional_findings:label=LVOT obstruction
lge_signal_characteristics:label=high signal
\end{verbatim}
\end{tcolorbox}

Given the atom sets \(A_{\mathrm{gt}}^{(i)}\) and \(A_{\mathrm{pred}}^{(i)}\) for the \(i\)-th sample, we compute the corpus-level micro F1 as
\begin{equation}
F1_{\mathrm{clinical}}
=
\frac{
2\sum_{i=1}^{N}|A_{\mathrm{pred}}^{(i)}\cap A_{\mathrm{gt}}^{(i)}|
}{
\sum_{i=1}^{N}|A_{\mathrm{pred}}^{(i)}|
+
\sum_{i=1}^{N}|A_{\mathrm{gt}}^{(i)}|
+\epsilon
}.
\end{equation}
This metric evaluates whether key clinical findings, normal/abnormal status, and additional findings are correctly reported, without requiring identical wording.

\smallskip
\noindent\textit{Checklist Graph-F1.}
Checklist Graph-F1 evaluates relations between findings and attributes. Each report is converted into graph triples \(G=\{(s,r,o)\}\), where \(s\) is a checklist finding, \(r\) is a relation type, and \(o\) is a status, label, or attribute value. The relation types include:
\begin{tcolorbox}[
  colback=gray!10,
  colframe=gray!50,
  boxrule=0.5pt,
  arc=2pt,
  left=6pt, right=6pt, top=4pt, bottom=4pt,
  breakable
]
\small
\begin{verbatim}
has_status
has_label
located_in_segment
located_at_wall
involves_layer
has_distribution
\end{verbatim}
\end{tcolorbox}

For example, ``patchy subepicardial high signal in the basal inferolateral wall'' yields triples such as:
\begin{tcolorbox}[
  colback=gray!10,
  colframe=gray!50,
  boxrule=0.5pt,
  arc=2pt,
  left=6pt, right=6pt, top=4pt, bottom=4pt,
  breakable
]
\small
\begin{verbatim}
(lge_high_signal_location, located_in_segment, basal)
(lge_high_signal_location, located_at_wall, inferolateral)
(lge_high_signal_location, involves_layer, subepicardial)
(lge_high_signal_location, has_distribution, patchy)
\end{verbatim}
\end{tcolorbox}

Given \(G_{\mathrm{gt}}^{(i)}\) and \(G_{\mathrm{pred}}^{(i)}\) for the \(i\)-th sample, we compute
\begin{equation}
F1_{\mathrm{graph}}
=
\frac{
2\sum_{i=1}^{N}|G_{\mathrm{pred}}^{(i)}\cap G_{\mathrm{gt}}^{(i)}|
}{
\sum_{i=1}^{N}|G_{\mathrm{pred}}^{(i)}|
+
\sum_{i=1}^{N}|G_{\mathrm{gt}}^{(i)}|
+\epsilon
}.
\end{equation}
Compared with Clinical Findings F1, this metric more strongly emphasizes structural completeness, such as whether an abnormality is assigned to the correct segment, wall region, myocardial layer, and distribution pattern.

\smallskip
\noindent\textit{Clinical BERTScore.}
Clinical BERTScore measures contextual semantic similarity between generated and ground-truth reports. Let the ground-truth and generated token sequences be \(X=\{x_j\}_{j=1}^{m}\) and \(\hat{X}=\{\hat{x}_i\}_{i=1}^{n}\), with contextual embeddings \(h_j\) and \(\hat{h}_i\). We compute
\begin{equation}
P_{\mathrm{BERT}}
=
\frac{1}{n}\sum_{i=1}^{n}\max_j \cos(\hat{h}_i,h_j),
\quad
R_{\mathrm{BERT}}
=
\frac{1}{m}\sum_{j=1}^{m}\max_i \cos(\hat{h}_i,h_j),
\end{equation}
\begin{equation}
F1_{\mathrm{BERT}}
=
\frac{2P_{\mathrm{BERT}}R_{\mathrm{BERT}}}
{P_{\mathrm{BERT}}+R_{\mathrm{BERT}}+\epsilon},
\qquad
\overline{F1}_{\mathrm{BERT}}
=
\frac{1}{N}\sum_{i=1}^{N}F1_{\mathrm{BERT}}^{(i)}.
\end{equation}
This metric reduces penalties for semantically equivalent wording, but does not explicitly model negation, abnormality status, or attribute grounding.

\smallskip
\noindent\textit{ROUGE-L.}
ROUGE-L measures lexical sequence overlap using the longest common subsequence. Let \(m\) and \(n\) be the lengths of the ground-truth and generated reports, and \(L=\mathrm{LCS}(X,\hat{X})\). We compute
\begin{equation}
F1_{\mathrm{ROUGE-L}}
=
\frac{2\cdot \frac{L}{n+\epsilon}\cdot \frac{L}{m+\epsilon}}
{\frac{L}{n+\epsilon}+\frac{L}{m+\epsilon}+\epsilon},
\qquad
\overline{F1}_{\mathrm{ROUGE-L}}
=
\frac{1}{N}\sum_{i=1}^{N}F1_{\mathrm{ROUGE-L}}^{(i)}.
\end{equation}
ROUGE-L captures surface-form similarity and ordering, but does not explicitly assess medical entities, negation, or attribute grounding.

\smallskip
\noindent\textit{CIDEr-lite.}
CIDEr-lite measures TF-IDF-weighted \(n\)-gram similarity between generated and ground-truth reports. For the \(k\)-th \(n\)-gram with document frequency \(df_k\) in a corpus of size \(D\), its inverse document frequency is
\begin{equation}
\mathrm{idf}_k
=
\log\frac{D+1}{df_k+1}.
\end{equation}
Let \(\mathbf{w}_n(s)\) denote the TF-IDF vector for \(n\)-grams of order \(n\). CIDEr-lite is computed as
\begin{equation}
\mathrm{CIDEr\text{-}lite}
=
\frac{1}{4}\sum_{n=1}^{4}
\cos\left(\mathbf{w}_n(s_{\mathrm{pred}}),\mathbf{w}_n(s_{\mathrm{gt}})\right),
\qquad
\overline{\mathrm{CIDEr\text{-}lite}}
=
\frac{1}{N}\sum_{i=1}^{N}
\mathrm{CIDEr\text{-}lite}^{(i)}.
\end{equation}
CIDEr-lite gives higher weight to less frequent but clinically specific phrases. Together with ROUGE-L and Clinical BERTScore, it provides complementary text-quality assessment, while clinical factual correctness is mainly captured by Clinical Findings F1 and Checklist Graph-F1.

\paragraph{Disease Diagnosis.}
Disease diagnosis is evaluated as a single-label multi-class classification task. Accuracy follows the same exact-match definition as the single-label setting in Image Understanding.

For precision, recall, and F1, we report macro-averaged scores. In single-label multi-class classification, micro-F1 is largely aligned with overall accuracy, so it provides limited additional information beyond exact-match accuracy. Macro-averaging instead computes the metric separately for each disease category and then averages across categories, giving each category equal weight. This makes the evaluation more sensitive to model performance on minority disease categories and reduces the chance that rare diseases are masked by larger categories.

For $M\in\{P,R,F1\}$, the reported macro score is
\begin{equation}
 M_{\mathrm{macro}} = \frac{1}{|\mathcal{C}|}\sum_{c\in\mathcal{C}} M_c,
\end{equation}
where $\mathcal{C}$ denotes the set of diagnosis categories. For each category $c$, $P_c$, $R_c$, and $F1_c$ are computed in a one-vs-rest manner: samples of category $c$ are treated as positives, all other categories are treated as negatives, and the resulting category-level precision, recall, and F1 are then averaged across categories.

\clearpage

\section{Related Benchmarks and Datasets}
\label{app:related_works_and_datasets}
\subsection{Datasets of Cardiac Images}
\label{app:Datasets_of_Cardiac_Images}

Compared with representative cardiac MRI datasets, CardioLens is designed to cover a broader and more clinically realistic spectrum of cardiac characteristics.
As shown in Table~\ref{tab:cardiac_dataset_comparison}, widely-used public benchmarks such as ACDC~\cite{bernard2018deep}, EMIDEC~\cite{lalande2020emidec},  CARE MyoPS++~\cite{li2023myops,qiu2023myops,ding2023aligning}, LAScarQS++~\cite{li2022review,li2020atrialscar,li2022atrialjsqnet,li2021atrialgeneral}, MBAS 2024~\cite{zhao2024mbas}, and HVSMR-2.0~\cite{pace2024hvsmr} collectively cover cine-based cardiac function assessment, LGE-based myocardial infarction evaluation, whole-heart segmentation, multi-sequence characterization, atrial segmentation, and congenital heart disease assessment. However, these resources are still typically centered on a specific modality, sequence, anatomical target, or task, and most do not jointly provide multi-sequence cardiac MRI, dynamic cine information, multiple clinical views, paired reports, and patient-level diagnostic labels within a unified benchmark. In contrast, CardioLens integrates Cine, LGE, Perfusion, and T2-weighted examinations from real clinical archives, together with report-aligned annotations and diagnosis-oriented evaluation, thereby offering a more complete representation of the real cardiac MRI diagnostic workflow.

\begin{table*}[htbp]
\centering
\caption{Comparison with representative cardiac MRI datasets. Each dataset is characterized by five binary attributes, its sequences, and its patient/volume counts. The binary attributes are: \textit{Multi-Seq.}  --  provides more than one MRI sequence; \textit{Multi-View}  --  provides multiple acquisition views (e.g., short-axis, 2-/3-/4-chamber long-axis); \textit{Reports}  --  provides paired radiology reports; \textit{Diagnosis}  --  provides patient-level diagnostic labels; \textit{Temporal}  --  provides time-resolved imaging across the cardiac cycle (e.g., cine sequences or first-pass perfusion dynamics) rather than only static snapshots. \textit{Sequences} lists the MRI sequences included, and \textit{Scale} reports the number of unique patients and the total number of imaging volumes. A red \ding{51} indicates the attribute is supported; \ding{55} indicates it is not.}
\label{tab:cardiac_dataset_comparison}
\tiny
\setlength{\tabcolsep}{4pt}
\renewcommand{\arraystretch}{1.1}
\renewcommand{\tabularxcolumn}[1]{m{#1}}
\begin{tabularx}{\textwidth}{>{\raggedright\arraybackslash}m{2.4cm} c c c c c >{\raggedright\arraybackslash}X >{\raggedright\arraybackslash}m{2.4cm}}
\toprule
\textbf{Dataset} & \textbf{Multi-Seq} & \textbf{Multi-View} & \textbf{Reports} & \textbf{Diagnosis} & \textbf{Temporal} & \textbf{Sequences}\textsuperscript{\dag} & \textbf{Scale (Patients / Volumes)} \\
\midrule
\href{https://www.creatis.insa-lyon.fr/Challenge/acdc/}{ACDC}
& \ding{55} & \ding{55} & \ding{55} & {\ding{51}} & {\ding{51}}
& Cine
& 150 / 150 \\
\href{https://emidec.com}{EMIDEC}
& \ding{55} & \ding{55} & \ding{55} & {\ding{51}} & \ding{55}
& LGE
& 150 / 150 \\
\href{https://zmic.org.cn/care_2024/track4/}{CARE MyoPS++}
& {\ding{51}} & \ding{55} & \ding{55} & \ding{55} & \ding{55}
& bSSFP, LGE, T2
& 250 / 564 \\
\href{https://www.zmic.org.cn/care_2024/track2/}{LAScarQS++}
& \ding{55} & \ding{55} & \ding{55} & {\ding{51}} & \ding{55}
& LGE
& 194 / 194 \\
\href{https://codalab.lisn.upsaclay.fr/competitions/18516\#learn_the_details-overview}{MBAS 2024}
& \ding{55} & \ding{55} & \ding{55} & \ding{55} & \ding{55}
& LGE
& 200 / 200 \\
\href{https://www.nature.com/articles/s41597-024-03469-9}{HVSMR-2.0}
& \ding{55} & \ding{55} & \ding{55} & \ding{55} & \ding{55}
& bSSFP
& 60 / 60 \\
\rowcolor{gray!20}
\textbf{CardioLens (Ours)}
& {\ding{51}} & {\ding{51}} & {\ding{51}} & {\ding{51}} & {\ding{51}}
& Cine, LGE, T2, Perfusion
& 585 / 5969 \\
\bottomrule
\end{tabularx}

\footnotesize\textsuperscript{\dag}\textit{Cine} denotes dynamic cine imaging acquired with a balanced SSFP (bSSFP) sequence at multiple time points across the cardiac cycle; \textit{bSSFP} alone denotes a single-phase (static) bSSFP acquisition. \textit{LGE} = late gadolinium enhancement; \textit{T2} = T2-weighted imaging; \textit{Perfusion} = first-pass perfusion imaging.
\end{table*}

\subsection{Datasets of Medical Benchmarks}
\label{app:Datasets_of_Medical_MLLM}
\begin{table*}[htbp]
\centering
\caption{Comparison with representative medical LLM/MLLM benchmarks.
\textbf{Data}: data modality and native dimensionality; CardioLens evaluation uses selected original slices or frames for current MLLMs.
\textbf{Source}: Pub.\ (public repositories), Priv.\ (private hospital archives).
\textbf{Scale}: number of evaluation items (QA pairs, reports, or rationales).
\textbf{Integration}: the level of multi-source information a task requires the model to jointly reason over—e.g., multiple views, imaging sequences, or temporal frames of the same patient.
\textbf{Perception}: image-level understanding tasks (VQA, modality/anatomy recognition).
\textbf{Reporting}: free-form long report or rationale generation.
\textbf{Slice Protocol}: standardized slice-selection comparison for examining input-construction effects.}
\label{tab:mllm_benchmark_comparison}
\scriptsize
\setlength{\tabcolsep}{4pt}
\renewcommand{\arraystretch}{1.25}
\renewcommand{\tabularxcolumn}[1]{m{#1}}
\begin{tabularx}{\textwidth}{
>{\raggedright\arraybackslash\hsize=1.55\hsize}X
>{\centering\arraybackslash\hsize=1.05\hsize}X
>{\centering\arraybackslash\hsize=0.70\hsize}X
>{\centering\arraybackslash\hsize=0.70\hsize}X
>{\centering\arraybackslash\hsize=1.55\hsize}X
>{\centering\arraybackslash\hsize=0.75\hsize}X
>{\centering\arraybackslash\hsize=0.75\hsize}X
>{\centering\arraybackslash\hsize=0.95\hsize}X
}
\toprule
\textbf{Benchmark} & \textbf{Data} & \textbf{Source} & \textbf{Scale} & \textbf{Integration} & \textbf{Percep.} & \textbf{Report.} & \makecell{\textbf{Slice}\\\textbf{Protocol}} \\
\midrule
\multicolumn{8}{@{}l}{\textit{Text-only medical benchmarks}} \\
\href{https://huggingface.co/datasets/bigbio/med_qa}{MedQA}         & Text             & Pub.  & 61K   & --                       & \ding{55} & \ding{55} & \ding{55} \\
\href{https://pubmedqa.github.io/}{PubMedQA}      & Text             & Pub.  & 273K  & --                       & \ding{55} & \ding{55} & \ding{55} \\
\midrule
\multicolumn{8}{@{}l}{\textit{Multimodal medical benchmarks (static 2D / 3D)}} \\
\href{https://huggingface.co/datasets/flaviagiammarino/vqa-rad}{VQA-RAD}       & Text + 2D        & Pub.  & 3.5K  & Single image             & {\ding{51}} & \ding{55} & \ding{55} \\
\href{https://www.med-vqa.com/slake/}{SLAKE}       & Text + 2D        & Pub.  & 14K   & Single image             & {\ding{51}} & \ding{55} & \ding{55} \\
\href{https://huggingface.co/datasets/foreverbeliever/OmniMedVQA}{OmniMedVQA}    & Text + 2D        & Pub.  & 118K  & Single image             & {\ding{51}} & \ding{55} & \ding{55} \\
\href{https://hyper.ai/en/datasets/35476}{GMAI-MMBench}  & Text + 2D        & Pub.  & 26K   & Single image             & {\ding{51}} & \ding{55} & \ding{55} \\
\href{https://cares-ai.github.io/}{CARES}         & Text + 2D        & Pub.  & 41K   & Single image             & {\ding{51}} & {\ding{51}} & \ding{55} \\
\href{https://github.com/BAAI-DCAI/M3D}{M3D}     & Text + 3D        & Pub.  & 120K  & Single volume            & {\ding{51}} & {\ding{51}} & \ding{55} \\
\midrule
\rowcolor{gray!20}
\textbf{CardioLens} & \textbf{Text + 3D / 4D} & \textbf{Priv.} & \textbf{13.5K} & \textbf{Views, Seqs., Phases} & {\ding{51}} & {\ding{51}} & {\ding{51}} \\
\bottomrule
\end{tabularx}
\end{table*}

Compared with representative medical LLM/MLLM benchmarks, CardioLens is designed not as another multimodal QA resource, but as a clinically grounded stress test for high-dimensional cardiac prediction. As summarized in Table~\ref{tab:mllm_benchmark_comparison}, text-only benchmarks such as MedQA~\cite{jin2020disease} and PubMedQA~\cite{jin2019pubmedqa} evaluate knowledge recall but provide no visual grounding. Multimodal benchmarks including SLAKE~\cite{liu2021slake}, VQA-RAD~\cite{lau2018dataset}, OmniMedVQA~\cite{hu2024omnimedvqa}, GMAI-MMBench~\cite{chen2024gmai}, CARES~\cite{xia2024cares}, and M3D-Bench~\cite{bai2024m3d} have substantially advanced the field in scale and modality coverage, yet remain confined to static or single-volume inputs.  The benchmarks summarized here draw on public resources and do not provide the same controlled comparisons of slice selection, leaving the effect of input construction less explicit. CardioLens addresses these gaps along three axes: (i) it is built from private hospital archives of 585 patients with four cardiac MRI sequence types, preserving their spatial, temporal, and study-level organization within a unified resource; (ii) it covers the full clinical workflow from image perception  to free-form report generation and diagnosis; and (iii) it introduces controlled slice-selection protocols (random / heuristic / data-driven) to measure how model performance changes with the images selected for input.

\clearpage

\section{Detailed Evaluation Results in CardioLens}
\label{app:more_detailed_results}

\subsection{Report Generation}
\label{app:detailed_report_results}

\begin{table*}[htbp]
\centering
\small
\setlength{\tabcolsep}{2.2pt}
\renewcommand{\arraystretch}{1.15}
\caption{\textbf{Detailed report generation performance (Level 2).} Clinical Findings F1, Checklist Graph-F1, Clinical BERTScore, ROUGE-L, and CIDEr-lite (\%) under three slice-selection strategies. Metrics are defined in Appendix~\ref{app:metrics}.}
\label{tab:report_generation}

\newcommand{\best}[1]{\textbf{#1}}
\newcommand{\second}[1]{\underline{#1}}

\resizebox{1.0\textwidth}{!}{
\begin{tabular}{llccccccccccccccc}
\toprule
& \textbf{Model} &
\multicolumn{3}{c}{\textbf{Clinical Findings F1 (\%)}} &
\multicolumn{3}{c}{\textbf{Checklist Graph-F1 (\%)}} &
\multicolumn{3}{c}{\textbf{Clinical BERTScore (\%)}} &
\multicolumn{3}{c}{\textbf{ROUGE-L (\%)}} &
\multicolumn{3}{c}{\textbf{CIDEr-lite (\%)}} \\

\cmidrule(lr){3-5}
\cmidrule(lr){6-8}
\cmidrule(lr){9-11}
\cmidrule(lr){12-14}
\cmidrule(lr){15-17}

& & Random & Heuristic & MIL
& Random & Heuristic & MIL
& Random & Heuristic & MIL
& Random & Heuristic & MIL
& Random & Heuristic & MIL \\

\midrule

\multicolumn{17}{l}{\textit{Open-Source Models ($\leq$10B)}} \\
\midrule

\multirow{5}{*}{\rotatebox{90}{\scriptsize Medical}}
& MedGemma-4B
& 45.03 & 45.88 & \second{48.25}
& 35.07 & 35.58 & \second{37.54}
& \best{93.47} & \best{93.52} & \best{93.62}
& \best{18.47} & \best{18.77} & \best{19.27}
& 11.03 & \second{11.45} & 11.32 \\

& Lingshu-7B
& \second{47.04} & \second{46.21} & 47.80
& \second{37.30} & \second{36.21} & 37.31
& 93.08 & 93.06 & 93.06
& 14.87 & 15.12 & 14.95
& 11.32 & 11.38 & \second{11.34} \\

& Fleming-8B
& 44.99 & 45.22 & 45.50
& 35.85 & 35.69 & 35.92
& 93.25 & 93.24 & 93.23
& 15.84 & 15.74 & 15.66
& \second{11.74} & 11.44 & 11.23 \\

& HuluMed-7B
& 42.30 & 42.04 & 42.36
& 32.99 & 32.66 & 32.96
& \second{93.44} & \second{93.47} & \second{93.44}
& \second{17.26} & \second{17.31} & \second{17.69}
& \best{12.39} & \best{12.39} & \best{12.61} \\

& QoQMed-7B
& \best{50.22} & \best{49.10} & \best{49.90}
& \best{39.80} & \best{38.84} & \best{39.48}
& 92.66 & 92.68 & 92.71
& 12.02 & 12.23 & 12.52
& 11.09 & 11.20 & 11.16 \\

\cmidrule(l){2-17}

\multirow{3}{*}{\rotatebox{90}{\scriptsize General}}
& Gemma-4-4B
& \second{43.27} & \second{45.77} & \second{48.68}
& \second{34.61} & \second{36.57} & \second{39.23}
& \second{93.11} & \second{93.18} & \second{93.22}
& \best{16.36} & \best{16.78} & \best{17.74}
& \second{10.41} & \second{10.39} & \second{12.10} \\

& InternVL3.5-8B
& 11.57 & 11.02 & 13.39
& 7.40 & 7.12 & 8.62
& 92.93 & 92.94 & 92.93
& \second{14.98} & \second{15.44} & 14.97
& 8.41 & 8.64 & 8.62 \\

& Qwen3-VL-8B
& \best{51.39} & \best{50.70} & \best{50.94}
& \best{40.29} & \best{39.70} & \best{40.73}
& \best{93.21} & \best{93.23} & \best{93.29}
& 14.61 & 14.87 & \second{15.00}
& \best{12.35} & \best{12.49} & \best{12.80} \\

\midrule

\multicolumn{17}{l}{\textit{Open-Source Models ($>$10B)}} \\
\midrule

\multirow{4}{*}{\rotatebox{90}{\scriptsize Medical}}
& MedGemma-27B
& 48.29 & \second{48.05} & 47.76
& \best{40.37} & \best{39.94} & \second{39.75}
& \best{93.43} & \best{93.36} & \best{93.40}
& \best{19.66} & \best{19.50} & \best{19.32}
& \best{13.01} & \best{13.06} & \best{12.71} \\

& Hulu-30B
& 46.93 & 46.80 & \second{48.41}
& 35.68 & 35.07 & 37.17
& 93.07 & \second{93.10} & 93.07
& \second{15.23} & \second{15.48} & \second{15.12}
& \second{11.35} & \second{11.28} & 11.09 \\

& Lingshu-32B
& \best{52.36} & \best{52.66} & \best{51.50}
& \second{40.30} & \second{39.63} & \best{40.25}
& \second{93.12} & 93.09 & \second{93.18}
& 14.65 & 14.74 & 15.02
& 11.27 & 11.11 & \second{11.78} \\

& Fleming-38B
& \second{49.56} & 47.16 & 48.38
& 39.51 & 36.15 & 37.81
& 92.95 & 92.95 & 92.92
& 13.97 & 14.05 & 13.77
& 9.89 & 10.03 & 9.75 \\

\cmidrule(l){2-17}

\multirow{9}{*}{\rotatebox{90}{\scriptsize General}}
& Llama-3.2-11B
& 54.97 & 52.50 & 49.64
& 41.30 & 39.20 & 37.32
& 92.84 & 92.81 & 92.80
& 16.41 & 16.34 & 16.35
& \best{13.69} & \best{13.69} & \second{13.19} \\

& Phi-4-14B
& 41.67 & 40.68 & 41.12
& 32.47 & 31.46 & 32.02
& 93.33 & 93.33 & 93.33
& 17.40 & 17.48 & 17.33
& 12.31 & 12.46 & 12.22 \\

& DeepSeek-VL2-16B
& 44.10 & 43.05 & 42.35
& 34.15 & 33.37 & 32.65
& 93.26 & 93.28 & 93.32
& \second{17.65} & \best{18.04} & \best{18.46}
& 11.45 & 11.37 & 11.61 \\

& Mistral-Small-24B
& 4.49 & 6.02 & 5.00
& 3.05 & 4.04 & 3.30
& 87.97 & 88.09 & 87.86
& 9.08 & 9.06 & 9.03
& 6.66 & 6.88 & 6.84 \\

& DeepSeek-VL2-27B
& 44.98 & 45.35 & 44.48
& 33.96 & 34.19 & 33.53
& \second{93.48} & \second{93.47} & \second{93.44}
& \best{17.87} & \second{18.03} & \second{17.88}
& \second{13.30} & \second{13.39} & \best{13.28} \\

& Gemma-3-27B
& 50.25 & 49.84 & 50.82
& 39.14 & 38.81 & 40.36
& 93.28 & 93.28 & 93.36
& 17.21 & 17.28 & 17.31
& 11.10 & 11.59 & 11.40 \\

& Gemma-4-31B
& \second{59.40} & \best{59.54} & \best{59.62}
& \best{47.13} & \best{47.31} & \best{46.65}
& \best{93.55} & \best{93.58} & \best{93.49}
& 17.26 & 17.41 & 17.36
& 13.12 & 13.22 & 12.55 \\

& Qwen3-VL-32B
& \best{59.58} & \second{59.14} & \second{58.46}
& \second{45.55} & \second{45.10} & \second{44.95}
& 93.08 & 93.07 & 93.14
& 13.67 & 13.73 & 13.89
& 12.36 & 12.35 & 12.57 \\

& InternVL3.5-38B
& 20.34 & 20.05 & 21.34
& 13.62 & 13.36 & 14.31
& 93.26 & 93.30 & 93.27
& 13.89 & 13.91 & 13.91
& 7.96 & 8.04 & 7.94 \\

\midrule

\multicolumn{17}{l}{\textit{Closed-Source Models}} \\
\midrule

& GPT-5
& 46.34 & \second{48.65} & 46.53
& 34.30 & \second{37.05} & 35.13
& \second{93.06} & \second{93.04} & \second{93.04}
& \best{15.64} & \second{15.38} & \second{15.44}
& \second{12.56} & \second{12.54} & \second{12.53} \\

& Gemini-3.1-Pro
& \second{46.60} & 48.13 & \second{48.16}
& \second{35.88} & 36.91 & \second{37.46}
& \best{93.12} & \best{93.09} & \best{93.26}
& \second{15.59} & \best{15.50} & \best{16.84}
& 11.47 & 11.58 & 11.75 \\

& Claude-Opus-4.6
& \best{50.37} & \best{52.11} & \best{51.33}
& \best{40.70} & \best{42.28} & \best{42.56}
& 92.71 & 92.75 & 92.87
& 12.51 & 12.96 & 13.66
& \best{13.83} & \best{13.51} & \best{14.18} \\

\bottomrule
\end{tabular}
}
\par\smallskip
\scriptsize\textsuperscript{\textdagger}\textit{ Open-source: full 585-patient cohort. Closed-source: 100-patient public subset.}
\end{table*}

\paragraph{Clinical correctness and text quality decouple in report generation.}
The five metrics in Table~\ref{tab:report_generation} probe two distinct aspects of report quality. Two observations stand out.
First, Clinical BERTScore is approximately 92--94\% for most models and remains around 88\% even for Mistral-Small-24B, whose Clinical Findings F1 is only about 5\%. This contrast shows that high contextual similarity can coexist with substantial clinical errors.
Second, the rankings induced by clinical-factual and surface-form metrics are not concordant:
under MIL selection, MedGemma-4B achieves a ROUGE-L of 19.27\% but only 48.25\% Clinical Findings F1, whereas Qwen3-VL-32B obtains a lower ROUGE-L of 13.89\% but a higher Clinical Findings F1 of 58.46\%.
Together, these observations show that neither high semantic similarity nor greater lexical overlap is sufficient to establish clinical faithfulness.

\paragraph{A clear ceiling on clinical reporting.}
Even the strongest models cap at $\approx$60\% Clinical Findings F1 and $\approx$47\% Checklist Graph-F1.
The lower Graph-F1 highlights the difficulty of accurately linking abnormalities to their anatomical locations, tissue layers, and spatial distribution patterns, suggesting that recognizing clinical findings does not necessarily translate into precise anatomical characterization.
Several mid- and large-sized general models (InternVL3.5-8B, InternVL3.5-38B, Mistral-Small-24B)
exhibit catastrophic behavior on clinical-factual metrics (4--21\% Clinical Findings F1), despite competitive Level-1 image-understanding scores (Table~\ref{tab:image_understanding}). This discrepancy points to difficulties in producing faithful structured reports, including possible failures to follow the required format.

\subsection{Disease Diagnosis}
\label{app:detailed_diagnosis_results}

\begin{table*}[htbp]
\centering
\small
\setlength{\tabcolsep}{2.8pt}
\renewcommand{\arraystretch}{1.15}
\caption{\textbf{Detailed disease diagnosis performance (Level 3).} Accuracy, Precision, Recall, and F1 (\%) under three slice-selection strategies, averaged over the etiology classification and NICM subtyping tasks.}
\label{tab:diagnosis}

\newcommand{\best}[1]{\textbf{#1}}
\newcommand{\second}[1]{\underline{#1}}

\resizebox{1.0\textwidth}{!}{
\begin{tabular}{llcccccccccccc}
\toprule
& \textbf{Model} &
\multicolumn{3}{c}{\textbf{Accuracy (\%)}} &
\multicolumn{3}{c}{\textbf{Precision (\%)}} &
\multicolumn{3}{c}{\textbf{Recall (\%)}} &
\multicolumn{3}{c}{\textbf{F1 Score (\%)}} \\

\cmidrule(lr){3-5}
\cmidrule(lr){6-8}
\cmidrule(lr){9-11}
\cmidrule(lr){12-14}

& & Random & Heuristic & MIL
& Random & Heuristic & MIL
& Random & Heuristic & MIL
& Random & Heuristic & MIL \\

\midrule

\multicolumn{14}{l}{\textit{Open-Source Models ($\leq$10B)}} \\
\midrule

\multirow{5}{*}{\rotatebox{90}{\scriptsize Medical}}
& MedGemma-4B
& \second{41.71} & 42.38 & 36.87
& 21.56 & 22.16 & 20.81
& 21.23 & 21.15 & 20.65
& 19.73 & 18.73 & 20.31 \\

& Lingshu-7B
& 38.84 & 39.74 & \second{38.72}
& \second{23.04} & \second{23.50} & \second{24.01}
& \second{22.05} & \second{22.37} & \best{22.70}
& \second{22.35} & \second{22.75} & \best{22.84} \\

& Fleming-8B
& 39.35 & \second{42.50} & 26.97
& \best{24.60} & \best{25.89} & 23.09
& \best{22.48} & \best{23.15} & 18.34
& \best{23.31} & \best{23.99} & 16.80 \\

& HuluMed-7B
& \best{43.28} & \best{43.33} & \best{43.44}
& 15.55 & 15.57 & \best{35.58}
& 21.21 & 21.24 & \second{21.34}
& 17.94 & 17.96 & 18.21 \\

& QoQMed-7B
& 36.32 & 41.39 & 35.57
& 20.10 & 21.84 & 21.09
& 20.21 & 21.66 & 20.70
& 20.03 & 20.88 & \second{20.48} \\

\cmidrule(l){2-14}

\multirow{3}{*}{\rotatebox{90}{\scriptsize General}}
& Gemma-4-4B
& \best{43.24} & \second{43.33} & \best{43.24}
& 15.55 & 15.57 & \second{20.03}
& \second{21.20} & \second{21.24} & \second{21.30}
& 17.94 & 17.96 & \second{18.37} \\

& InternVL3.5-8B
& \second{43.16} & 43.21 & \second{43.16}
& \second{15.99} & \second{15.93} & 15.99
& 21.16 & 21.19 & 21.16
& \second{18.22} & \second{18.19} & 18.22 \\

& Qwen3-VL-8B
& \second{43.16} & \best{43.40} & 42.65
& \best{26.00} & \best{27.49} & \best{24.38}
& \best{21.24} & \best{21.45} & \best{21.98}
& \best{18.50} & \best{18.97} & \best{21.02} \\

\midrule

\multicolumn{14}{l}{\textit{Open-Source Models ($>$10B)}} \\
\midrule

\multirow{4}{*}{\rotatebox{90}{\scriptsize Medical}}
& MedGemma-27B
& 21.03 & 23.85 & 21.46
& 21.67 & 23.49 & 21.90
& 16.16 & 17.20 & 16.42
& 11.73 & 14.24 & 12.02 \\

& Hulu-30B
& \second{44.58} & \second{45.22} & \second{44.03}
& \best{30.40} & \best{35.89} & \best{25.93}
& 21.74 & 21.95 & 22.17
& 18.66 & 18.68 & 20.44 \\

& Lingshu-32B
& \best{47.29} & \best{46.52} & \best{46.27}
& 25.78 & 26.35 & \second{25.34}
& \best{24.27} & \second{23.56} & \best{24.21}
& \second{23.48} & \second{22.60} & \best{23.54} \\

& Fleming-38B
& 43.83 & 43.80 & 39.66
& \second{27.00} & \second{26.75} & 24.88
& \second{23.98} & \best{23.66} & \second{22.95}
& \best{24.79} & \best{24.32} & \second{23.44} \\

\cmidrule(l){2-14}

\multirow{7}{*}{\rotatebox{90}{\scriptsize General}}
& Llama-3.2-11B
& 24.69 & 24.68 & 24.69
& 12.15 & 4.95 & 10.59
& 20.02 & 20.00 & 20.01
& 8.08 & 7.94 & 8.44 \\

& Phi-4-14B
& 25.79 & 26.21 & 24.65
& 8.68 & 8.52 & 7.41
& 19.34 & 19.69 & 18.43
& 10.70 & 10.65 & 9.78 \\

& Mistral-Small-24B
& 17.80 & 16.11 & 18.86
& 8.76 & 10.96 & 8.77
& 5.99 & 6.06 & 5.82
& 4.69 & 4.56 & 4.57 \\

& Gemma-3-27B
& 42.57 & 41.09 & 38.05
& 19.65 & 20.01 & 20.03
& 21.06 & 20.92 & 20.43
& 18.30 & 19.43 & 19.91 \\

& Gemma-4-31B
& \best{51.42} & \best{51.24} & \best{49.80}
& \best{31.39} & \best{32.71} & \best{29.54}
& \best{29.71} & \best{29.28} & \best{28.64}
& \best{28.07} & \best{27.43} & \best{27.48} \\

& Qwen3-VL-32B
& \second{43.00} & \second{42.38} & \second{40.76}
& 21.07 & 21.20 & 21.65
& 21.22 & 21.25 & 21.52
& 18.35 & 19.15 & 20.63 \\

& InternVL3.5-38B
& 42.92 & 41.20 & 38.88
& \second{23.61} & \second{22.93} & \second{22.99}
& \second{23.05} & \second{22.81} & \second{22.58}
& \second{22.80} & \second{22.73} & \second{22.31} \\

\midrule

\multicolumn{14}{l}{\textit{Closed-Source Models}} \\
\midrule

& GPT-5
& 27.91 & 30.77 & 28.24
& 20.06 & \second{23.93} & 18.69
& \best{38.54} & 20.23 & 17.87
& 16.10 & 15.24 & 16.56 \\

& Gemini-3.1-Pro
& \second{39.83} & \second{42.50} & \second{32.77}
& \second{23.53} & \best{25.98} & \second{19.29}
& 25.04 & \best{25.73} & \second{19.34}
& \second{23.82} & \best{25.67} & \second{18.48} \\

& Claude-Opus-4.6
& \best{43.51} & \best{38.93} & \best{45.80}
& \best{25.30} & 21.41 & \best{26.34}
& \second{25.83} & \second{22.94} & \best{27.20}
& \best{25.45} & \second{22.12} & \best{26.73} \\

\bottomrule
\end{tabular}
}
\vspace{1mm}
\begin{minipage}{0.98\textwidth}
\footnotesize
\textit{Note.} DeepSeek-VL2 models are excluded from the diagnosis evaluation because their maximum sequence length is limited to 4096 tokens, which is insufficient to accommodate the patient-level diagnosis with 32 images input together with the textual prompt and output budget; see \href{https://huggingface.co/deepseek-ai/deepseek-vl2}{DeepSeek-VL2}.  Open-source: full 585-patient cohort. Closed-source: 100-patient public subset.
\end{minipage}

\end{table*}

\paragraph{Larger models do not consistently improve diagnosis.}
The effect of model size varies across families in Table~\ref{tab:diagnosis}. Under random selection, Lingshu improves only slightly from 7B to 32B (22.35 to 23.48 F1), while MedGemma declines from 4B to 27B (19.73 to 11.73). Gemma-4 improves more substantially from 4B to 31B (17.94 to 28.07), but its diagnosis performance remains limited. These comparisons show no uniform benefit from larger model variants and do not support parameter count alone as a solution to the diagnosis gap.

\paragraph{Aggregate accuracy can conceal poor disease discrimination.}
Several models obtain moderate accuracy despite low macro-F1 in Table~\ref{tab:diagnosis}. Under random selection, HuluMed-7B reaches 43.28\% accuracy but only 17.94\% macro-F1; Gemma-4-4B and InternVL3.5-8B show similar gaps. Accuracy gives more weight to common diseases through their larger case counts, whereas macro-F1 weights each disease category equally. The low macro-F1 scores therefore show that these accuracy levels do not reflect reliable performance across disease categories. The confusion analysis of Gemma-4-31B in Section~\ref{sec:main_results} illustrates how this can occur: predictions concentrate on NICM and HCM, allowing the model to correctly classify many cases from these common categories while misclassifying cases from other categories.

\paragraph{Slice selection has limited leverage at higher tiers.}
Consistent with the Level-1 finding in Section~\ref{sec:bottleneck_analysis},
slice-selection effects at Levels~2--3 are generally modest:
for most models in Table~\ref{tab:report_generation}, the spread of Clinical Findings F1 across the three strategies is below 3 points.
Diagnosis F1 in Table~\ref{tab:diagnosis} also varies modestly for most models, with larger exceptions such as Fleming-8B
(23.31 under Random versus 16.80 under MIL).
MIL selection can benefit individual models
(e.g., Qwen3-VL-8B gains $\approx$2.5 F1 points over Random selection on diagnosis),
but the higher-tier performance gap persists.

\subsection{Locating the Image-to-Diagnosis Bottleneck}
\label{app:diagnostic_controls}

{Level-3 diagnosis maps multi-sequence CMR images directly to patient-level disease categories. To better understand the sources of diagnostic failures, we conduct three complementary control experiments examining single- versus multi-sequence input, key-sequence removal, and diagnosis from oracle structured reports.}

\subsubsection{Single-sequence versus multi-sequence input}
\label{app:single_sequence_control}

Restricting a model to one sequence removes the need to combine evidence across sequences, but also removes potentially useful diagnostic information. We therefore use this comparison to examine whether the model struggles with the same patient-level diagnosis task under single-sequence input and how much it benefits from additional complementary sequences.

We evaluate Gemma-4-31B using the Level-3 heuristic protocol. The diagnostic tasks and labels remain fixed, while the visual input contains either Cine, LGE, T2, or Perfusion alone, or all four sequences together. The per-sequence inputs follow the selection procedure in Appendix~\ref{sec:slice-selection-across-task-tiers}. The all-sequence condition corresponds to the heuristic baseline in Table~\ref{tab:diagnosis}.

\begin{table}[htbp]
\centering
\small

\caption{Single-sequence versus multi-sequence Level-3 diagnosis with Gemma-4-31B under heuristic selection. The diagnosis tasks are unchanged across input conditions. Accuracy and macro-F1 are percentages.}
\label{tab:single_sequence_control}
\begin{tabular}{lrr}
\toprule
Visual input & Accuracy & Macro-F1 \\
\midrule
Perfusion only & 34.71 & 19.62 \\
T2 only & 39.08 & 22.03 \\
Cine only & 45.45 & 26.21 \\
LGE only & 48.29 & 25.88 \\
All four sequences & 51.24 & 27.43 \\
\bottomrule
\end{tabular}
\end{table}

All four sequences give the highest accuracy and macro-F1 (Table~\ref{tab:single_sequence_control}), but the gains over the strongest single-sequence conditions are modest. LGE provides the highest single-sequence accuracy, 48.29\%, compared with 51.24\% for all four sequences, a gain of 2.95 percentage points. Cine provides the highest single-sequence macro-F1, 26.21\%, compared with 27.43\% for all sequences, a gain of 1.22 points. These results suggest that cross-sequence integration is not the only source of Level-3 difficulty: performance remains limited even when the model must map a single sequence to a patient-level diagnosis. At the same time, the modest gains from combining four clinically complementary sequences suggest that the model does not fully exploit the additional evidence. Restricting the input to one sequence also removes diagnostically relevant information, so this comparison cannot completely separate task difficulty from cross-sequence integration. Nevertheless, the results point to difficulties in both translating sequence-specific visual evidence into diagnostic decisions and integrating complementary findings across sequences.

\subsubsection{Key-sequence removal in disease diagnosis}
\label{app:key_sequence_control}

To examine whether the category-collapse behavior observed in Level-3 diagnosis is associated with insufficient reliance on category-relevant CMR evidence, we perform a key-sequence ablation. Specifically, we remove LGE for etiology classification, T2 for myocarditis, and Cine for the remaining NICM subtypes, while keeping all other inputs unchanged. The removed sequence is determined by the ground-truth diagnosis rather than the model's prediction. If diagnostic predictions strongly depend on these visual cues, removing them should lead to a noticeable performance decline.

\begin{table}[h]
\centering
\small

\caption{Level-3 diagnosis after removing the key sequence for each diagnostic group under heuristic selection. Accuracy and macro-F1 are percentages; $\Delta$ is ablated minus the four-sequence baseline in percentage points.}
\label{tab:key_sequence_control}
\setlength{\tabcolsep}{9pt}
\renewcommand{\arraystretch}{1.22}
{\footnotesize
\begin{tabular}{@{}llr@{\hspace{1.5em}}lr@{}}
\toprule
& \multicolumn{2}{c}{Accuracy} & \multicolumn{2}{c}{Macro-F1} \\
\cmidrule(lr){2-3}\cmidrule(l){4-5}
Model & Four-seq. $\rightarrow$ Ablated & $\Delta$ & Four-seq. $\rightarrow$ Ablated & $\Delta$ \\
\midrule
Gemma-4-31B      & 51.24 $\rightarrow$ 48.82 & $-2.42$ & 27.43 $\rightarrow$ 25.36 & $-2.07$ \\
InternVL3.5-38B  & 41.20 $\rightarrow$ 38.51 & $-2.69$ & 22.73 $\rightarrow$ 21.41 & $-1.32$ \\
Lingshu-7B       & 39.74 $\rightarrow$ 33.65 & $-6.09$ & 22.75 $\rightarrow$ 19.96 & $-2.79$ \\
\bottomrule
\end{tabular}
}
\end{table}

All three models perform worse after the relevant sequence is removed (Table~\ref{tab:key_sequence_control}). Accuracy decreases by 2.42--6.09 points and macro-F1 by 1.32--2.79 points. The relatively modest F1 decreases suggest limited
sensitivity to the removal of category-relevant sequences,
although remaining sequences may still provide
complementary diagnostic evidence. These results indicate
some reliance on the removed sequences, but their
individual contributions cannot be fully isolated
through this ablation.

\subsubsection{Diagnosis from oracle structured reports}
\label{app:oracle_report_control}

To examine whether the Level-3 degradation arises primarily from the final diagnostic decision or from earlier failures in extracting and organizing visual evidence, we conduct an oracle-report experiment. We replace the CMR images with ground-truth structured radiology reports while keeping the Level-3 diagnosis task unchanged. The reports provide the structured clinical findings but exclude the final diagnosis name. We evaluate Gemma-4-31B, InternVL3.5-38B, and Lingshu-7B against their respective image-input baselines.

\begin{table}[t]
\centering
\small

\caption{Level-3 diagnosis from images or oracle structured reports, averaged over etiology classification and NICM subtyping. Accuracy and macro-F1 are percentages; $\Delta$ is oracle minus image input in percentage points.}
\label{tab:oracle_report_control}
\setlength{\tabcolsep}{5pt}
\begin{tabular}{lrrrrrr}
\toprule
& \multicolumn{3}{c}{Accuracy} & \multicolumn{3}{c}{Macro-F1} \\
\cmidrule(lr){2-4}\cmidrule(lr){5-7}
Model & Images & Oracle & $\Delta$ & Images & Oracle & $\Delta$ \\
\midrule
Gemma-4-31B & 51.24 & 84.42 & +33.18 & 27.43 & 51.01 & +23.58 \\
InternVL3.5-38B & 41.20 & 68.60 & +27.40 & 22.73 & 44.98 & +22.25 \\
Lingshu-7B & 39.74 & 68.48 & +28.74 & 22.75 & 44.00 & +21.25 \\
\bottomrule
\end{tabular}
\end{table}

All three models improve substantially when the structured clinical findings are directly supplied (Table~\ref{tab:oracle_report_control}), with accuracy gains of 27.40--33.18 percentage points and macro-F1 gains of 21.25--23.58 points. The improvement occurs without changing the diagnostic targets, indicating that the difficulty of making the final disease decision does not by itself explain the poor image-based results.

These results suggest that upstream failures in visual perception, finding extraction, and sequence-level synthesis contribute substantially to Level-3 degradation. While oracle structured reports make the final diagnostic decision considerably easier, constructing such representations from CMR images requires not only visual perception but also medical knowledge and clinical reasoning to interpret and integrate the evidence. Thus, a major bottleneck appears to lie in the image-to-clinical transformation, rather than solely in diagnosis from already-structured findings.

\clearpage

\section{Public Subset Details and Consistency}
\label{app:public_subset}
\subsection{Public Subset Statistics}
\label{app:subset_statistics}

\begin{table}[htbp]
\centering
\caption{CardioLens public subset statistics: 2,297 QA pairs from 100 patients. Image understanding subtasks are single-label (S) or multi-label (M) with different classes(\#C).}
\label{tab:subset_statistics}
\setlength{\tabcolsep}{4pt}
\renewcommand{\arraystretch}{0.95}
\resizebox{1.0\textwidth}{!}{
\footnotesize
\begin{tabular}{@{}lcrr@{\hskip 10pt}lcrr@{\hskip 10pt}lcrr@{\hskip 10pt}lcrr@{}}
\toprule
\multicolumn{4}{c}{\textbf{Cine (889)}} & \multicolumn{4}{c}{\textbf{LGE (549)}} & \multicolumn{4}{c}{\textbf{Perfusion (190)}} & \multicolumn{4}{c}{\textbf{T2 (177)}} \\
\cmidrule(r){1-4}\cmidrule(lr){5-8}\cmidrule(lr){9-12}\cmidrule(l){13-16}
Subtask & T & \#C & \#QA & Subtask & T & \#C & \#QA & Subtask & T & \#C & \#QA & Subtask & T & \#C & \#QA \\
\midrule
Wall Thickness      & M & 5 &  99 & Enhance. Status  & S & 2 &  98 & Perf.\ Status       & S & 2 &  96 & T2 Signal        & S & 4 &  91 \\
Special Signs       & M & 6 &  99 & Special Findings    & M & 4 &  99 & Signal Char.        & S & 3 &  29 & Abn.\ Regions    & M & 6 &  24 \\
Pericardial Eff.    & S & 2 & 100 & Abn.\ Signal        & S & 2 &  79 & Abn.\ Regions       & M & 6 &  29 & Abn.\ Segments   & M & 4 &  24 \\
Pleural Eff.        & S & 2 & 100 & High-Sig Pattern      & M & 5 &  66 & Abn.\ Segments      & M & 4 &  19 & Sig.\ Distrib.   & M & 5 &  25 \\
Mitral Valve        & S & 2 & 100 & High-Sig Layer        & M & 5 &  57 & Myo.\ Layer     & M & 5 &  17 & Myo.\ Layer  & M & 5 &  13 \\
Tricuspid Valve     & S & 2 & 100 & High-Sig Region       & M & 6 &  65 &                     &   &   &     &                  &   &   &     \\
Aortic Valve        & S & 2 &  99 & High-Sig Segment      & M & 5 &  63 &                     &   &   &     &                  &   &   &     \\
Motion Ampl.        & M & 5 &  71 & Low-Sig Pattern      & M & 5 &   9 &                     &   &   &     &                  &   &   &     \\
Systolic Func.      & S & 3 &  61 & Low-Sig Region       & M & 6 &   5 &                     &   &   &     &                  &   &   &     \\
Motion Coord.       & S & 2 &  36 & Low-Sig Segment      & M & 4 &   5 &                     &   &   &     &                  &   &   &     \\
Diastolic Func.     & S & 2 &  24 & Low-Sig Layer        & M & 5 &   3 &                     &   &   &     &                  &   &   &     \\
\midrule
\multicolumn{16}{@{}l}{\textbf{Report Generation (361):}~~Cine 99~~$\mid$~~LGE 99~~$\mid$~~Perfusion 96~~$\mid$~~T2 67} \\
\midrule
\multicolumn{16}{@{}l}{\textbf{Diagnosis (131):}~~\textit{Etiology (98):} NH 20 / IHD 45 / NICM 33~~$\mid$~~\textit{NICM Subtype (33):} HCM 20 / Myocarditis 6 / DCM 5 / RCM 1 / ACM 1} \\
\bottomrule
\end{tabular}
}
\end{table}

The 100-patient public subset comprises 2,297 verified QA
pairs, with the underlying CMR studies containing a total
of 78,707 image slices. Detailed statistics are provided
in Table~\ref{tab:subset_statistics}. All four imaging sequences (Cine, LGE, Perfusion, T2) and
all 32 image-understanding subtasks are retained, with both single- and
multi-label formulations preserved. The three task tiers contain 1{,}805 image-understanding QA pairs, 361 reports, and 131 diagnosis QA items, and the diagnosis hierarchy preserves all
three etiology classes (NH/IHD/NICM) and all five NICM subtypes
(HCM/DCM/RCM/ACM/Myocarditis).

\subsection{Consistency Between the Full Cohort and Public Subset}
\label{app:subset_consistency}

We compare the public subset with the full cohort to examine whether the main evaluation patterns hold across both cohorts.
Table~\ref{tab:subset_full_consistency_all}(a)--(c) reports the absolute
gap $|\Delta|$ between full-cohort and subset performance for three
representative open-source models across the three evaluation levels.

\paragraph{Level 1 and Level 2 show strong subset-to-full consistency.}
For image understanding (panel~a), gaps are uniformly small: across all
9 model$\times$strategy comparisons, the mean $|\Delta|$
on F1 is 1.15 points and the maximum is 1.79 (InternVL3.5-38B, Random);
accuracy and precision gaps are similarly bounded.
Report generation (panel~b) shows similarly small gaps: Clinical BERTScore differs by at most 0.05 points and Clinical Findings F1 by at most 1.43 points across the nine model--strategy comparisons. Two comparative patterns are also preserved: (i)~among the three models, the Level-1 F1 and Level-2 Clinical Findings F1 rankings are unchanged---Lingshu-7B leads on image understanding and Gemma-4-31B on report generation; and (ii)~for these metrics, the direction of the Heuristic and MIL changes relative to Random is the same on both cohorts. The public subset therefore reproduces the main Level-1 and Level-2 comparisons for the tested models.

\begin{figure*}[htbp]
    \centering

    \begin{minipage}{0.49\textwidth}
        \centering
        \includegraphics[width=\linewidth]{Figures/confusion_matrix_combined_main_category_sub_category.pdf}
    \end{minipage}
    \hfill
    \begin{minipage}{0.49\textwidth}
        \centering
        \includegraphics[width=\linewidth]{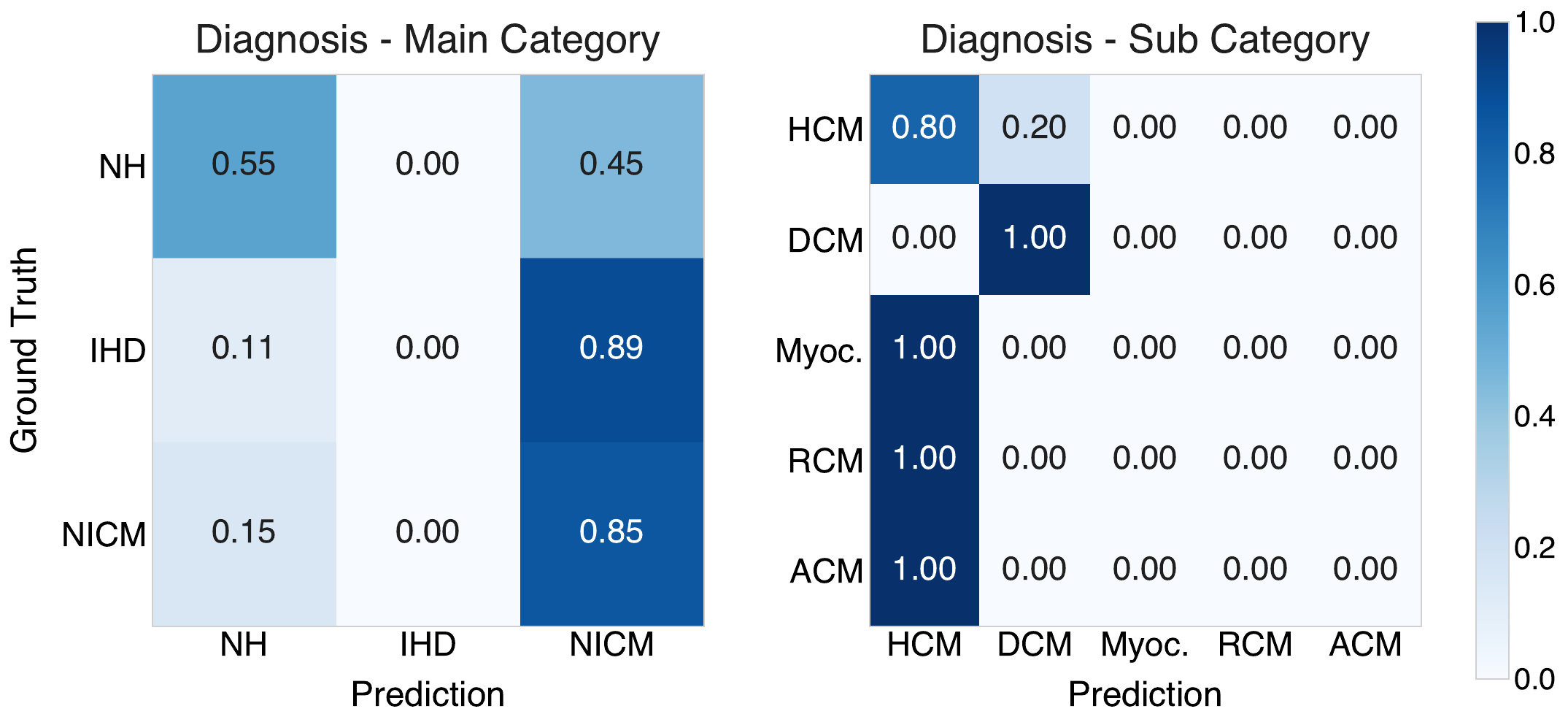}
    \end{minipage}

    \caption{\textbf{Full v.s. Sub cohort performance on Level-3 task for Gemma-4-31B (random sampling).}
 {Left}: Confusion matrices for etiology classification and NICM subtyping on \textbf{full cohort}. \textbf{Right}: Confusion matrices for etiology classification and NICM subtyping on \textbf{public subset}. Predictions in both cohorts concentrate on NICM at the etiology level and HCM at the subtype level.}

    \label{fig:subset_diagnosis}
\end{figure*}

\paragraph{Level-3 Failure Patterns Remain Consistent Across Cohorts.}
The full-to-subset differences in diagnosis F1 are larger than those at Levels~1--2, reaching 7.65 points (Table~\ref{tab:subset_full_consistency_all}(c)).
At the etiology level, NICM accounts for 53.3\% (295/553) of the full cohort but only 33.7\% (33/98) of the subset, where IHD instead becomes the largest class (45.9\%).
Despite this shift, Gemma-4-31B exhibits similar confusion patterns under random sampling (Figure~\ref{fig:subset_diagnosis}): 89\% versus 88\% of IHD cases are classified as NICM, while 85\% versus 88\% of NICM cases are correctly classified (subset versus full). A similar concentration of predictions on HCM is observed in the NICM subtyping task.
With only one RCM and one ACM case in the public subset, fine-grained subtype scores are sensitive to individual predictions (Table~\ref{tab:subset_statistics}).
Together, these findings suggest that the larger Level-3 score differences partly reflect cohort composition, while the public subset reproduces the main category-collapse patterns observed on the full benchmark.

\begin{table*}[htbp]
\centering
\scriptsize
\renewcommand{\arraystretch}{1.10}
\caption{\textbf{Consistency between the full cohort and the 100-patient public subset across all three evaluation levels.} For three representative open-source models (Lingshu-7B, Gemma-4-31B, InternVL3.5-38B) we report performance under three slice-selection strategies (Random, Heuristic, MIL) on the full cohort (\textit{Full}) and on the public subset (\textit{Sub.}); $|\Delta|$ is the absolute gap. }
\label{tab:subset_full_consistency_all}

\textbf{(a) Level 1: Image Understanding}\\[1pt]
\setlength{\tabcolsep}{8pt}
\resizebox{1.0\textwidth}{!}{
\begin{tabular}{llcccccccccccc}
\toprule
 & & \multicolumn{3}{c}{\textbf{Accuracy (\%)}} & \multicolumn{3}{c}{\textbf{Precision (\%)}} & \multicolumn{3}{c}{\textbf{Recall (\%)}} & \multicolumn{3}{c}{\textbf{F1 Score (\%)}} \\
\cmidrule(lr){3-5}\cmidrule(lr){6-8}\cmidrule(lr){9-11}\cmidrule(lr){12-14}
\textbf{Model} & \textbf{Source} & Rand. & Heur. & MIL & Rand. & Heur. & MIL & Rand. & Heur. & MIL & Rand. & Heur. & MIL \\
\midrule
\multirow{3}{*}{Lingshu-7B} & Full       & 65.22 & 66.75 & 67.52 & 50.76 & 54.67 & 55.08 & 63.43 & 65.58 & 65.98 & 56.22 & 59.34 & 59.82 \\
                              & Sub.       & 66.71 & 68.53 & 68.74 & 52.30 & 56.27 & 56.94 & 64.52 & 66.76 & 66.75 & 57.56 & 60.70 & 61.16 \\
                              & $|\Delta|$ & 1.49 & 1.78 & 1.22 & 1.54 & 1.60 & 1.86 & 1.09 & 1.18 & 0.77 & 1.34 & 1.36 & 1.34 \\
\midrule
\multirow{3}{*}{Gemma-4-31B} & Full       & 61.83 & 61.53 & 62.56 & 46.62 & 46.47 & 47.75 & 57.86 & 57.26 & 58.12 & 51.61 & 51.29 & 52.40 \\
                              & Sub.       & 62.57 & 62.01 & 63.56 & 47.23 & 46.72 & 48.40 & 58.75 & 58.10 & 59.07 & 52.35 & 51.79 & 53.17 \\
                              & $|\Delta|$ & 0.74 & 0.48 & 1.00 & 0.61 & 0.25 & 0.65 & 0.89 & 0.84 & 0.95 & 0.74 & 0.50 & 0.77 \\
\midrule
\multirow{3}{*}{InternVL3.5-38B} & Full       & 63.97 & 64.72 & 65.29 & 51.09 & 52.67 & 53.73 & 60.27 & 62.46 & 60.98 & 55.25 & 57.09 & 57.08 \\
                              & Sub.       & 65.40 & 65.80 & 66.29 & 52.77 & 54.01 & 54.94 & 62.19 & 63.37 & 62.50 & 57.04 & 58.24 & 58.41 \\
                              & $|\Delta|$ & 1.43 & 1.08 & 1.00 & 1.68 & 1.34 & 1.21 & 1.92 & 0.91 & 1.52 & 1.79 & 1.15 & 1.33 \\
\bottomrule
\end{tabular}}

\vspace{2.0mm}

\textbf{(b) Level 2: Report Generation}\\[1pt]
\setlength{\tabcolsep}{4.0pt}
\resizebox{1.0\textwidth}{!}{
\begin{tabular}{llccccccccccccccc}
\toprule
 & & \multicolumn{3}{c}{\textbf{Clinical Findings F1 (\%)}} & \multicolumn{3}{c}{\textbf{Checklist Graph-F1 (\%)}} & \multicolumn{3}{c}{\textbf{Clinical BERTScore (\%)}} & \multicolumn{3}{c}{\textbf{ROUGE-L (\%)}} & \multicolumn{3}{c}{\textbf{CIDEr-lite (\%)}} \\
\cmidrule(lr){3-5}\cmidrule(lr){6-8}\cmidrule(lr){9-11}\cmidrule(lr){12-14}\cmidrule(lr){15-17}
\textbf{Model} & \textbf{Source} & Rand. & Heur. & MIL & Rand. & Heur. & MIL & Rand. & Heur. & MIL & Rand. & Heur. & MIL & Rand. & Heur. & MIL \\
\midrule
\multirow{3}{*}{Lingshu-7B} & Full       & 47.04 & 46.21 & 47.80 & 37.30 & 36.21 & 37.31 & 93.08 & 93.06 & 93.06 & 14.87 & 15.12 & 14.95 & 11.32 & 11.38 & 11.34 \\
                              & Sub.       & 46.56 & 45.48 & 47.91 & 36.00 & 34.89 & 36.26 & 93.04 & 93.03 & 93.02 & 15.06 & 15.35 & 15.22 & 12.16 & 11.97 & 11.99 \\
                              & $|\Delta|$ & 0.48 & 0.73 & 0.11 & 1.30 & 1.32 & 1.05 & 0.04 & 0.03 & 0.04 & 0.19 & 0.23 & 0.27 & 0.84 & 0.59 & 0.65 \\
\midrule
\multirow{3}{*}{Gemma-4-31B} & Full       & 59.40 & 59.54 & 59.62 & 47.13 & 47.31 & 46.65 & 93.55 & 93.58 & 93.49 & 17.26 & 17.41 & 17.36 & 13.12 & 13.22 & 12.55 \\
                              & Sub.       & 59.60 & 60.06 & 60.68 & 46.34 & 47.05 & 47.03 & 93.50 & 93.56 & 93.48 & 17.33 & 17.65 & 17.50 & 13.95 & 14.05 & 13.37 \\
                              & $|\Delta|$ & 0.20 & 0.52 & 1.06 & 0.79 & 0.26 & 0.38 & 0.05 & 0.02 & 0.01 & 0.07 & 0.24 & 0.14 & 0.83 & 0.83 & 0.82 \\
\midrule
\multirow{3}{*}{InternVL3.5-38B} & Full       & 20.34 & 20.05 & 21.34 & 13.62 & 13.36 & 14.31 & 93.26 & 93.30 & 93.27 & 13.89 & 13.91 & 13.91 & 7.96 & 8.04 & 7.94 \\
                              & Sub.       & 19.47 & 18.62 & 20.32 & 12.71 & 12.07 & 13.27 & 93.24 & 93.28 & 93.24 & 13.77 & 13.93 & 13.73 & 8.04 & 8.21 & 8.04 \\
                              & $|\Delta|$ & 0.87 & 1.43 & 1.02 & 0.91 & 1.29 & 1.04 & 0.02 & 0.02 & 0.03 & 0.12 & 0.02 & 0.18 & 0.08 & 0.17 & 0.10 \\
\bottomrule
\end{tabular}}

\vspace{2.0mm}

\textbf{(c) Level 3: Disease Diagnosis}\\[1pt]
\setlength{\tabcolsep}{8pt}
\resizebox{1.0\textwidth}{!}{
\begin{tabular}{llcccccccccccc}
\toprule
 & & \multicolumn{3}{c}{\textbf{Accuracy (\%)}} & \multicolumn{3}{c}{\textbf{Precision (\%)}} & \multicolumn{3}{c}{\textbf{Recall (\%)}} & \multicolumn{3}{c}{\textbf{F1 Score (\%)}} \\
\cmidrule(lr){3-5}\cmidrule(lr){6-8}\cmidrule(lr){9-11}\cmidrule(lr){12-14}
\textbf{Model} & \textbf{Source} & Rand. & Heur. & MIL & Rand. & Heur. & MIL & Rand. & Heur. & MIL & Rand. & Heur. & MIL \\
\midrule
\multirow{3}{*}{Lingshu-7B} & Full       & 38.84 & 39.74 & 38.72 & 23.04 & 23.50 & 24.01 & 22.05 & 22.37 & 22.70 & 22.35 & 22.75 & 22.84 \\
                              & Sub.       & 28.75 & 29.77 & 29.77 & 17.30 & 18.09 & 18.74 & 18.61 & 19.29 & 19.22 & 17.63 & 18.31 & 18.66 \\
                              & $|\Delta|$ & 10.09 & 9.97 & 8.95 & 5.74 & 5.41 & 5.27 & 3.44 & 3.08 & 3.48 & 4.72 & 4.44 & 4.18 \\
\midrule
\multirow{3}{*}{Gemma-4-31B} & Full       & 51.42 & 51.24 & 49.80 & 31.39 & 32.71 & 29.54 & 29.71 & 29.28 & 28.64 & 28.07 & 27.43 & 27.48 \\
                              & Sub.       & 45.04 & 42.75 & 41.73 & 32.00 & 30.68 & 25.47 & 29.41 & 27.97 & 27.40 & 25.24 & 23.39 & 22.94 \\
                              & $|\Delta|$ & 6.38 & 8.49 & 8.07 & 0.61 & 2.03 & 4.07 & 0.30 & 1.31 & 1.24 & 2.83 & 4.04 & 4.54 \\
\midrule
\multirow{3}{*}{InternVL3.5-38B} & Full       & 42.92 & 41.20 & 38.88 & 23.61 & 22.93 & 22.99 & 23.05 & 22.81 & 22.58 & 22.80 & 22.73 & 22.31 \\
                              & Sub.       & 29.77 & 31.30 & 29.77 & 15.71 & 16.53 & 17.05 & 19.38 & 20.36 & 19.26 & 15.15 & 17.77 & 17.51 \\
                              & $|\Delta|$ & 13.15 & 9.90 & 9.11 & 7.90 & 6.40 & 5.94 & 3.67 & 2.45 & 3.32 & 7.65 & 4.96 & 4.80 \\
\bottomrule
\end{tabular}}

\end{table*}

\clearpage

\section{Clinical References and Independent Reader Evaluation}
\label{app:clinical_references}

A natural question is how MLLMs compare with radiologists.
We consider two complementary references: a report-derived clinical reference based on finalized radiology reports and an independent radiologist evaluation under matched visual inputs.

\subsection{Report-derived Clinical Reference}
\label{app:radiologist_baseline}

The image-understanding labels are extracted and verified from radiologist-finalized CMR reports. Agreement with those reference labels is therefore $100\%$ by construction; this value is not an independently measured radiologist score. Each source report was drafted by a radiologist and reviewed and signed by a senior radiologist, as described in Section~\ref{sec:data_construction}.

For LGE and T2, larger input budgets allow us to compare model predictions with the report-derived findings under broader visual coverage.
Their typical per-study slice counts
($\approx$15--18 for LGE across SAX/4CH/2CH and
$\approx$10 for T2 SAX\footnote{
Clinical T2 assessment is primarily based on T2 SAX images.
T2 4CH images, whose slice/frame counts vary across studies,
are included in our input construction as auxiliary visual context.
Therefore, at larger budgets such as $K{=}32$ and $K{=}128$,
the additional inclusion of T2 4CH content may affect model
performance even though the T2 SAX view itself is already largely covered.
})
are typically covered by $K=32$ and $K=128$.

\begin{table}[htbp]
\centering
\caption{\textbf{Report-derived clinical reference for LGE and T2 image understanding.} F1 (\%) under random selection at $K\in\{8,32,128\}$. Larger budgets typically cover the LGE and T2 images described in the text. The final row is the highest model F1 minus 100\% reference-label agreement; it is not a measured reader--model gap.}
\label{tab:radiologist_baseline}
\small
\setlength{\tabcolsep}{6pt}
\begin{tabular}{lcccccc}
\toprule
\multirow{2}{*}{\textbf{Model}}
 & \multicolumn{3}{c}{\textbf{LGE F1 (\%)}}
 & \multicolumn{3}{c}{\textbf{T2 F1 (\%)}} \\
\cmidrule(lr){2-4}\cmidrule(lr){5-7}
 & $K{=}8$ & $K{=}32$ & $K{=}128$ & $K{=}8$ & $K{=}32$ & $K{=}128$ \\
\midrule
MedGemma-4B       & 51.7 & 50.8 & {50.8} & {57.5} & 55.2 & 54.9 \\
Lingshu-7B        & 56.8 & 54.4 & 54.4          & 50.8          & 46.0 & 45.2 \\
Gemma-4-31B       & 48.3 & 49.0 & 49.0          & 53.6          & 51.9 & 50.2 \\
Qwen3-VL-32B      & 49.5 & 47.8 & 47.7          & 50.1          & 44.1 & 39.9 \\
InternVL3.5-38B   & 54.8 & 56.0 & 56.0          & 44.9          & 44.2 & 44.5 \\
\midrule
\textit{Best $-$ Reference}
                  & $-43.2$ & $-44.0$ & $-44.0$
                  & $-42.5$ & $-44.8$ & $-45.1$ \\
\bottomrule
\end{tabular}
\end{table}

Table~\ref{tab:radiologist_baseline} reports this comparison for five representative models.
At larger slice budgets, the best LGE model reaches only $56.0$ F1,
leaving a $44.0$-point gap to the report-derived reference;
on T2, the best score across the tested budgets is $57.5$ F1,
leaving a $42.5$-point gap.
Moreover, performance changes little across $K \in \{8,32,128\}$,
and several models even decline on T2 as $K$ increases.
These results show that expanding the sampled visual coverage does not close the gap between model predictions and the report-derived findings, pointing to difficulties in interpreting the available CMR evidence.

\subsection{Independent Reader Evaluation}
\label{app:independent_reader}

We further conduct an independent reader study to assess radiologist performance under the same selected-image inputs provided to the compared models.

An independent radiologist, who was not involved in the original reporting, evaluates the 98 public cases with etiology labels: 20 NH, 45 IHD, and 33 NICM. For each anonymized case, the reader receives eight heuristic-selected images from each of Cine, LGE, T2, and Perfusion, giving 32 images in total. These are the same images supplied to the compared MLLMs. Patient information, clinical history, reports, discharge records, and ground-truth labels are withheld. The diagnoses are recorded independently and compared with the discharge diagnoses only after the reading is complete.

\begin{table}[htbp]
\centering
\small

\caption{Independent reader and MLLMs on the same 98 public etiology cases with heuristic-selected images. Scores are percentages. Distributions follow NH / IHD / NICM.}
\label{tab:independent_reader_control}
\begin{tabular}{lrrc}
\toprule
Reader / model & Accuracy & Macro-F1 & Prediction distribution \\
\midrule
Radiologist & 97.96 & 97.66 & 22 / 43 / 33 \\
Gemma-4-31B & 37.76 & 30.44 & 14 / 5 / 79 \\
InternVL3.5-38B & 36.73 & 29.21 & 37 / 0 / 61 \\
Qwen3-VL-32B & 34.69 & 23.84 & 9 / 0 / 89 \\
\midrule
Ground truth & -- & -- & 20 / 45 / 33 \\
\bottomrule
\end{tabular}
\end{table}

\paragraph{Results.}
The reader achieves 97.96\% accuracy and 97.66 macro-F1 (Table~\ref{tab:independent_reader_control}), correctly classifying 20/20 NH, 43/45 IHD, and 33/33 NICM cases. Both errors are IHD cases classified as NH. In contrast, all three MLLMs remain below 38\% accuracy. Their prediction distributions are concentrated on a subset of the categories, whereas the reader's distribution is close to the ground truth.

The independent radiologist achieved high accuracy on the NH/IHD/NICM etiology task using the same heuristic-selected CMR images provided to the models. For this task, LGE distribution and myocardial layer provide important evidence for distinguishing ischemic from non-ischemic conditions, complemented by functional and tissue information from other sequences. The strong reader performance indicates that the selected images contain sufficient diagnostic evidence for this distinction, highlighting the models' limitations in interpreting and integrating the available visual information.

\clearpage

\section{Visual Input and Enumeration Bias}
\label{app:visual_controls}

\label{app:image_blind_control}

The enumeration bias observed in medically adapted models
raises the question of whether medical adaptation weakens
visual grounding in favor of clinical priors.
To examine this, we compare Qwen2.5-VL-7B and its medically
adapted counterpart, Lingshu-7B, on identical Level-1
LGE questions with and without CMR images.
The image-blind condition retains the questions and answer
options to assess the contribution of visual evidence
to each model's predictions.

\begin{table}[!htbp]
\centering
\small

\caption{Image input versus image-blind Level-1 LGE. Accuracy, precision, recall, and F1 are percentages; $\Delta$ is image minus blind input in percentage points.}
\label{tab:image_blind_control}
\setlength{\tabcolsep}{8pt}
\renewcommand{\arraystretch}{1.16}
{\footnotesize
\begin{tabular}{@{}llrrrr@{}}
\toprule
Model & Input & Accuracy & Precision & Recall & F1 \\
\midrule
\multirow{3}{*}{Qwen2.5-VL-7B}
  & Image   & 40.88 & 36.01 & 67.73 & 47.02 \\
  & Blind   & 45.14 & 36.56 & 65.93 & 47.04 \\
  & $\Delta$ & $-4.26$ & $-0.55$ & $+1.80$ & $-0.02$ \\
\addlinespace[3pt]
\multirow{3}{*}{Lingshu-7B}
  & Image   & 62.58 & 50.26 & 69.60 & 58.37 \\
  & Blind   & 54.96 & 44.10 & 85.27 & 58.13 \\
  & $\Delta$ & $+7.62$ & $+6.16$ & $-15.67$ & $+0.24$ \\
\bottomrule
\end{tabular}
}
\end{table}

For Lingshu-7B, adding images improves accuracy by 7.62
percentage points and precision by 6.16 points, while
reducing recall by 15.67 points. Conversely, without
images, the model achieves substantially higher recall
at the cost of lower precision. This suggests that visual
evidence helps constrain the model's tendency to predict
plausible abnormalities, making its predictions more
selective. Thus, the observed enumeration bias does not
imply that the medically adapted model disregards
visual information.

Qwen2.5-VL-7B exhibits a different response to visual
input. Adding images slightly increases recall (+1.80
points) but reduces accuracy (-4.26 points) and precision
(-0.55 points). The accuracy decline is primarily
associated with the Abnormal Signal subtask, where image
input increases confusion between High Signal and Mixed
Signal categories. Unlike Lingshu-7B, the base model
does not show corresponding improvements in accuracy
or precision when images are provided.

Despite these differences, F1 remains nearly unchanged
for both models, as changes in precision and recall
largely offset each other. This highlights how similar
F1 scores can mask distinct responses to visual evidence.
Importantly, Lingshu-7B benefits more clearly from
visual input than its base counterpart, suggesting that
medical adaptation does not necessarily weaken visual
grounding. Instead, its high recall under image-blind
conditions points to a tendency to assert plausible
findings when visual evidence is unavailable.

\clearpage

\section{Task Adaptation and Interactive Evaluation}

Our main evaluation uses fixed sets of selected CMR images without task-specific adaptation. Here, we test three extensions: providing task-specific examples, lightweight fine-tuning, and allowing models to request additional sequences during diagnosis.

\subsection{Few-Shot Prompting and Lightweight Fine-Tuning}
\label{app:adaptation_controls}

We evaluate three-shot prompting for Level-3 diagnosis and QLoRA fine-tuning for Level-1 image understanding to assess the effectiveness of in-context examples and lightweight parameter adaptation.

\label{app:additional_evaluation}
\begin{table}[!htbp]
\centering
\small
\caption{Few-shot prompting and QLoRA adaptation on the public subset. Values are percentages; $\Delta$ is in percentage points.}
\label{tab:adaptation_summary}
\captionsetup[subtable]{font=small,labelfont=bf,justification=raggedright,singlelinecheck=false,skip=3pt}
\begin{subtable}{\linewidth}
\centering
\caption{Three-shot Level-3 diagnosis ($\Delta$ = three-shot $-$ zero-shot).}
\label{tab:fewshot_control}
\setlength{\tabcolsep}{9pt}
\renewcommand{\arraystretch}{1.15}
{\footnotesize
\begin{tabular}{@{}llr@{\hspace{1.5em}}lr@{}}
\toprule
& \multicolumn{2}{c}{Accuracy} & \multicolumn{2}{c}{Macro-F1} \\
\cmidrule(lr){2-3}\cmidrule(l){4-5}
Model & 0-shot $\rightarrow$ 3-shot & $\Delta$ & 0-shot $\rightarrow$ 3-shot & $\Delta$ \\
\midrule
Gemma-4-31B     & 42.75 $\rightarrow$ 36.64 & $-6.11$ & 23.39 $\rightarrow$ 20.36 & $-3.03$ \\
InternVL3.5-38B & 31.30 $\rightarrow$ 29.77 & $-1.53$ & 17.77 $\rightarrow$ 13.28 & $-4.49$ \\
Lingshu-7B      & 29.77 $\rightarrow$ 31.30 & $+1.53$ & 18.31 $\rightarrow$ 18.56 & $+0.25$ \\
\bottomrule
\end{tabular}
}
\end{subtable}
\par\vspace{0.7em}
\begin{subtable}{\linewidth}
\centering
\caption{Gemma-4-31B: Level-1 QLoRA adaptation ($\Delta$ = adapted $-$ zero-shot).}
\label{tab:qlora_control}
\setlength{\tabcolsep}{10pt}
\renewcommand{\arraystretch}{1.18}
{\footnotesize
\begin{tabular}{@{}lrrrr@{}}
\toprule
Setting & Accuracy & Precision & Recall & F1 \\
\midrule
Zero-shot & 62.01 & 46.72 & 58.10 & 51.79 \\
QLoRA     & 52.00 & 47.93 & 65.97 & 55.52 \\
$\Delta$ & $-10.01$ & $+1.21$ & $+7.87$ & $+3.73$ \\
\bottomrule
\end{tabular}
}
\end{subtable}
\end{table}

\subsubsection{Three-shot prompting for diagnosis}
\label{app:fewshot_control}

For each Level-3 test case in the 100-patient public subset, we prepend three solved examples drawn from other CardioLens patients. The demonstrations exclude the test patient and are sampled to balance answer classes. We compare zero-shot and three-shot performance for Gemma-4-31B, InternVL3.5-38B, and Lingshu-7B.

Three-shot prompting produces no consistent gains (Table~\ref{tab:fewshot_control}). Gemma-4-31B loses 6.11 accuracy points and 3.03 macro-F1 points, while InternVL3.5-38B loses 1.53 and 4.49 points, respectively. Lingshu-7B gains 1.53 accuracy points but only 0.25 macro-F1 points. These results suggest that few-shot prompting does not consistently improve diagnostic performance, despite providing labeled examples of the task.

\subsubsection{QLoRA adaptation for image understanding}
\label{app:qlora_control}

We additionally fine-tune Gemma-4-31B on Level-1 image-understanding questions from patients outside the public subset and evaluate on the public subset using heuristic selection. The adaptation targets Level~1 rather than Level~3, for which only 848 diagnosis QA items are available. We use 4-bit NF4 quantization and LoRA rank 16, with alpha 32 and dropout 0.05, applied to language-model projections while freezing the vision encoder.

QLoRA improves F1 from 51.79 to 55.52, a gain of 3.73 points (Table~\ref{tab:qlora_control}). This is mainly associated with higher recall (58.10 to 65.97), whereas precision increases only slightly (46.72 to 47.93) and accuracy decreases from 62.01\% to 52.00\%. Consequently, the recall--precision gap widens from 11.38 to 18.04 points. The model identifies more true findings, but this improvement is not accompanied by a comparable gain in precision or accuracy.

These results are consistent with report-derived supervision shifting the model's tendency to assert abnormalities, rather than establishing improved visual grounding or diagnostic discrimination. Taken together, the three-shot and QLoRA controls show that few-shot prompting and lightweight adaptation do not uniformly improve clinical performance: the former provides no reliable diagnosis benefit, while the latter trades lower accuracy for greater finding coverage.

\subsection{Interactive Sequence Acquisition}
\label{app:interactive_control}

To examine whether models can determine when additional CMR
evidence is needed, we simulate interactive sequence access
during Level-3 diagnosis. Starting with Cine, each model
may request LGE, T2, or Perfusion over up to three rounds
before making a diagnosis. We evaluate Gemma-4-31B and
InternVL3.5-38B against the static four-sequence baseline,
examining their diagnostic performance, sequence request
patterns, and self-reported confidence.

\begin{table}[t]
\centering
\small
\caption{Interactive sequence acquisition: diagnostic performance, sequence requests, and confidence. Performance and request rates are percentages; $\Delta$ is in percentage points.}
\label{tab:interactive_summary}
\captionsetup[subtable]{font=small,labelfont=bf,justification=raggedright,singlelinecheck=false,skip=3pt}

\begin{subtable}{\linewidth}
\centering
\caption{Diagnostic performance compared with static four-sequence input.}
\label{tab:interactive_performance}
\setlength{\tabcolsep}{9pt}
\renewcommand{\arraystretch}{1.15}
{\footnotesize
\begin{tabular}{@{}llr@{\hspace{1.5em}}lr@{}}
\toprule
& \multicolumn{2}{c}{Accuracy} & \multicolumn{2}{c}{Macro-F1} \\
\cmidrule(lr){2-3}\cmidrule(l){4-5}
Model & Static $\rightarrow$ Interactive & $\Delta$ & Static $\rightarrow$ Interactive & $\Delta$ \\
\midrule
Gemma-4-31B & 51.24 $\rightarrow$ 40.26 & $-10.98$ & 27.43 $\rightarrow$ 24.41 & $-3.02$ \\
InternVL3.5-38B & 41.20 $\rightarrow$ 21.37 & $-19.83$ & 22.73 $\rightarrow$ 11.81 & $-10.92$ \\
\bottomrule
\end{tabular}
}
\end{subtable}

\par\vspace{0.65em}

\begin{subtable}{\linewidth}
\centering
\caption{Distribution of additional sequence requests.}
\label{tab:interactive_requests}
\setlength{\tabcolsep}{10pt}
\renewcommand{\arraystretch}{1.15}
{\footnotesize
\begin{tabular}{@{}lrrrr@{}}
\toprule
Model & 0 requests & 1 request & $\geq$2 requests & Mean sequences used \\
\midrule
Gemma-4-31B & 12.0 & 88.0 & 0.0 & 1.88 \\
InternVL3.5-38B & 0.6 & 31.9 & 67.5 & 2.72 \\
\bottomrule
\end{tabular}
}
\end{subtable}

\par\vspace{0.65em}

\begin{subtable}{\linewidth}
\centering
\caption{Confidence and diagnostic accuracy by request count.}
\label{tab:interactive_confidence}
\setlength{\tabcolsep}{10pt}
\renewcommand{\arraystretch}{1.15}
{\footnotesize
\begin{tabular}{@{}lrrl@{}}
\toprule
Model & Conf. (correct) & Conf. (incorrect) & Mean confidence by round \\
\midrule
Gemma-4-31B & 0.897 & 0.899 & $0.78\rightarrow0.90$ \\
InternVL3.5-38B & 0.973 & 0.975 & $0.81\rightarrow0.88\rightarrow0.96\rightarrow0.99$ \\
\bottomrule
\end{tabular}
}

\par\vspace{0.45em}
{\footnotesize\textbf{Accuracy by number of additional requests (\%)}\par}
\par\vspace{0.2em}

{\footnotesize
\begin{tabular}{@{}lrrrr@{}}
\toprule
Model & 0 & 1 & 2 & 3 \\
\midrule
Gemma-4-31B & 45.1 & 39.6 & -- & -- \\
InternVL3.5-38B & 40.0 & 22.6 & 21.0 & 13.3 \\
\bottomrule
\end{tabular}
}
\end{subtable}

\par\vspace{0.5em}
{\footnotesize\raggedright\noindent
\textit{Notes.} Mean sequences include the initial Cine sequence.
Gemma requests only LGE; InternVL requests LGE (821), T2 (486), and Perfusion (152).
Confidence is self-reported on a 0--1 scale.
Accuracy by request count refers to different case groups and does not measure the causal effect of additional requests.\par}

\end{table}

\subsubsection{Diagnostic Performance and Acquisition Behavior}
\label{app:interactive_results}
Neither model matches the static four-sequence baseline under
interactive access (Table~\ref{tab:interactive_performance}).
Gemma-4-31B loses 10.98 accuracy points and 3.02 macro-F1
points, while InternVL3.5-38B loses 19.83 and 10.92 points,
respectively. These results indicate that interactive
sequence acquisition does not recover the performance
achieved with full sequence availability in this setting.

The models also exhibit distinct acquisition behaviors
(Table~\ref{tab:interactive_requests}). Gemma-4-31B requests
exactly one additional sequence in 88.0\% of evaluations,
always selecting LGE, and uses 1.88 sequences on average.
In contrast, InternVL3.5-38B requests at least two additional
sequences in 67.5\% of evaluations, drawing from LGE, T2,
and Perfusion, with an average of 2.72 sequences.
Despite requesting more diverse evidence, InternVL3.5-38B still falls substantially below its static baseline, suggesting that acquiring additional sequences does not necessarily lead to better diagnostic performance.

\subsubsection{Confidence, Correctness, and Stopping}
\label{app:interactive_confidence}

Self-reported confidence shows little separation between
correct and incorrect diagnoses
(Table~\ref{tab:interactive_confidence}): 0.897 versus 0.899
for Gemma-4-31B, and 0.973 versus 0.975 for
InternVL3.5-38B. Meanwhile, mean confidence increases
across acquisition rounds, from 0.78 to 0.90 for Gemma
and from 0.81 to 0.99 for InternVL. These observations
suggest that increasing reported confidence does not
necessarily reflect diagnostic correctness. Together with
the limited interactive performance, they highlight the
difficulty of using self-reported uncertainty to guide
sequence acquisition and stopping decisions.

\clearpage

\section{Clinical Reasoning Prompts and Prediction Behavior}
\label{app:reasoning_analysis}

Section~\ref{sec:reasoning_effect} shows that clinical reasoning prompts alter model predictions without consistently improving performance. Here, we further examine the effects of clinical frameworks and answer--rationale ordering through controlled prompt comparisons. We then present an LGE case study illustrating how a seemingly coherent clinical explanation can accompany predictions that contradict the underlying visual evidence. Prompt templates are provided in Appendix~\ref{app:prompts}.

\subsection{Clinical Framework and Output Order}
\label{app:prompt_order_controls}

\paragraph{Purpose and experimental conditions.}
The original reasoning prompt both supplies a clinical analysis framework and requests an explanation after the answer. To examine the contribution of these choices, we compare four conditions on Level-1 image understanding for MedGemma-4B and Gemma-4-31B:
\begin{itemize}[leftmargin=*, itemsep=1pt]
    \item \textbf{R0: standard prompt.} The model provides only its answer, with no rationale requested.
    \item \textbf{R1: answer-first with framework.} The answer is followed by a justification guided by the task-specific clinical framework. This is the original reasoning-prompt condition.
    \item \textbf{R2: answer-first without framework.} The framework is removed, but the model is still asked to provide an answer followed by a rationale.
    \item \textbf{R3: reason-first with framework.} The framework is retained, but the model is asked to produce its clinical reasoning before the final answer.
\end{itemize}
R0 and R1 correspond to the standard and reasoning-prompt results in Table~\ref{tab:reasoning_comparison_random}. Comparing R1 with R2 examines the framework under answer-first output; comparing R1 with R3 examines output order with the framework retained. In R1 and R2, the explanation is post-answer text, whereas R3 explicitly places the rationale before the decision.

\begin{table}[htbp]
\centering
\small

\caption{Level-1 clinical-framework and output-order controls with random selection. Scores are percentages. R0: answer only; R1: answer then rationale, with framework; R2: answer then rationale, without framework; R3: rationale then answer, with framework.}
\label{tab:prompt_order_control}
\begin{tabular}{llrrrr}
\toprule
Model & Setting & Accuracy & Precision & Recall & F1 \\
\midrule
MedGemma-4B & R0 & 58.72 & 41.83 & 67.44 & 51.53 \\
& R1 & 47.17 & 37.76 & 53.77 & 44.24 \\
& R2 & 47.24 & 36.65 & 56.50 & 44.46 \\
& R3 & 60.19 & 48.27 & 49.33 & 48.79 \\
\midrule
Gemma-4-31B & R0 & 61.83 & 46.62 & 57.86 & 51.61 \\
& R1 & 64.80 & 52.42 & 57.81 & 54.95 \\
& R2 & 61.11 & 45.96 & 52.02 & 48.80 \\
& R3 & 62.36 & 47.16 & 51.04 & 49.02 \\
\bottomrule
\end{tabular}
\end{table}

\paragraph{Effect of the clinical framework (R1 vs.\ R2).}
Removing the framework changes MedGemma-4B's F1 only slightly, from 44.24 to 44.46, while recall rises from 53.77 to 56.50. The model still remains below its standard-prompt recall of 67.44 and F1 of 51.53. For Gemma-4-31B, removing the framework reduces F1 from 54.95 to 48.80 and recall from 57.81 to 52.02. The framework therefore benefits Gemma in this comparison and has little effect on MedGemma's F1. Removing the enumerated clinical rules does not recover standard-prompt performance in either model, so the recall loss cannot be attributed simply to the rules being too restrictive.

\paragraph{Effect of generating the rationale first (R1 vs.\ R3).}
Moving the rationale before the answer has different effects in the two models. MedGemma-4B improves from 44.24 to 48.79 F1 relative to R1, while Gemma-4-31B declines from 54.95 to 49.02. Thus, correcting the output order changes the result, but not in a uniformly beneficial direction.

The comparison with R0 is also mixed. Under R3, both models gain accuracy and precision, but lose recall and F1. MedGemma's recall falls from 67.44 to 49.33 and F1 from 51.53 to 48.79; Gemma's recall falls from 57.86 to 51.04 and F1 from 51.61 to 49.02. The respective F1 decreases are 2.74 and 2.59 points. Reason-first prompting therefore does not eliminate the lower-recall pattern, even though it improves some other metrics.

\paragraph{Interpretation.}
The effects of the clinical framework and output order vary across models, with neither consistently improving performance. Reasoning prompts often reduce recall, even when accuracy or precision improves, suggesting a shift toward more conservative predictions rather than more reliable identification of clinical findings. The following case illustrates how a model can provide a detailed clinical justification while still missing true abnormalities and reporting unsupported findings.

\subsection{Case Study of Clinical Reasoning Prompt}
\label{app:case_reasoning}

\begin{figure}[htbp]
\centering
\begin{subfigure}[b]{0.24\linewidth}
  \includegraphics[width=\linewidth]{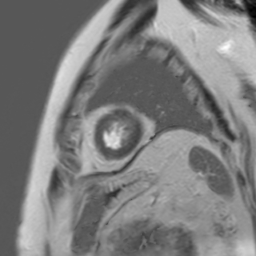}
  \caption{}\label{fig:case_f1}
\end{subfigure}\hfill
\begin{subfigure}[b]{0.24\linewidth}
  \includegraphics[width=\linewidth]{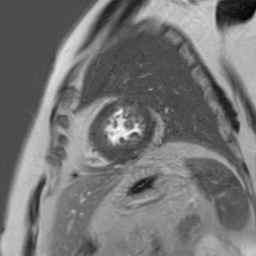}
  \caption{}\label{fig:case_f2}
\end{subfigure}\hfill
\begin{subfigure}[b]{0.24\linewidth}
  \includegraphics[width=\linewidth]{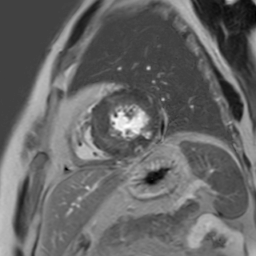}
  \caption{}\label{fig:case_f3}
\end{subfigure}\hfill
\begin{subfigure}[b]{0.24\linewidth}
  \includegraphics[width=\linewidth]{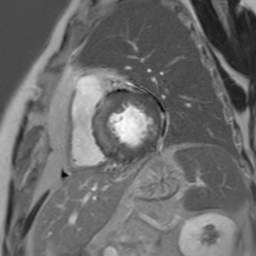}
  \caption{}\label{fig:case_f4}
\end{subfigure}

\vspace{2pt}

\begin{subfigure}[b]{0.24\linewidth}
  \includegraphics[width=\linewidth]{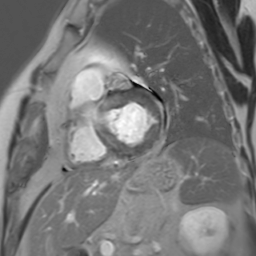}
  \caption{}\label{fig:case_f5}
\end{subfigure}\hfill
\begin{subfigure}[b]{0.24\linewidth}
  \includegraphics[width=\linewidth]{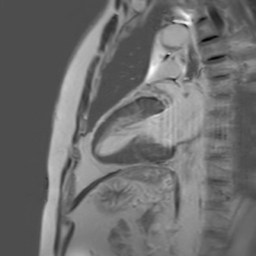}
  \caption{}\label{fig:case_f6}
\end{subfigure}\hfill
\begin{subfigure}[b]{0.24\linewidth}
  \includegraphics[width=\linewidth]{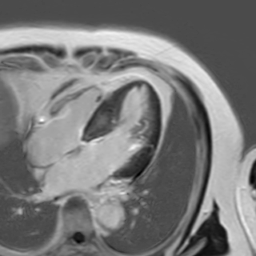}
  \caption{}\label{fig:case_f7}
\end{subfigure}\hfill
\begin{subfigure}[b]{0.24\linewidth}
  \includegraphics[width=\linewidth]{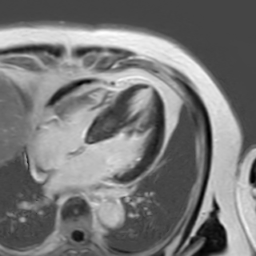}
  \caption{}\label{fig:case_f8}
\end{subfigure}

\caption{LGE frames provided to the model in the case study. Myocardial late
enhancement is visible in the \textbf{basal} and \textbf{mid} segments across
multiple slices; the apical segment, apex, and papillary muscles are spared.
Ground truth: A (Basal), B (Mid).}
\label{fig:case_reasoning}
\end{figure}

To illustrate how clinical explanations may diverge from visual evidence, we compare the standard (R0) and answer-first reasoning (R1) prompts on an LGE multi-segment-selection case. Both conditions use the same model and the same eight LGE frames (Figure~\ref{fig:case_reasoning}), with confirmed high-signal enhancement in the basal and mid segments (ground truth: \textbf{A, B}).

We query the same MLLM with identical visual inputs under two prompt
conditions.

\noindent\textit{Standard prompt:}
\begin{tcolorbox}[colback=gray!10, colframe=gray!50, boxrule=0.5pt, arc=2pt,
  left=6pt, right=6pt, top=4pt, bottom=4pt, breakable]
\small
\begin{verbatim}
You are a Vision-Language Model (VLM) for cardiac LGE MRI.
Task: Using the provided image frames, select which segments show
high-signal abnormal enhancement (multi-select).

Strict rules:
- Select one or more from A/B/C/D/E.
- Output letters only; use English commas for multiple selections
  (e.g., B,C,D).
- No explanations, reasoning, descriptions, confidence, or extra
  characters.

Options:
A. Basal Segment
B. Mid Segment
C. Apical Segment
D. Apex
E. Papillary Muscle
\end{verbatim}
\end{tcolorbox}

\noindent\textit{Model output:}
\begin{tcolorbox}[colback=gray!10, colframe=gray!50, boxrule=0.5pt, arc=2pt,
  left=6pt, right=6pt, top=4pt, bottom=4pt, breakable]
\texttt{A,B,C,D,E}
\end{tcolorbox}

\noindent\textit{Reasoning prompt:}
\begin{tcolorbox}[colback=gray!10, colframe=gray!50, boxrule=0.5pt, arc=2pt,
  left=6pt, right=6pt, top=4pt, bottom=4pt, breakable]
\small
\begin{verbatim}
You are a Vision-Language Model (VLM) for cardiac LGE MRI.
Task: Using the provided image frames, select which segments show
 high-signal abnormal enhancement (multi-select).

Strict rules:
- Select one or more from A/B/C/D/E.
- After providing your answer, you must provide a reason for your
  choice.
- Output format:
  - Line 1: Answer letter(s) only (e.g., A or A,B)
  - Line 2: "Reason: [your explanation]"

When providing your reason, please follow this analysis framework
(image-based):
1) Segment identification: which parts of LV are visible.
2) Enhancement presence by segment: basal / mid / apical / apex
/ Papillary Muscle.
3) Distribution along long axis: where it starts and ends.
4) Consistency across frames: persists vs transient artifact.
5) Final selected segments.

Options:
A. Basal Segment
B. Mid Segment
C. Apical Segment
D. Apex
E. Papillary Muscle
\end{verbatim}
\end{tcolorbox}

\noindent\textit{Model output:}
\begin{tcolorbox}[colback=gray!10, colframe=gray!50, boxrule=0.5pt, arc=2pt,
  left=6pt, right=6pt, top=4pt, bottom=4pt, breakable]
\texttt{B,C,D}\\[3pt]
\textbf{Reason:}\\
1) Segment identification: the images show short- and long-axis views of the
left ventricle.\\
2) Enhancement presence by segment: a prominent high-signal region of late
gadolinium enhancement is visible in the mid-ventricular and apical
myocardium.\\
3) Distribution along long axis: the high-signal area begins in the mid
segment and extends through the apical segment, involving the apex itself;
\textbf{the basal segment appears relatively spared}.\\
4) Consistency across frames: the enhancement is consistent across multiple
slices, confirming a real myocardial finding rather than a transient
artifact.\\
5) Final selected segments: Mid (B), Apical (C), Apex (D).
\end{tcolorbox}

\begin{table}[h]
\centering
\caption{Standard and answer-first reasoning responses in the LGE case. TP, FP, and FN list the true-positive, false-positive, and false-negative segments; P/R/F1 are percentages.}
\label{tab:case_reasoning_comparison}
\small
\setlength{\tabcolsep}{6pt}
\begin{tabular}{lccccc}
\toprule
Prompt & Prediction & TP & FP & FN & P / R / F1 (\%) \\
\midrule
Standard  & A,B,C,D,E & \{A,B\} & \{C,D,E\} & $\emptyset$  & 40.0 / \textbf{100.0} / 57.1 \\
Reasoning & B,C,D     & \{B\}   & \{C,D\}   & \{A\}        & 33.3 / \textbf{50.0}  / 40.0 \\
\bottomrule
\end{tabular}
\end{table}

\paragraph{Analysis.}
Under the standard prompt, the model selects all five segments, capturing both true abnormalities but also introducing three false positives. With the answer-first reasoning prompt, it selects B, C, and D, omitting the true-positive basal segment (A) while retaining false positives C and D. Consequently, recall falls from 100\% to 50\%, precision from 40.0\% to 33.3\%, and F1 from 57.1\% to 40.0\%.
Despite the lower performance, the model provides a structured clinical explanation that states ``the basal segment appears relatively spared'' and describes enhancement extending toward the apex. Both statements contradict the ground-truth distribution, yet appear to justify the incorrect selection. This case illustrates how a seemingly coherent clinical explanation can accompany an anatomically inaccurate prediction without ensuring that the reported findings are supported by visual evidence. Such errors may distort the assessment of LGE distribution, an important aspect of clinical CMR interpretation.

\clearpage

\section{Broader Impacts and Limitations}
\label{app:impacts_limitations}

CardioLens aims to advance the evaluation of medical MLLMs by assessing their capabilities across image understanding, clinical reporting, and patient-level diagnosis. Its findings can inform future model development and more reliable assessments of clinical capabilities, while the current dataset and evaluation protocols impose several limitations.

\paragraph{Positive impacts.}
CardioLens provides a clinically grounded benchmark for evaluating medical MLLMs beyond isolated image-level recognition. By quantifying the performance gap between image understanding, clinical reporting, and patient-level diagnosis, it can (i) encourage more careful assessments of clinical capabilities, (ii) motivate research into models that better capture 3D anatomy, temporal dynamics, and multi-sequence evidence integration, and (iii) offer a consistent reference for tracking progress as models capable of native 3D/4D medical image processing emerge.

\paragraph{Risks and mitigations.}
Two considerations are particularly important. \textit{(i) Premature clinical reliance:} despite stronger performance on image-level tasks, current MLLMs remain unreliable in patient-level diagnosis and may favor prevalent diagnostic categories over less common conditions. Benchmark scores should therefore not be interpreted as evidence of clinical readiness or used to justify clinical deployment. \textit{(ii) Data privacy:} all data were collected under IRB approval and de-identified before benchmark construction. The public release is restricted to a 100-patient subset, distributed under a data-use agreement that prohibits re-identification and restricts clinical use.

\paragraph{Intended use.}
CardioLens is intended for research on medical MLLM evaluation, high-dimensional CMR interpretation, and methodological studies of model inputs and clinical assessment. It is not intended for clinical decision-making, patient triage, or automated diagnosis.

\paragraph{Scope and limitations.}
Several aspects of the dataset and evaluation protocols limit the generalizability of the current findings.

\textit{(i) Single-institution sourcing:}
CardioLens is constructed from a single hospital's archives, providing a consistent reporting context but leaving cross-site generalization untested. Model performance may differ across institutions, scanner characteristics, clinical populations, and reporting conventions.

\textit{(ii) Sequence and disease coverage:}
The benchmark focuses on four major CMR sequences (Cine, LGE, T2, and Perfusion) and common diagnostic categories in etiology classification and NICM subtyping. Additional sequences, including T1 mapping, T2 mapping, and 4D flow, are not covered, while rare cardiomyopathies such as ACM and RCM remain under-represented.

\textit{(iii) Scope of reader evaluation:}
The independent reader study involves a single radiologist performing three-class etiology classification using the same selected images as the models. It does not measure inter-observer variability or establish human performance on finer-grained diagnostic tasks (Appendix~\ref{app:clinical_references}).

\textit{(iv) Selected-image interfaces:}
Current evaluations rely on selected slices and frames to accommodate existing MLLMs. The interactive experiment allows sequence-level requests but not navigation across individual slices, views, or temporal phases (Appendix~\ref{app:interactive_control}). The results therefore do not establish how these models would perform with native 3D/4D inputs or more flexible interactive access. The preserved NIfTI studies provide a basis for investigating these settings in future work.

\textit{(v) Public release scope:}
Institutional data-governance requirements limit public release to a de-identified 100-patient subset. Level-1 and Level-2 comparisons show broadly consistent performance patterns between the subset and full cohort, whereas smaller class counts and differences in disease prevalence make Level-3 results less directly comparable (Appendix~\ref{app:subset_consistency}).

\textit{(vi) Coverage of controlled analyses:}
The image-blind analysis covers two models on Level-1 LGE, while the clinical-framework and output-order controls examine two models. The QLoRA experiment is restricted to Level-1 adaptation of language-model projections, with the visual encoder frozen, leaving visual-encoder adaptation and fine-tuning for other task tiers unexplored. Broader comparisons across models and task tiers, together with input-budget analyses for individual Cine motion and function subtasks, remain directions for future work.

\clearpage
\section*{NeurIPS Paper Checklist}

\begin{enumerate}

\item {\bf Claims}
    \item[] Question: Do the main claims made in the abstract and introduction accurately reflect the paper's contributions and scope?
    \item[] Answer: \answerYes{}
    \item[] Justification: The abstract and introduction state our contributions: a clinically grounded multi-sequence CMR testbed (CardioLens) with 473{,}896 slices and 13{,}494 verified QA pairs, a three-tier evaluation (image understanding, report generation, discharge-grounded diagnosis), and empirical findings of a clinical reality gap -- as supported by Sections~\ref{sec:CardioLens} and~\ref{sec:experiments_analysis} and the Appendix.

\item {\bf Limitations}
    \item[] Question: Does the paper discuss the limitations of the work performed by the authors?
    \item[] Answer: \answerYes{}
    \item[] Justification: \CRrecent{Appendix~\ref{app:impacts_limitations} discusses single-institution sourcing, restricted sequence and disease coverage, the single-reader scope of the independent radiologist evaluation, selected-image input constraints, and the limited public-release scope.}

\item {\bf Theory assumptions and proofs}
    \item[] Question: For each theoretical result, does the paper provide the full set of assumptions and a complete (and correct) proof?
    \item[] Answer: \answerNA{}
    \item[] Justification: This is an empirical evaluation paper and contains no theoretical results requiring formal proof.

    \item {\bf Experimental result reproducibility}
    \item[] Question: Does the paper fully disclose all the information needed to reproduce the main experimental results of the paper to the extent that it affects the main claims and/or conclusions of the paper (regardless of whether the code and data are provided or not)?
    \item[] Answer: \answerYes{}
    \item[] Justification: 
    Section~\ref{sec:data_construction} details the data construction pipeline; 
    Section~\ref{sec:slice_selection} specifies all three slice-selection protocols with details stated in Appendix~\ref{app:heuristic}, ~\ref{sec:slice-selection-across-task-tiers}, and~\ref{app:mil} (architecture, loss, hyperparameters, algorithms, usage instructions); 
    Appendix~\ref{app:experiments} specifies models, prompts, and metrics.

\item {\bf Open access to data and code}
    \item[] Question: Does the paper provide open access to the data and code, with sufficient instructions to faithfully reproduce the main experimental results, as described in supplemental material?
    \item[] Answer: \answerYes{}
    \item[] Justification: \CRrecent{Section~\ref{sec:data-governance} provides a repository link for CardioLens data and code and specifies the de-identified 100-patient subset permitted for public release. The full 585-patient cohort remains restricted by institutional data-sharing requirements (Appendix~\ref{app:public_subset}).}

\item {\bf Experimental setting/details}
    \item[] Question: Does the paper specify all the training and test details (e.g., data splits, hyperparameters, how they were chosen, type of optimizer) necessary to understand the results?
    \item[] Answer: \answerYes{}
    \item[] Justification: \CRrecent{Section~\ref{sec:experimental_setup} specifies the main zero-shot evaluation protocol; Appendix~\ref{app:model-list} lists the evaluated MLLMs; Appendix~\ref{app:prompts} provides the prompt templates; Appendix~\ref{app:mil_data_split} details the MIL data splits; Appendix~\ref{app:metrics} defines the evaluation metrics; and Appendix~\ref{app:mil:impl} lists MIL training hyperparameters (Table~\ref{tab:mil:config}). Additional diagnostic, reader, adaptation, and interactive evaluations are described in Appendices~\ref{app:diagnostic_controls},~\ref{app:independent_reader},~\ref{app:adaptation_controls}, and~\ref{app:interactive_control}.}

\item {\bf Experiment statistical significance}
    \item[] Question: Does the paper report error bars suitably and correctly defined or other appropriate information about the statistical significance of the experiments?
    \item[] Answer: \answerYes{}
    \item[] Justification: \CRrecent{Table~\ref{tab:image_understanding} reports mean$\pm$std across three inference runs for Random and MIL selection (Heuristic is deterministic). Table~\ref{tab:mil:perf} reports mean$\pm$std over three independently trained MIL models on the slice-selection cohort.}

\item {\bf Experiments compute resources}
    \item[] Question: For each experiment, does the paper provide sufficient information on the computer resources (type of compute workers, memory, time of execution) needed to reproduce the experiments?
    \item[] Answer: \answerYes{}
    \item[] Justification: \CRrecent{Open-source MLLM inference uses NVIDIA H100 GPUs (80\,GB), while the MIL framework is trained on a single NVIDIA RTX~4090 ($\sim$24\,GB; Appendix~\ref{app:mil:impl}). Closed-source MLLMs are evaluated through their official APIs.}
    
\item {\bf Code of ethics}
    \item[] Question: Does the research conducted in the paper conform, in every respect, with the NeurIPS Code of Ethics \url{https://neurips.cc/public/EthicsGuidelines}?
    \item[] Answer: \answerYes{}
    \item[] Justification: All CMR data were collected under IRB approval and fully de-identified before benchmark construction (Section~\ref{sec:data-governance}).

\item {\bf Broader impacts}
    \item[] Question: Does the paper discuss both potential positive societal impacts and negative societal impacts of the work performed?
    \item[] Answer: \answerYes{}
     \item[] Justification: Appendix~\ref{app:impacts_limitations} discusses positive impacts (more reliable assessment of clinical capabilities and guidance for high-dimensional medical interpretation), potential risks (premature clinical reliance and data privacy), and the corresponding mitigations and intended-use restrictions. Single-institution sourcing and other scope limitations are discussed separately in the same appendix.
    
\item {\bf Safeguards}
    \item[] Question: Does the paper describe safeguards that have been put in place for responsible release of data or models that have a high risk for misuse (e.g., pre-trained language models, image generators, or scraped datasets)?
    \item[] Answer: \answerYes{}
    \item[] Justification: All data are de-identified; the public release is limited to a 100-patient subset and will be distributed under a data-use agreement prohibiting re-identification and unsanctioned clinical deployment (Section~\ref{sec:data-governance}, Appendix~\ref{app:impacts_limitations}).

\item {\bf Licenses for existing assets}
    \item[] Question: Are the creators or original owners of assets (e.g., code, data, models), used in the paper, properly credited and are the license and terms of use explicitly mentioned and properly respected?
    \item[] Answer: \answerYes{}
    \item[] Justification: All evaluated MLLMs are cited with their original publications or system cards (Appendix~\ref{app:model-list}, Table~\ref{tab:model-list}) and used in accordance with their respective open-source or API licenses; the ResNet-18 backbone is used under the standard torchvision license.

\item {\bf New assets}
    \item[] Question: Are new assets introduced in the paper well documented and is the documentation provided alongside the assets?
    \item[] Answer: \answerYes{}
    \item[] Justification: \CRrecent{CardioLens is a new asset; Section~\ref{sec:CardioLens} and Appendices~\ref{app:cardiolens_task-details},~\ref{app:heuristic}, and~\ref{app:mil} document its data composition, task taxonomy, label construction, slice-selection protocols, and evaluation procedures. Section~\ref{sec:data-governance} and Appendix~\ref{app:public_subset} specify the public-release scope and subset contents.}

\item {\bf Crowdsourcing and research with human subjects}
    \item[] Question: For crowdsourcing experiments and research with human subjects, does the paper include the full text of instructions given to participants and screenshots, if applicable, as well as details about compensation (if any)? 
    \item[] Answer: \answerYes{}
    \item[] Justification: \CRrecent{No crowdsourcing was involved. Annotation followed a two-tier clinical reporting workflow with subsequent label verification (Section~\ref{sec:data_construction}). Appendix~\ref{app:independent_reader} also describes a blinded independent radiologist evaluation of 98 etiology cases using the same selected-image inputs provided to the compared models.}

\item {\bf Institutional review board (IRB) approvals or equivalent for research with human subjects}
    \item[] Question: Does the paper describe potential risks incurred by study participants, whether such risks were disclosed to the subjects, and whether Institutional Review Board (IRB) approvals (or an equivalent approval/review based on the requirements of your country or institution) were obtained?
    \item[] Answer: \answerYes{}
    \item[] Justification: \CRrecent{All CMR studies were collected under the Institutional Review Board (IRB) approval of Beijing Anzhen Hospital affiliated to Capital Medical University (2025216x) and de-identified prior to benchmark construction (Section~\ref{sec:data-governance}).}

\item {\bf Declaration of LLM usage}
    \item[] Question: Does the paper describe the usage of LLMs if it is an important, original, or non-standard component of the core methods in this research? Note that if the LLM is used only for writing, editing, or formatting purposes and does \emph{not} impact the core methodology, scientific rigor, or originality of the research, declaration is not required.
    \item[] Answer: \answerNA{}
    \item[] Justification: LLMs/MLLMs are the subjects of evaluation in this work, not components of the core methodology. No LLMs were used as important, original, or non-standard components of the methods, dataset construction, or analysis.

\end{enumerate}

\end{document}